\RequirePackage{fix-cm}
\PassOptionsToPackage{table}{xcolor}
\PassOptionsToPackage{hypertexnames=false}{hyperref}
\documentclass{fpt_seed}
\usepackage{manuscript-support}
\usepackage{toc-times}
\usepackage{fontawesome5}
\newcommand{\BibTeX}{\rm B\kern-.05em{\sc i\kern-.025em b}\kern-.08em\TeX}

\newcommand{\method}{\mbox{\textnormal{\textsc{Agentic4PDM}}}}
\newcommand{\protocol}{\mbox{\textnormal{\textsc{CCGA}}}}
\newcommand{\agent}[1]{\textnormal{\textsc{#1}}}
\newcommand{\host}{\texorpdfstring{\textnormal{Host}}{Host}}
\providecommand{\rev}[1]{#1}
\providecommand{\revcolor}{}
\providecommand{\revr}[1]{#1}
\newcommand{\revred}[1]{#1}
\newcommand{\newred}[1]{#1}

\newcommand{\evidencestream}[1]{\hspace{0.75em}\textbf{#1}~{\scriptsize(ev.)}}
\newcommand{\bluetick}{\textcolor{blue!70!black}{\ding{51}}}
\newcommand{\redcross}{\textcolor{red!75!black}{\ding{55}}}

\definecolor{gainxgb}{HTML}{E9C46A}
\definecolor{gainsvm}{HTML}{F4A261}
\definecolor{gainagent}{HTML}{2F80ED}
\definecolor{gainarrow}{HTML}{667085}

\newcommand{\system}{\textnormal{\textsc{Agentic4PDM}}\xspace}

\definecolor{jColor}{HTML}{1F4E79}
\definecolor{hColor}{HTML}{1E6B3A}
\definecolor{wColor}{HTML}{9C4A00}
\newcommand{\jturn}[2]{\textcolor{jColor}{\textbf{#1}}\ \textcolor{jColor}{\small#2}}
\newcommand{\hturn}[2]{\textcolor{hColor}{\textbf{#1}}\ \textcolor{hColor}{\small#2}}
\newcommand{\stg}[2]{\textcolor{hColor}{\textbf{#1}}\ \textcolor{hColor}{\small#2}}
\newcommand{\note}[1]{\textit{#1}}
\newcommand{\lab}[1]{\textcolor{black!55}{\scriptsize\textsf{#1}}}
\newcommand{\hsep}{\vspace{5pt}\hrule\vspace{5pt}}
\newcommand{\seriesstrip}[1]{%
  \begin{tcolorbox}[colback=black!4,colframe=black!18,boxrule=0.4pt,arc=1.5pt,
    left=6pt,right=6pt,top=3pt,bottom=3pt]
  {\scriptsize\textsf{How \method{} reaches a diagnosis, in five steps (Appendix \tracesec{1}--\tracesec{5}), all on the same \newred{sample}:}\\[2pt]
  \textsf{#1}}
  \end{tcolorbox}}
\newcommand{\traceapp}{\ref*{app:agent-workflow-trace}}
\newcommand{\tracesec}[1]{\traceapp.#1}
\newcommand{\stepon}{\textcolor{black!35}{\tracesec{1} Evidence Construction}}
\newcommand{\steptwo}{\textcolor{black!35}{\tracesec{2} Comparative Reasoning}}
\newcommand{\stepthree}{\textcolor{black!35}{\tracesec{3} Request Gate}}
\newcommand{\stepfour}{\textcolor{black!35}{\tracesec{4} Packet \& Influence Gate}}
\newcommand{\stepfive}{\textcolor{black!35}{\tracesec{5} Finalization}}
\newcommand{\here}[1]{\textcolor{wColor}{\textbf{#1}}}

\author[1]{Nam V. Nguyen}
\author[1]{Hoang Van Truong}
\author[1]{Hieu D. D.}
\author[1]{Nguyen Duy Khuong}
\author[2]{Quang Pham}
\author[3]{Tung Kieu}
\affiliation[1]{FPT Corporation}
\affiliation[2]{Independent Researcher}
\affiliation[3]{Aalborg University}
\correspondence{\email{namnguyen16ai@gmail.com}}
\newcommand{\pdfauthors}{Nam V. Nguyen; Hoang Van Truong; Hieu D. D.; Nguyen Duy Khuong; Quang Pham; Tung Kieu}

\title{Agentic4PDM: Host-Governed Evidence Admissibility for Industrial Fault Diagnosis}
\abstract{
We study industrial fault diagnosis as a classification task that assigns sensor time-series windows to predefined health states or fault types. Scarce labels and changing operating conditions make repeated training of specialized classifiers costly. Yet structures such as periodicity, trends, and transients can remain diagnostically informative and be examined through reusable analytical tools. LLM agents can select and interpret these tools, but fixed rules or training-derived guidance may not resolve each signal’s ambiguity, while unconstrained interpretation may overstate weak or conflicting results. We introduce \system{}, a framework that separates adaptive investigation from control over evidence use. LLM agents compare each signal with labeled references from every candidate class, identify unresolved distinctions, and propose follow-up measurements. A deterministic software controller, called the host, enforces contract-checked gated adjudication (CCGA): it checks requests, requires commitment to a complete tool batch before results are observed, and assigns evidence roles based on tool checks against labeled references. These roles distinguish corroborative support from advisory information. The Judge Agent proposes a diagnosis, while the controller checks protocol compliance and records the investigation. \system{} supports training-free diagnosis using pretrained representations and labeled references without task-specific diagnostic models, and a standard setting that additionally investigates trained-model predictions. Experiments on selected bearing and motor diagnosis challenges and an unfiltered industrial test set supplied by our customer show gains in aggregate macro F1 over the evaluated diagnostic and agentic baselines. Ablations further support the value of the governed workflow.
}

\checkdata[Keywords]{LLM agents; agent governance; agents for predictive maintenance}
\renewcommand{\titlelinks}{%
  \href{https://david-nguyen-s16.github.io/agentic4pdm_homepage/}{\shortstack[c]{{\Large\faGlobe}\\[-1pt]{\footnotesize Project page}}}%
}
\hypersetup{
  pdftitle={Agentic4PDM: Host-Governed Evidence Admissibility for Industrial Fault Diagnosis},
  pdfauthor={\pdfauthors},
  pdfsubject={Industrial fault diagnosis with host-governed LLM agents},
  pdfkeywords={LLM agents, agent governance, agents for predictive maintenance}
}

\begin{document}
\maketitle
\section{Introduction}
\label{sec:introduction}

Unexpected machine failures disrupt production and incur costs beyond the repair itself. \citet{Thomas2021Maintenance} estimated annual losses from unplanned downtime in U.S. discrete manufacturing at \$18.4 billion in 2016. Industrial fault diagnosis identifies the current health state or fault type from condition-monitoring data \cite{Jardine2006CBMReview}, supporting maintenance decisions and the coordination of repair resources and production schedules \cite{Bousdekis2021MaintenanceDecisions}. Unsupervised methods can detect departures from normal behavior, but identifying the specific fault type requires information linking the observed deviation to known fault categories. Labeled fault examples provide such information, but are often scarce~\cite{Lai2025FewShotFD}, and this limitation becomes especially severe under distribution shift. Changes in operating conditions or measurement quality can alter the observed signal \cite{Chen2025DGSurvey}, meaning that a model trained under one condition may need to be retrained and revalidated using labeled faults from the new condition, even though such data are difficult to obtain. This raises a fundamental question: \emph{how can fault diagnosis adapt to a new operating condition when only a few labeled fault examples are available?}

Time-series foundation models offer reusable representations learned through large-scale pretraining \cite{liang2024tsfmsurvey}. A common approach maps a time series to a fixed-dimensional representation and fits a downstream classifier on top of it \citep{franceschi2019unsupervised,yue2022ts2vec,Goswami2024MOMENT,li2026starembed}. However, different classifiers may exploit different distinctions within the same representation: one classifier may correctly classify a sample that another misclassifies, while making errors on samples that the other handles correctly. Replacing one classifier with another can therefore shift errors across samples rather than eliminate the underlying ambiguity. This suggests that the most useful diagnostic perspective may depend on the sample, and that reliable diagnosis may require examining multiple aspects of the observed signal rather than relying on a single predictive view. In industrial diagnosis, such complementary evidence can arise from waveform morphology, recurrence, spectral structure, and other signal characteristics \cite{randall2011tutorial,smith2015bearingbenchmark,kuzio2023faultfrequency}. The problem then becomes more specific: \emph{when existing diagnostic views remain inconclusive, what additional evidence should be acquired for the current case?}

Large language model (LLM) agents can support this investigation by interleaving reasoning with actions, such as invoking analytical tools to acquire additional evidence for the current signal~\cite{yao2023react}. Existing time-series agents, however, guide evidence collection through predefined rules, feature knowledge, or aggregation procedures: ARGOS~\citep{gu2025argos} applies detection rules generated offline, ZARA~\citep{li2026zara} selects among discriminative features distilled from the training data, and FETA~\citep{sui2025feta} combines channel-level decisions using the LLM's self-assessed confidence as aggregation weights. These designs raise three concerns. First, a predefined pipeline can limit the agent's freedom to decide what to investigate. Second, rules or features selected for their discriminative power on training data may not resolve the ambiguity of the current case, particularly when operating conditions shift. Third, when the LLM rates its own confidence, that rating may be an unreliable basis for weighting evidence, since LLMs can struggle to assess their own reasoning without external feedback~\cite{huang2024cannot}. 

To address these concerns, we propose \method{}, a multi-agent framework for industrial fault diagnosis. Several LLM agents, each with its own role and context, first build comparative evidence for the current case. The analysts compare stable regions and local deviations in the signal with labeled references from every candidate class, providing complementary views of the same input. The \agent{Judge} agent then investigates the case under \emph{contract-checked gated adjudication} (\protocol{}), a protocol that controls how the agent collects and uses evidence. \protocol{} relies on a \host{}, a deterministic software controller that applies predefined rules. Like a standard agent harness~\citep{yang2024sweagent,wu2024autogen}, the \host{} calls each agent, prepares its context, and executes its requested tools. Here, it also checks the agent's requests and determines how each result may be used. The \host{} enforces two gates, followed by a finalization check on the proposed class. The \emph{request gate} guides the agent toward tools that address the ambiguity or conflict in the current case. The agent must state which signal properties remain unresolved and commit to a batch of tools that measure them before seeing any result from that batch. This follows four steps: \textsc{Declare} the unresolved properties, \textsc{Offer} compatible tools, \textsc{Commit} to the complete batch, and \textsc{Execute} the approved measurements. The \emph{packet and influence gate} then determines how each tool result may influence the diagnosis. A reference preflight checks the tool on labeled references retrieved for the current case and assigns its result a \textbf{corroborative} or \textbf{advisory} role. Corroborative results may support a class as checked evidence, whereas advisory results can inform hypotheses and further investigation but remain unchecked. The permitted influence of each result therefore rests on these reference checks rather than on the LLM's own confidence. Finally, the \host{} checks the proposed class against the finalization requirements and records the measurements, their assigned roles, and whether the proposal is admitted.

\begin{figure*}[!t]
\centering
\includegraphics[width=1\textwidth]{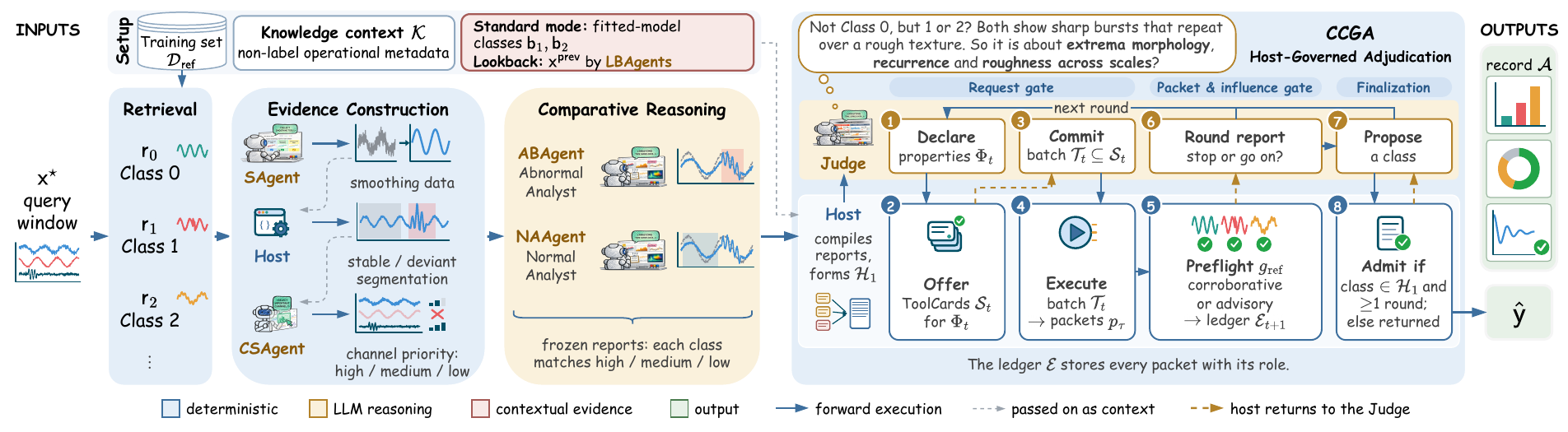}
\caption{\method{} workflow. The \host{} constructs and freezes class-balanced comparative evidence before the \agent{Judge} begins sample-specific investigation. The \agent{Judge} proposes follow-up measurements or a diagnosis, while the \host{} governs which measurements may run and how returned evidence may influence the decision. \revr{The \host{} then publishes the reported class, marked as admitted or not admitted, together with the record of the checks behind it.}}
\label{fig:stage2-inference}
\Description{The host chooses references and fixes how the current signal window will be viewed. Normal and Abnormal \agent{Analysts} then produce reports that cannot be changed. The \agent{Judge} may propose a class or ask for an extra measurement. Host checks decide which tools may run, test them on known references, and limit how each result may affect the diagnosis. At \revr{finalization}, the host reads the checked record rather than the raw signal, \revr{decides whether the proposed class is admitted, and publishes one class together with the record of the checks behind it}.}
\end{figure*}

Across the bearing and motor fault-diagnosis tasks, \method{} achieves the strongest overall macro-F1 performance, attaining the highest three-dataset averages on both BFH and GRS and the best result on the complete industrial test set provided by our customer. Our contributions are threefold:

\begin{enumerate}[noitemsep,topsep=2pt,leftmargin=*]

\item \textbf{Agentic4PDM: a context-rich framework for agentic industrial diagnosis.} It is an agentic framework that integrates complementary diagnostic context to support \rev{sample-specific investigation} in both training-free and standard settings.

\item \textbf{CCGA: Host-governed evidence adjudication.} We introduce a protocol that separates \emph{adaptive investigation} from \emph{evidentiary authority}, allowing the agent to decide what to investigate while a deterministic \host{} controls how each result may be used to support the diagnosis.

\item \textbf{Evaluation across industrial diagnosis challenges.} We evaluate \method{} across \rev{complementary bearing and motor fault-diagnosis settings}, examining both predictive performance and the behavior of governed investigation.

\end{enumerate}

\section{Related Work}
\label{sec:related}

\noindent\textit{Diagnosis with transferable representations.}
Industrial fault diagnosis must distinguish fault-related signal changes from those caused by operating conditions: load, rotational speed, and environmental factors can alter vibration signals even when component health remains unchanged, causing deployment data to differ from training data~\citep{Wei2021WDAN,Chen2022WheelDA}. This mismatch can reduce the accuracy of conventional fault classifiers~\citep{chen2020xaibearing,aburakhia2022bearing,kumar2023gearbox,Chen2025DGSurvey}, while adaptation is difficult because labeled fault data are scarce~\citep{Lai2025FewShotFD}. A common remedy is to train a classification head on pretrained time-series representations~\citep{franceschi2019unsupervised,yue2022ts2vec,li2026starembed}, including those from recent time-series foundation models~\citep{liang2024tsfmsurvey,das2024timesfm,garza2023timegpt,rasul2024lagllama,Goswami2024MOMENT,ansari2024chronos,ansari2025chronos2,woo2024moirai,liu2024moiraimoe,liu2026moirai20timeseries,ekambaram2024ttm,cohen2025toto,khwaja2026toto2,auer2025tirex,podest2026tirex2}. However, a fixed head returns a prediction using its learned decision boundaries without acquiring further evidence when the input is ambiguous or poorly represented in the training data. \method{} instead uses a pretrained encoder to retrieve class-balanced labeled references, while LLM agents investigate complementary aspects of the signal through analytical tools before proposing a class. Thus, the trained model predictions only served as consultation in \method{}.

\vspace{3pt}
\noindent\textit{Agentic evidence acquisition.}
LLM agents already reason over sensor data and acquire information beyond a fixed prediction: aligning sensor encoders with language~\citep{li2025sensorllm}, validating executable diagnostic rules offline~\citep{gu2025argos}, retrieving labeled exemplars and fusing channel judgments using self-assessed confidence~\citep{sui2025feta}, scoring predefined features for pairwise discriminativeness offline and pruning candidates online~\citep{li2026zara}, and orchestrating broader diagnostic or maintenance workflows~\citep{zhao2025timeseriesscientist,semas2026,lukens2024phmcopilot,he2024maintagt}. These designs raise two concerns: (i) predefined rules or historically discriminative features may not resolve the ambiguity of the current case; and (ii) when the same component selects evidence and determines its influence, self-assessed confidence provides no independent check when results conflict. \method{} addresses both: (i) the agent must state which signal properties remain unresolved in the current case and commit to a tool batch that covers them before seeing any result; and (ii) a deterministic \host{} determines each result's permitted influence through checks on labeled references retrieved for the current case, independently of the agent's own confidence.

\vspace{3pt}
\noindent\textit{Governing actions, data, and evidence.}
Related work provides three forms of agent governance. \emph{Interaction rules}, such as electronic institutions, regulated multi-agent systems, and commitment protocols, specify roles, permitted actions, and interaction obligations~\citep{esteva2001electronic,noriega2013regulated,yolum2004commitments}. \emph{Guardrails} integrated into the agent harness---the program that calls the LLM and executes its tools~\citep{yang2024sweagent,wu2024autogen}---enforce safety policies or restrict untrusted data flows~\citep{wang2026agentspec,chen2025shieldagent,xiang2025guardagent,debenedetti2025camel}. Some use an additional LLM to interpret policies or construct checking procedures~\citep{chen2025shieldagent,xiang2025guardagent}. \emph{Argumentation} studies which arguments or conclusions are acceptable given their supporting and conflicting relations~\citep{dung1995acceptability,cayrol2005bipolar,prakken2010structured}. These lines of work emphasize interaction permissions, safety and security policies, or argument acceptability. Our focus is the acquisition and use of diagnostic evidence during an adaptive investigation: whether a proposed measurement addresses the current ambiguity and what its result is permitted to support. This matters in time-series classification because the agent must choose measurements for each signal and interpret results that may conflict. Under \protocol{}, the agent retains initiative over \emph{what to investigate}, while a deterministic \host{} checks measurements against declared unresolved properties and controls \emph{what each result may establish}, from the first request to the admission of the final class (Section~\ref{sec:method}).

\section{Agentic4PDM Framework}
\label{sec:method}

\subsection{Framework Overview}
\label{sec:overview}

As illustrated in Figure~\ref{fig:stage2-inference}, \method{} consists of two main phases. (1)~In \emph{evidence construction} (Section~\ref{sec:stage2}), the \host{} retrieves one reference sample per class, the agents decide how the signal is viewed and which channels receive priority, and two \agent{Analysts} compare the query with each reference over complementary stable and deviant regions. (2)~In \emph{\protocol{}} (Section~\ref{sec:adjudication}), the \agent{Judge} resolves the ambiguity that remains after these comparisons: it requests additional measurements when needed and proposes a diagnosis, while three \host{} checks decide which measurements may run, how their results may influence the decision, and whether the proposed class is admitted. Notably, the LLM agents decide what to investigate, while the \host{} controls how the resulting evidence can be used in the final diagnosis.

\subsection{Problem Definition}
\label{sec:problem}
Let $x^\star \in \mathbb{R}^{T \times C}$ be a sample under diagnosis, with $T$ observation across $C$ sensor channels, and let $\mathcal{Y} = \{0, \dots, Y-1\}$ be the label space of $Y$ classes. Let $\mathcal D_{\mathrm{ref}}=\{(x_i,y_i)\}_{i=1}^{N}$ be the labeled training set, with $x_i\in\mathbb R^{T\times C}$ and $y_i\in\mathcal Y$. For each class $y\in\mathcal Y$, $\mathcal R_y=\{x_i:(x_i,y_i)\in\mathcal D_{\mathrm{ref}},\ y_i=y\}$ denotes the set of training samples with label $y$. Given a frozen embedding $\phi_{\mathrm{emb}}$ and the cosine similarity $s_{\mathrm{cos}}$ between embeddings, the class-balanced reference bank of a sample $x^\star$ holds the most similar training sample of each class:
\begin{equation}\label{eq:ret}
\begin{aligned}
  \mathcal R^\star(x^\star)&=\{r_0(x^\star),\dots,r_{Y-1}(x^\star)\},\\
  \textrm{ where } r_y(x^\star)&=\arg\max_{r\in\mathcal R_y,\;r\neq x^\star} s_{\mathrm{cos}}(x^\star,r).
\end{aligned}
\end{equation}
Thus $r_y(x^\star)$ is the class-$y$ sample closest to $x^\star$. Furthermore, we use $\mathcal{K}$ to denote a read-only knowledge context and metadata fields describing the operating condition of the input, such as machine types, rotational speeds, and recorded channels. Then \method{}, denoted $F(\cdot,\cdot,\cdot)$, is a diagnosis system comprising LLM agents and a deterministic \host{}. \method{} takes as input $x^\star$, $\mathcal R^\star$, and $\mathcal K$, and produces: (1)~the reported class $\hat y\in\mathcal Y$; and (2)~the record $\mathcal A$, a variable-length log of all actions taken and results revealed as:
\begin{equation}
(\hat{y}, \mathcal{A}) = F(x^\star, \mathcal{R}^\star, \mathcal{K}).
\label{eq:governed-inference}
\end{equation}
In many scenarios such as new machines or insufficient high quality reference samples, it is desirable for the system to notify that it is not confident in providing its prediction. \method{} achieved this behaviour via $\mathcal A$, which states whether $\hat y$ was \emph{admitted}, that is, whether it passed the \host{}'s finalization check. In all cases, $F$ always reports a valid class for every sample, however, $\mathcal A$ may indicate that this label is unreliable. 
We now describe the agents and the \host{} that make up $F$.

\subsection{Agents and Host}
\label{sec:agents-host}

\paragraph{Agents.}
\method{} uses five primary LLM agent roles, each with its own task and context. The \agent{Smoothing Agent} (\agent{SAgent}) selects a registered smoothing family for each channel; the \host{} determines its parameters. The \agent{Channel-Selection Agent} (\agent{CSAgent}) assigns each channel a priority tier by its diagnostic relevance, or removes it. The \agent{Normal Analyst} (\agent{NAAgent}) and \agent{Abnormal Analyst} (\agent{ABAgent}) compare the query's stable and deviant regions, respectively, with the class references and construct \agent{Analyst} reports that include a set of potential candidate classes, supporting observations, and contradictions. The \agent{Judge} synthesizes these reports, identifies unresolved signal properties, requests follow-up measurements, and proposes a diagnostic class. Then, the \agent{NAAgent} and \agent{ABAgent} repeat their respective comparisons on the preceding sample, collectively referred to as the \agent{Lookback Analysts} (\agent{LBAgents}). Their outputs serve only as context (Section~\ref{sec:adjudication}). 

We use separate agents instead of one LLM for every role. Specifically, each agent starts from a fresh context that holds only its own inputs and earlier agents’ outputs. This design provides two major advantages. First, it avoids context contamination where the thinking trace of one task is fed into next task's input context. Since only the final output is relevant, such thinking traces dilute the context and can quickly degrade performance~\citep{du-etal-2025-context,laban2026lost}, especially for small, local LLMs used in our work. Second, it provides a better tokenomics since using only one LLM for all tasks will quickly reach its token limit and require frequent context summarization, increasing costs and reducing inference speed. Separating the agents, therefore, keeps each context short and focused on one specific task. 
Lastly, Appendix~\ref{app:context-memory} describes how we manage the agents' contexts in more detail.

\paragraph{\host{}.}
The \host{} orchestrates the interactions among the \agent{SAgent}, \agent{CSAgent}, \agent{NAAgent}, \agent{ABAgent}, and \agent{LBAgents} to construct the evidence and \agent{Analyst} reports presented to the \agent{Judge}. It also governs how the \agent{Judge} progresses toward a diagnosis, from identifying the remaining ambiguity and selecting appropriate tools to incorporating the resulting evidence and proposing a final decision.


\paragraph{Two modes.}
In standard mode, $F$ also uses two fitted classifiers: the \agent{Judge} receives the classes $b_1,b_2\in\mathcal Y$ predicted for $x^\star$ by an SVM and an XGBoost classifier trained on frozen embeddings of the labeled samples. We call $b_1$ and $b_2$ the fitted-model predictions and treat them as individual contextual input sources that encourage the \agent{Judge} to pay greater attention to these candidate classes. The \agent{Judge} may use them as anchors and verify them by examining class characteristics that the \agent{Analyst} reports may have missed, using additional tools to investigate and validate the corresponding hypotheses. In this study, we report the standard mode as our main results. In Appendix~\ref{par:training-free-comparison}, we extend the evaluation to a training-free (TF) mode, where no classifier is trained and $b_1,b_2$ are omitted, to assess whether fitted-model predictions are necessary.

\subsection{Query-Adaptive Evidence Construction}
\label{sec:stage2}

Before the \agent{Judge} starts its analysis, the \host{} and the other agents prepare different types of evidence for the query sample $x^\star$. First, the \host{} retrieves the reference bank $\mathcal R^\star(x^\star)$ by Eq.~\ref{eq:ret}. Next, the \agent{SAgent} picks a smoothing family for each channel, and the \host{} marks the stable and deviant regions of each smoothed signal. The \agent{CSAgent} then removes channels that carry no useful information and gives each remaining channel $c$ a priority tier, $\mathrm{tier}(c)\in\{\texttt{high},\texttt{medium},\texttt{low}\}$; the \host{} checks this choice and may restore a removed channel. The final tiers give three sets of channels: $\mathcal P_1=\{c:\mathrm{tier}(c)=\texttt{high}\}$, $\mathcal P_2=\{c:\mathrm{tier}(c)=\texttt{medium}\}$, and $\mathcal P_3=\{c:\mathrm{tier}(c)=\texttt{low}\}$. In real sensor data, short peaks and irregular changes can come from either a real fault or a temporary change in operating or environmental conditions. If we treat all of these changes in the same way, we may confuse fault-related patterns with short-term noise or disturbances. To mitigate this ambiguity, we analyze the query from two complementary perspectives: the \agent{NAAgent} compares the query's stable region with the stable region of each reference, whereas the \agent{ABAgent} compares its deviant region with the corresponding deviant region of each reference. Each \agent{Analyst} report identifies a set of plausible classes and assigns each candidate a confidence level from $\{\texttt{high}, \texttt{medium}, \texttt{low}\}$, together with supporting and contradicting observations.

\subsection{CCGA: Host-Governed Adjudication}
\label{sec:adjudication}
\label{sec:evidence-contract}

Once both \agent{Analyst} reports are complete and frozen, the \host{} advances the case to the \agent{Judge}, providing the frozen reports together with the selected signal views, channel priority tiers, and, in standard mode, fitted-model predictions \revr{$b_1,b_2$}. If ambiguity remains, the \agent{Judge} may request the preceding \newred{sample} $x^{\mathrm{prev}}$. The \agent{LBAgents} then repeat the \agent{NAAgent} and \agent{ABAgent} comparisons on this \newred{sample} against the same references. Once these comparisons are complete, the \host{} aggregates their results into an overall report for the \agent{Judge} and constructs the initial candidate set $\mathcal H_1$ as the union of the plausible classes identified in all \agent{Analyst} reports (including lookback, if performed) and the two fitted-model \newred{predictions} $b_1$ and $b_2$. These complementary sources help the \agent{Judge} identify what remains unresolved and what additional evidence is needed. The subsequent investigation is governed by \protocol{} through three \newred{\host{} checks}: the request gate controls which measurements may run (Section~\ref{sec:request-gate}); the packet and influence gate controls what each result may support (Section~\ref{par:evidence-admission}); and finalization determines the reported diagnosis (Section~\ref{sec:finalization}).

\subsubsection{Request Gate}
\label{sec:request-gate}

This stage structures each round of the \agent{Judge}'s investigation around an unresolved property, proceeding through four steps: \textsc{Declare}, \textsc{Offer}, \textsc{Commit}, and \textsc{Execute}. Rounds are indexed by $t=1,2,\dots$; a subscript $t$ marks a quantity of round $t$, and the ledger $\mathcal E_t$ records every result returned before round $t$. 

\textbf{\textit{(1) Declare.}} The \agent{Judge} receives the \revr{live candidate set $\mathcal H_t$} (initially $\mathcal H_1$), the evidence ledger $\mathcal E_t$, and the registered set $\Phi_{\mathrm{reg}}$ of 48 signal properties (Table~\ref{tab:request-gate-property-registry} in Appendix~\ref{app:why-request-gate}). Based on the evidence already available for $x^\star$, the \agent{Judge} identifies the unresolved signal properties that could help distinguish among the candidate classes and declares a nonempty subset $\Phi_t \subseteq \Phi_{\mathrm{reg}}$ for further investigation.

\textbf{\textit{(2) Offer.}} The \host{} takes the requested properties $\Phi_t$, searches the registry of 25 default tools (Appendix~\ref{app:tool-registries}), and returns the set $\mathcal S_t$ of tools. Each tool is registered as a \emph{ToolCard} $\tau$, a fixed description specifying the set $\Phi(\tau)\subseteq\Phi_{\mathrm{reg}}$ of properties it measures, its input requirements, the measurements and outputs it provides, and its known limitations. This helps the \agent{Judge} determine which tools are suitable for investigating the unresolved signal properties.

\textbf{\textit{(3) Commit.}} The \agent{Judge} selects suitable tools from the offered set $\mathcal S_t$ and commits to them as a single batch $\mathcal T_t$ of ToolCards, subject to three conditions:
\begin{equation}
\mathcal T_t\subseteq\mathcal S_t,
\qquad
\Phi_t\subseteq\bigcup_{\tau\in\mathcal T_t}\Phi(\tau),
\qquad
\Phi_t\cap\Phi(\mathcal E_t)=\varnothing.
\label{eq:request-property-coverage}
\end{equation}
Together, these conditions ensure that the \agent{Judge} uses only offered tools, covers all declared properties, and does not revisit properties measured in earlier rounds. This prevents selective investigation and repeated requests for more favorable results. The \host{} checks all three conditions before execution. If any condition fails, the \agent{Judge} must select again.

\textbf{\textit{(4) Execute.}} The \host{} runs each ToolCard $\tau\in\mathcal T_t$ on the channels in $\mathcal P_1$, $\mathcal P_2$, and $\mathcal P_3$. On each channel, the tool compares $x^\star$ with the reference of every class in $\mathcal R^\star$ and ranks the classes. To combine the channels, the \host{} uses only the highest nonempty tier, $\mathcal P^\star=\mathcal P_k$ with $k=\min\{\ell:\mathcal P_\ell\neq\varnothing\}$. The tool's \emph{winning class} $\hat y_\tau(x^\star;\mathcal R^\star)$ is the class that every channel in $\mathcal P^\star$ ranks first; if these channels disagree, there is no winning class ($\hat y_\tau=\bot$). The \host{} returns the result as a \emph{packet} $p_\tau$, which holds the per-channel rankings, the winning class, and the tool's stated limitations. If the tool fails or there is no winning class, the packet is \emph{excluded}: it stays in the ledger but supports no class. Any other packet is \emph{usable}. 

Taken together, the four steps order the reasoning without limiting what may be investigated. Before accessing new evidence, the \agent{Judge} must name what remains unresolved as measurable signal properties and cite the evidence behind the request, so the reason for each measurement is recorded and can be checked. A complete agent-\host{} exchange is given in Appendix~\ref{app:workflow-request-gate}.

\FloatBarrier
\AddToHookNext{shipout/after}{\flushbottom}
\raggedbottom





\subsubsection{Packet and Influence Gate: Controlling What Each Result Can Support}
\label{par:evidence-admission}

A successfully executed measurement is not automatically reliable diagnostic evidence. Its result may be ambiguous or poorly supported, yet an unconstrained agent could still overuse it. The packet and influence gate therefore controls how much each returned result is allowed to influence the diagnosis, through a single check on the tool itself: a reference preflight. 

The winning class $\hat y_\tau(x^\star;\mathcal R^\star)$ shows which class one measurement favors. But when several tools name different classes, the \agent{Judge} still needs a basis to decide which results are reliable enough to influence the diagnosis. We therefore add a reference preflight before assigning any influence role. For each class $y$, the \host{} treats its retrieved reference $r_y^{\mathrm{ctrl}}=r_y$ (Section~\ref{sec:problem}) as a pseudo-query\newred{, called a \emph{control},} whose label it knows. The alternate bank is $\mathcal R^{\mathrm{alt}}(r_y^{\mathrm{ctrl}})=\mathcal R^\star(r_y^{\mathrm{ctrl}})$, which by Eq.~\ref{eq:ret} never contains \newred{$r_y^{\mathrm{ctrl}}$} itself. The \host{} runs the same ToolCard $\tau$ on each control with the same \newred{rule for the winning class}. The tool passes the preflight only if it recovers the known class of every control:
\begin{equation}
g_{\mathrm{ref}}(\tau)=
\begin{cases}
1, & \text{if } \hat y_\tau\bigl(r_y^{\mathrm{ctrl}};\,\mathcal R^{\mathrm{alt}}(r_y^{\mathrm{ctrl}})\bigr)=y \text{ for every } y\in\mathcal Y,\\
0, & \text{otherwise.}
\end{cases}
\label{eq:reference-sanity}
\end{equation}

\revr{Each} non-excluded packet then receives its \emph{influence role}:
\begin{equation}
\mathrm{role}(p_\tau)=
\begin{cases}
\text{corroborative}, & g_{\mathrm{ref}}(\tau)=1,\\
\text{advisory}, & g_{\mathrm{ref}}(\tau)=0.
\end{cases}
\label{eq:packet-role}
\end{equation}
\revred{The roles specify how a packet may be used, not how confident it is. A \emph{corroborative} packet passed the preflight, so its class counts as checked evidence and can support a proposed class. An \emph{advisory} packet failed it, so its class is unchecked: it may shape the \agent{Judge}'s hypothesis and its next request, and may add support to a proposal, but should not by itself overturn another result. This holds even when the packet's class on $x^\star$ is clear, because the same tool failed to recover a known control label. These roles fix how each packet is treated when a class is proposed (Section~\ref{sec:finalization}).} The \host{} then adds every packet of the round, with its role if it is usable, to the ledger, $\mathcal E_{t+1}=\mathcal E_t\cup\{p_\tau\}_{\tau\in\mathcal T_t}$. This closes the declared properties, $\Phi_t\subseteq\Phi(\mathcal E_{t+1})$, so later rounds cannot declare them again (Eq.~\ref{eq:request-property-coverage}).

\revred{Once every packet of the round has its role, the \agent{Judge} files one round report covering all of them: it states whether their results agree, conflict, or cannot be compared, names what remains uncertain, and decides whether to stop or open another round. The \host{} rejects a decision to stop while valid results still conflict and tool budget remains. If the \agent{Judge} continues, the next round begins at Declare; otherwise the case moves to finalization.}

\subsubsection{Finalization}
\label{sec:finalization}

The final stage decides which class is reported and whether it was admitted. When the \agent{Judge} is ready to conclude, it holds the final ledger $\mathcal E$: every executed packet together with its influence role. It proposes a class from this \newred{ledger}, weighing each packet by its role: a corroborative packet's class counts as checked support, whereas an advisory packet's class may support the proposal but should not by itself overturn another result. The \host{} then admits the proposal only when it \newred{meets two admission conditions, one on the candidate set and one on the measurements}: the proposed class must lie in $\mathcal H_1$\revred{, the candidate set before any measurement} \newred{(Section~\ref{sec:adjudication})}, and at least one round of quantitative measurement must have been completed (Appendix~\ref{app:workflow-finalization}). A proposal that fails either \newred{condition} is returned to the \agent{Judge} with the failed \newred{condition}, and the \agent{Judge} may revise it or measure further, within a budget of $N_J$ \newred{\agent{Judge}} steps (Table~\ref{tab:agent-config} in Appendix~\ref{app:experimental-protocol-details}). At the last step, the \agent{Judge} must propose a class in $\mathcal H_1$. The first proposal that \newred{meets both conditions}, denoted $\tilde y$, is admitted and reported, $\hat y=\tilde y$; if none does, the \host{} reports the last-step class and marks it as not admitted. This flag tells the user that the class did not pass the protocol checks and should be reviewed; it is not an estimate of diagnostic accuracy.


\subsubsection{Roles of the Judge and Host in CCGA}
\label{par:ccga-responsibilities}

The preceding sections describe how CCGA governs evidence acquisition, its influence on diagnosis, and finalization. Together, these mechanisms establish a clear division of responsibilities between the \agent{Judge} and the \host{}. The \agent{Judge} leads the investigation by identifying unresolved signal properties, selecting suitable ToolCards, interpreting their results, and proposing a diagnostic class. The \host{} oversees this process without making the diagnostic judgment itself. It checks whether tool requests satisfy the request conditions (Eq.~\ref{eq:request-property-coverage}), determines which returned packets are usable, and assigns their influence roles through the reference preflight. These roles guide how the \agent{Judge} uses the evidence: corroborative packets provide checked support, while advisory packets can guide further investigation and add support to a proposal. The \host{} also prevents premature stopping when valid results still conflict and tool budget remains, and checks the proposed class against the admission conditions (Section~\ref{sec:finalization}). In this way, the \agent{Judge} retains the freedom to investigate and reason about the current case, while the \host{} ensures that its decisions follow the evidence-handling constraints of CCGA.

\section{Experiments and Results}
\label{sec:results}

\subsection{Experimental Protocol}
\label{sec:exp-settings}
\paragraph{Datasets experimental settings.}
We evaluate three-class bearing diagnosis on the PU~\citep{Lessmeier2016PU}, UO~\citep{HuangBaddour2018UOBearing}, four-class induction-motor diagnosis on MCC5~\citep{chen2026motor}, and a proprietary customer dataset, Bel. All signals are segmented into non-overlapping windows of $T{=}2048$ observations, with 50 training samples retained per class. 
For the public benchmarks, launching the agents on the full datasets is infeasible. Therefore, we implement two subsampling techniques: Geometry-Stratified Response-Stability (GRS) for diversity and Baseline-Failure Hard (BFH) for selecting difficult samples. 
First, GRS uses $k-$means to cluster the held-out pool of each class and select only one representative window of the whole evaluation window where the baseline predictions are consistent.
This produces 906, 763, and 1152 evaluation windows for PU, UO, and MCC5, respectively. 
Second, for each dataset, BFH selects 100 approximately class-balanced windows from the held-out samples misclassified by a large fraction of the frozen baseline registry, yielding a hard test suite where the majority of the conventional baseline fail.
We provide the more details of GRS and BFH in Appendix~\ref{sec:grs-challenge-set} and Appendix~\ref{sec:bfh-challenge-set}) respectively. Bel is a proprietary dataset of raw sensor time series recorded by a bearing noise and vibration tester during quality inspection on our customer's production line. We evaluate all 330 test samples without BFH or GRS selection, preserving the customer-defined test distribution. Together, these evaluations assess performance from three complementary perspectives: coverage across observed regimes, robustness to baseline-failure cases, and generalization to an external customer dataset. Further details on the datasets and evaluation settings are provided in Appendix~\ref{app:datasets}.

\paragraph{Model settings and evaluation metrics.}
All agentic methods use Qwen3.6-27B~\citep{qwen36_27b} with greedy decoding under the same labeled-reference budget and resource limits. The main \method{} configuration uses frozen Moirai~1.1~R~Large~\citep{woo2024moirai} embeddings for class-balanced reference retrieval and for fitting the SVM and XGBoost heads supplied to the standard variant. Detailed inference, retrieval, and baseline configurations are provided in Appendices~\ref{app:experimental-protocol-details} and~\ref{app:agentic-baseline-boundaries}. Performance is measured using multiclass accuracy and macro F1 (F1-M).
We compare three families of methods. First, deep-learning baselines include CNN-TCC~\citep{eldele2021tstcc}, QCNN~\citep{QCNN2022}, and WDCNN~\citep{zhang2017wdcnn}. Second, foundation-model baselines pair frozen representations from Moirai~1.1~R~Large~\citep{woo2024moirai}, Moirai~2~Small~\citep{liu2026moirai20timeseries}, and Chronos~2 and Chronos~2~Small~\citep{ansari2025chronos2} with KNN, SVM, and XGBoost classification heads. Third, agentic methods include FETA~\citep{sui2025feta}, ZARA~\citep{li2026zara}, and our proposed \method{}.



\begin{figure*}[h]
\centering
\includegraphics[width=0.99\textwidth]{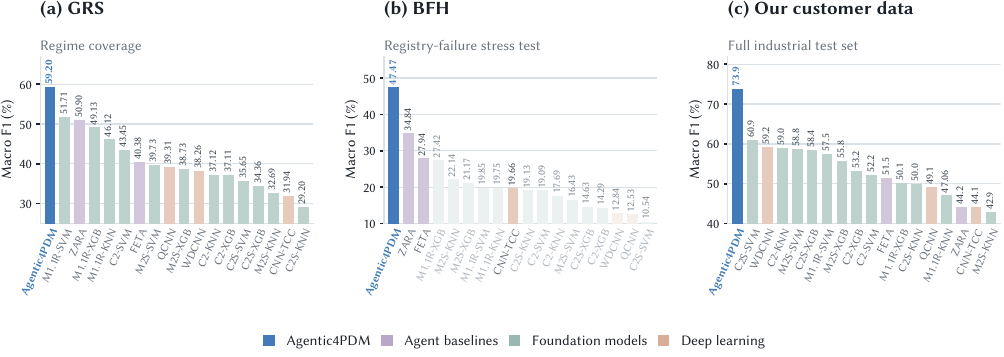}
\caption{Performance comparison between \method{} and baselines from three model families agentic methods, time-series foundation models, and deep-learning models across three evaluation settings: (a) GRS, (b) BFH, and (c) the customer \newred{test set (Bel)}. M1.1R, M2S, C2, and C2S denote Moirai~1.1~R~Large, Moirai~2~Small, Chronos~2, and Chronos~2~Small, respectively; XGB denotes XGBoost.}
\label{fig:challenge-results-avg}
\label{fig:bel-20window}
\end{figure*}

\subsection{Main Results}
\label{sec:main-results}
Figure~\ref{fig:challenge-results-avg} reports the performance of the three baseline families across the three benchmarks considered. 
Overall, \method{} consistently achieves the highest average performance across all three tests. We first examine GRS, where \method{} outperforms the strongest baseline by $+7.5\%$. Interestingly, the agentic baselines ZARA and FETA even fall behind the strongest time-series foundation models, suggesting that agentic reasoning alone does not necessarily yield a stronger diagnosis. The advantage of \method{} becomes more pronounced in BFH, which focuses on samples misclassified by baseline models. In this hard setting, \method{} surpasses ZARA by $+12\%$ and FETA by $+19.53\%$, demonstrating a substantially stronger ability to recover difficult cases. More importantly, we extend the evaluation to the full test set of a real customer dataset, BEL, narrowing the gap between benchmark evaluation and practical industrial deployment. Here, \method{} again ranks first, exceeding the strongest baseline by more than $+12\%$. Notably, the agentic baselines degrade substantially in this setting: ZARA ranks close to the lowest-performing methods, while FETA performs better than ZARA but still remains far behind \method{}.

Overall, across diverse benchmark conditions, difficult baseline-failure cases, and real industrial data, \method{} achieves the highest average performance among all evaluated methods. Additionally, Appendix~\ref{app:run-variability} examines run-to-run variation over three repeated executions. Despite the multi-round reasoning process of \method{}, the observed variation remains small: the largest standard deviation across datasets and metrics is approximately $2.3$ points on BFH and $1.2$ points on GRS. More importantly, these fluctuations are substantially smaller than the performance margins over the agentic baselines, showing that the reported gains are not simply due to a favorable individual run. For transparency, we provide detailed performance results in Appendix~\ref{app:performance_bfh_and_grs}.

\subsection{Agent Behaviour Analysis}
\label{sec:comprehensive-analysis}

Beyond aggregate performance, we ask a broader question: does Agentic4PDM exhibit sample-specific diagnostic reasoning, or can its behavior be explained by simpler fixed evidence strategies? We examine this question from three complementary perspectives.

\begin{figure}[h]
\centering
\includegraphics[width=0.7\linewidth]{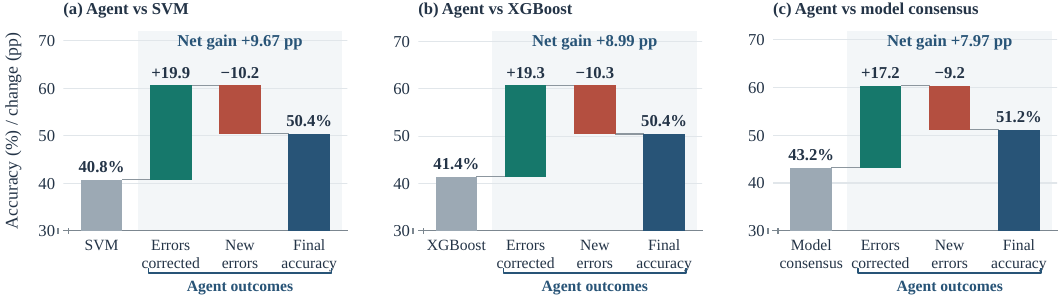}
\caption{Prediction revisions relative to fitted-model predictions on PU and MCC5 GRS. The consensus panel includes only cases where SVM and XGBoost initially predict the same class.}
\label{fig:accuracy-bridge}
\end{figure}

\paragraph{a) Does the agent add diagnostic value beyond fitted-model predictions?}
\label{par:agent-baseline-gains}
We further decompose the performance changes between \method{} and the fitted-model predictions of SVM and XGBoost baselines to determine whether the agent merely follows their predictions or can meaningfully revise them. Figure~\ref{fig:accuracy-bridge} separates three outcomes: \emph{errors corrected}, where an incorrect fitted-model prediction becomes correct after agent reasoning; \emph{new errors}, where a correct fitted-model prediction becomes incorrect; and \emph{final accuracy}, averaged over PU and MCC5 under GRS. Relative to SVM, \method{} corrects $19.9$ percentage points of errors while introducing $10.2$ points of new errors, yielding a net gain of $9.67$ points; against XGBoost, the corresponding values are $+19.3$, $-10.3$, and $+8.99$ points. More importantly, the same pattern persists when SVM and XGBoost initially agree. Even in these cases, \method{} corrects $17.2$ percentage points of their shared errors while introducing $9.2$ points of new errors, producing a net gain of $7.97$ points. This consensus result provides strong evidence that \method{} does not merely reproduce or arbitrate between fitted-model outputs; instead, it treats these predictions as diagnostic hypotheses and extends the analysis by acquiring additional evidence through its tools.

\paragraph{b) Can simpler combinations of the available evidence reproduce the workflow?}
We further examine whether the performance of \method{} can be explained by simpler ways of using its available evidence. First, Appendix~\ref{app:evidence-ensemble-comparison} combines the evidence available to the \agent{Judge} before its subsequent investigation (the \agent{NAAgent} and \agent{ABAgent} outputs, the final \agent{LBAgents} conclusion, and the SVM and XGBoost predictions) using flat plurality. \method{} still achieves higher macro-F1 on the same covered samples, by 5.83 points on MCC5 and 9.45 points on PU, showing that its gains are not reproduced by simply voting over the labels already present in its context. Second, Appendix~\ref{app:single-tool-transfer} asks whether one strong quantitative tool could replace the adaptive investigation. The best tool differs across datasets and transfers poorly: for example, Crossing-run Statistics is the strongest single tool on PU at 37.94\% macro-F1, but drops to 22.89\% on MCC5 and 16.73\% on UO, decreases of 15.05 and 21.21 points, respectively. Moreover, \method{} workflow outperforms the best single tool identified for each dataset. Together, these results show that neither fixed aggregation of the initial evidence nor reliance on a single strong measurement reproduces the full workflow, and are consistent with the value of selecting and combining measurements according to the diagnostic situation of the current case.

\paragraph{c) Does the investigation adapt to the current case?}
\label{par:sample-adaptivity}

\begin{figure}[h]
\centering
\includegraphics[width=0.5\linewidth]{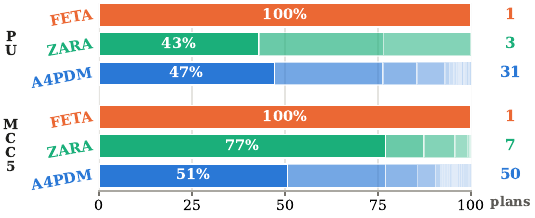}
\caption{ Observed plans across agentic methods on PU and MCC5. Bar segments show the frequency of each plan, the right number indicates the total number of distinct plans.}
\label{fig:sample-adaptivity}
\end{figure}
In this section, we examine whether each method changes what it inspects from one sample to another (Figure~\ref{fig:sample-adaptivity}). We define a \emph{plan} as the combination of analysis choices used for a sample before the final decision: the channels inspected by FETA, the features selected by ZARA, or the signal properties requested by \method{}. The three methods exhibit markedly different behaviors. FETA applies a single fixed plan to every sample on both datasets. ZARA shows limited variation, using only 3 plans on PU and 7 on MCC5; moreover, its most frequent plan covers 43\% and 77\% of the samples, respectively. This variation is largely tied to which class pair remains after pruning, so the selected plan is primarily determined by the surviving candidate pair rather than by a new investigation of each individual sample. In contrast, \method{} produces 31 distinct plans on PU and 50 on MCC5, while its most frequent plan covers only 47\% and 51\% of samples. This flexibility allows \method{} to redirect its investigation toward the signal properties that are most relevant to the ambiguity of the current case, rather than repeatedly applying the same plan. Thus, nearly half of the samples follow a plan different from the dominant one. Overall, the results show a clear progression from fixed analysis in FETA, to candidate-pair-dependent analysis in ZARA, to substantially more sample-specific investigation in \method{}. While this analysis does not by itself establish that plan diversity causes the performance gains, it shows that \method{} has the capacity to adapt its evidence acquisition to the diagnostic situation presented by each sample.

\subsection{Ablation Study}
\label{sec:planned-component-ablation}

\begin{table}[h]
\centering
\caption{Ablation of the main components in \method{} on the PU data under GRS sampling.}
\label{tab:core-ablation-plan}
\scriptsize
\setlength{\tabcolsep}{2pt}
\resizebox{0.7\linewidth}{!}{%
\begin{tabular}{@{}lccccc|rr@{}}
\toprule
& \multicolumn{5}{c}{Enabled mechanisms}
& \multicolumn{2}{c}{Performance (\%)} \\
\cmidrule(lr){2-6}\cmidrule(lr){7-8}
Variant
& \makecell{CB\\retrieval}
& \makecell{\agent{Analyst}\\reports}
& \makecell{Request\\gate}
& \makecell{Packet \newred{\&}\\\newred{influence}\\gate}
& \makecell{Finali-\\zation}
& \makecell{ACC $\uparrow$}
& \makecell{\newred{F1-M} $\uparrow$} \\
\midrule
\textbf{A1 Full workflow}
& \bluetick & \bluetick & \bluetick & \bluetick & \bluetick
& 47.24 & 38.54 \\
\specialrule{0.25pt}{2pt}{2pt}
A2 Global top-$K$
& \redcross & \bluetick & \bluetick & \bluetick & \bluetick
& 37.20 & 29.89 \\
\specialrule{0.25pt}{2pt}{2pt}
A3 No \agent{Analyst} reports
& \bluetick & \redcross & \bluetick & \bluetick & \bluetick
& 46.35 & 37.72 \\
\specialrule{0.25pt}{2pt}{2pt}
A4 Ungoverned
& \bluetick & \bluetick & \redcross & \redcross & \bluetick
& 36.31 & 30.53 \\
A5 No request gate
& \bluetick & \bluetick & \redcross & \bluetick & \bluetick
& 39.96 & 33.35 \\
\makecell[l]{A6 No packet \newred{\&}\\influence gate}
& \bluetick & \bluetick & \bluetick & \redcross & \bluetick
& 42.60 & 36.70 \\
\specialrule{0.25pt}{2pt}{2pt}
A7 No finalization
& \bluetick & \bluetick & \bluetick & \bluetick & \redcross
& 45.40 & 38.02 \\
\bottomrule
\end{tabular}%
}
\vspace{3pt}
\begin{minipage}{\linewidth}
\scriptsize
\end{minipage}
\end{table}

In this section, we perform a comprehensive ablation analysis of the main components of \method{}. To avoid test-specific design bias and ensure a controlled comparison, all variants are evaluated under the same PU GRS setting, as reported in Table~\ref{tab:core-ablation-plan}.

\paragraph{Preserving candidate hypotheses is critical.}
Class-balanced retrieval and the \agent{Analyst} reports both keep plausible classes in view before the \agent{Judge} investigates. Replacing class-balanced retrieval with global top-$K$ retrieval (A2) causes a large drops (accuracy $-10.04$ and macro-F1 $-8.65$ points): several retrieved references may share one class, leaving other classes with nothing to be compared with. Removing the \agent{Analyst} reports (A3) has a smaller effect (accuracy $-0.89$ and macro-F1 $-0.82$ points), as the \agent{Judge} loses the candidate classes, observations, and contradictions raised by the \agent{NAAgent}, \agent{ABAgent}, and \agent{LBAgents}. Retrieval thus keeps every class at the reference level, and the reports keep plausible classes at the reasoning level. The drop in A3 is likely smaller because, in the standard setting, fitted-model predictions and quantitative measurements partly replace the reports, whereas a class without a reference cannot be compared at all.

\paragraph{Governance has the largest effect.}
We next remove the controls of \protocol{} while keeping the prompts, tool registry, references, and \host{} limits of the full workflow unchanged. Removing both gates (A4) causes the largest drop of all variants: accuracy falls by 10.93 points and macro-F1 by 8.01 points. Each gate also contributes on its own. Without the request gate (A5), the \agent{Judge} no longer has to state which unresolved property a tool should measure before using it; accuracy falls by 7.28 points and macro-F1 by 5.19 points. Without the packet and influence gate (A6), the \agent{Judge} no longer knows which of its selected tools are reliable for the current case. When several tools return different classes, it has no basis for deciding which results should matter most and may simply follow the class that most tools agree on; accuracy falls by 4.64 points and macro-F1 by 1.84 points. Appendix~\ref{app:gate-saved-log-audit} further examines the influence roles assigned by the reference preflight in the packet and influence gate. Corroborative packets appear less frequently than advisory packets, but provide more accurate guidanceon most datasets. This supports the purpose of the packet and influence gate: the preflight does not merely partition tool outputs, but identifies a smaller subset that can provide more reliable guidance in the settings where the distinction is most pronounced. Removing both gates hurts more than removing either one, indicating that they play complementary roles: the request gate ties each measurement to a stated diagnostic question, while the packet and influence gate determines how much each returned result is permitted to influence the diagnosis.

\paragraph{Finalization controls unreliable Judge proposals.}
In A7 (\emph{No Finalization}), we examine if the \agent{Judge} can reliably determine the final diagnosis without the admission conditions defined in Section~\ref{sec:finalization}. We directly take the \agent{Judge}'s decisions from the A1 agent conversation traces, thereby avoiding additional run-to-run variability. Removing finalization reduces accuracy by 1.84 percentage points and macro-F1 by 0.52 points. These results suggest that finalization provides a meaningful safeguard against unreliable \agent{Judge} proposals, preventing decisions that lack sufficient supporting evidence.

\subsection{Computational Cost Analysis}


\begin{figure}[!ht]
\centering
\includegraphics[width=0.68\columnwidth]{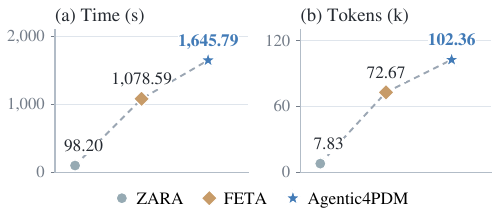}
\caption{Mean recorded time and total tokens per query for ZARA, FETA, and standard \method{}. Here, k denotes 1,000 tokens. Marker shapes identify methods; dashed lines guide the eye}
\label{fig:resource-time-total}
\Description{Two side-by-side point charts show recorded time and total tokens per query. A shared legend identifies ZARA with circles, FETA with diamonds, and Agentic4PDM with stars. Gray dashed lines connect the method summaries in each panel. Times are 98.20, 1078.59, and 1645.79 seconds; token counts are 7.83, 72.67, and 102.36 thousand, respectively.}
\end{figure}
In Figure~\ref{fig:resource-time-total} compares recorded processing time and total token use per query. Mean token usage is 102.36k for \method{}, 72.67k for FETA, and 7.83k for ZARA, where k denotes one thousand tokens. Their estimated API charges are \$0.0989, \$0.0575, and \$0.0129 per query. \method{} is meant for on-demand diagnosis of difficult cases, not for real-time monitoring; its time per sample is reported in Appendix~\ref{app:resource-complexity}. Relative to FETA, the additional estimated API charge is \$0.0414 per query, accompanied by macro-F1 gains of 18.82 points on GRS, 19.53 points on BFH, and 22.47 points on the full Bel test set. ZARA provides a substantially cheaper option, but has lower macro-F1 in all three settings; on Bel, \method{} reaches 73.95\% versus 44.17\% for ZARA. These results suggest that the additional computation may be justified for on-demand diagnosis of complex and high-value industrial equipment. Appendix~\ref{app:resource-complexity} details recorded runtime, token usage, LLM calls, and the pricing assumptions.

\section{Conclusion}
\label{sec:conclusion}

We introduce \method{}, an agentic framework for industrial fault diagnosis under label scarcity and potentially shifting operating conditions. Although raw signal statistics may change, diagnostically meaningful structures such as periodicity, trends, and transients can remain informative and be measured by analytical tools. \method{} guides LLM agents using these tools through contract-checked gated adjudication (\protocol{}). The agent identifies ambiguity among candidate classes and selects measurements that could resolve it, while \protocol{} controls each result's permitted influence through reference checks rather than the LLM's confidence. This combines investigation tailored to the current case with external control over evidence use, giving unchecked outputs less evidential authority than corroborative results. Across two bearing benchmarks, an induction-motor benchmark, and real-world customer data, \method{} achieves the strongest aggregate diagnostic performance among the evaluated deep-learning, time-series foundation-model, and agentic baselines. Our analyses and ablations support the value of the governed investigation and the controls of \protocol{}.


These results suggest two directions for future research. First, consolidating tools while broadening their functionality could reduce the context needed to describe them, simplify tool selection, and support richer signal analysis. Second, \method{} records whether a diagnosis is admitted in $\mathcal A$, but does not yet recommend follow-up actions. Providing concrete suggestions for additional measurements or inspections could help operators investigate unresolved cases and improve the system's practical usefulness.


\FloatBarrier
\clearpage
\bibliographystyle{plainnat}
\bibliography{acml26}

\clearpage

\appendix
\fontsize{10}{12}\selectfont

\begin{center}
{}\textbf{\Large{Supplement to
``Agentic4PDM: Host-Governed Evidence Admissibility for
Industrial Fault Diagnosis''}}
\end{center}
This document provides the supplementary materials for the paper Agentic4PDM: Host-Governed Evidence Admissibility for Industrial Fault Diagnosis, and is organized as follows.

\raggedbottom
\begingroup
\hypersetup{linkcolor=black}
\renewcommand{\rmdefault}{toctnr}
\renewcommand{\sfdefault}{toctnr}
\renewcommand{\ttdefault}{toctnr}
\fontfamily{toctnr}\selectfont
\renewcommand{\textbf}[1]{{\fontsize{8.8}{10}\selectfont\bfseries #1}}
\noindent\textbf{\large Contents}\par
\vspace{6pt}
\providecommand{\agentrev}[1]{#1}
\renewcommand{\rev}[1]{#1}
\renewcommand{\revr}[1]{#1}
\renewcommand{\revred}[1]{#1}
\renewcommand{\revcolor}{}
\footnotesize
\setlength{\tabcolsep}{0pt}
\renewcommand{\arraystretch}{1.05}
\noindent\begin{tabular*}{\textwidth}{@{}>{\raggedright\arraybackslash}p{0.05\textwidth}>{\raggedright\arraybackslash}p{0.80\textwidth}@{\extracolsep{\fill}}r@{}}

\multicolumn{3}{@{}l}{\textit{System implementation}} \\[-2pt]
\textbf{\ref{app:context-memory}} & \textbf{\nameref{app:context-memory}} & \pageref{app:context-memory} \\
\textbf{\ref{app:information-isolation}} & \textbf{\nameref{app:information-isolation}} & \pageref{app:information-isolation} \\

\addlinespace[4pt]
\multicolumn{3}{@{}l}{\textit{Part I --- Results and analyses}} \\[-2pt]
\textbf{\ref{app:complete-challenge-results}} & \textbf{\nameref{app:complete-challenge-results}} & \pageref{app:complete-challenge-results} \\
 & \hspace{1.5em}\hyperref[app:performance_bfh_and_grs]{\ref*{app:performance_bfh_and_grs}\quad\nameref*{app:performance_bfh_and_grs}} & \pageref{app:performance_bfh_and_grs} \\
 & \hspace{1.5em}\hyperref[par:training-free-comparison]{\ref*{par:training-free-comparison}\quad\nameref*{par:training-free-comparison}} & \pageref{par:training-free-comparison} \\
 & \hspace{1.5em}\hyperref[app:run-variability]{\ref*{app:run-variability}\quad\nameref*{app:run-variability}} & \pageref{app:run-variability} \\
 & \hspace{1.5em}\hyperref[app:evidence-ensemble-comparison]{\ref*{app:evidence-ensemble-comparison}\quad\nameref*{app:evidence-ensemble-comparison}} & \pageref{app:evidence-ensemble-comparison} \\
 & \hspace{1.5em}\hyperref[app:gate-saved-log-audit]{\ref*{app:gate-saved-log-audit}\quad\nameref*{app:gate-saved-log-audit}} & \pageref{app:gate-saved-log-audit} \\
\textbf{\ref{app:single-tool-transfer}} & \textbf{\nameref{app:single-tool-transfer}} & \pageref{app:single-tool-transfer} \\
\textbf{\ref{app:backbone-substitution}} & \textbf{\nameref{app:backbone-substitution}} & \pageref{app:backbone-substitution} \\
\textbf{\ref{app:resource-complexity}} & \textbf{\nameref{app:resource-complexity}} & \pageref{app:resource-complexity} \\
\textbf{\ref{par:condition-accuracy}} & \textbf{\nameref{par:condition-accuracy}} & \pageref{par:condition-accuracy} \\

\addlinespace[4pt]
\multicolumn{3}{@{}l}{\textit{Part II --- Experimental setup and benchmarks}} \\[-2pt]
\textbf{\ref{app:experimental-setup}} & \textbf{\nameref{app:experimental-setup}} & \pageref{app:experimental-setup} \\
 & \hspace{1.5em}\hyperref[app:datasets]{\ref*{app:datasets}\quad\nameref*{app:datasets}} & \pageref{app:datasets} \\
 & \hspace{3em}\hyperref[app:benchmark-manifests]{\ref*{app:benchmark-manifests}\quad\nameref*{app:benchmark-manifests}} & \pageref{app:benchmark-manifests} \\
 & \hspace{1.5em}\hyperref[app:datasets-baselines]{\ref*{app:datasets-baselines}\quad\nameref*{app:datasets-baselines}} & \pageref{app:datasets-baselines} \\
 & \hspace{1.5em}\hyperref[app:experimental-protocol-details]{\ref*{app:experimental-protocol-details}\quad\nameref*{app:experimental-protocol-details}} & \pageref{app:experimental-protocol-details} \\
 & \hspace{1.5em}\hyperref[app:deep-baseline-settings]{\ref*{app:deep-baseline-settings}\quad\nameref*{app:deep-baseline-settings}} & \pageref{app:deep-baseline-settings} \\
 & \hspace{1.5em}\hyperref[app:foundation-baseline-settings]{\ref*{app:foundation-baseline-settings}\quad\nameref*{app:foundation-baseline-settings}} & \pageref{app:foundation-baseline-settings} \\
 & \hspace{1.5em}\hyperref[app:agentic-baseline-boundaries]{\ref*{app:agentic-baseline-boundaries}\quad\nameref*{app:agentic-baseline-boundaries}} & \pageref{app:agentic-baseline-boundaries} \\
\textbf{\ref{app:benchmark-provenance}} & \textbf{\nameref{app:benchmark-provenance}} & \pageref{app:benchmark-provenance} \\
 & \hspace{1.5em}\hyperref[app:benchmark-bfh]{\ref*{app:benchmark-bfh}\quad\nameref*{app:benchmark-bfh}} & \pageref{app:benchmark-bfh} \\
 & \hspace{1.5em}\hyperref[app:benchmark-grs]{\ref*{app:benchmark-grs}\quad\nameref*{app:benchmark-grs}} & \pageref{app:benchmark-grs} \\
 & \hspace{1.5em}\hyperref[app:benchmark-budget-rationale]{\ref*{app:benchmark-budget-rationale}\quad\nameref*{app:benchmark-budget-rationale}} & \pageref{app:benchmark-budget-rationale} \\

\addlinespace[4pt]
\multicolumn{3}{@{}l}{\textit{Part III --- Method: notation, protocol, tools, and prompts}} \\[-2pt]
\textbf{\ref{app:method-notation}} & \textbf{\nameref{app:method-notation}} & \pageref{app:method-notation} \\
\textbf{\ref{app:method-details}} & \textbf{\nameref{app:method-details}} & \pageref{app:method-details} \\
 & \hspace{1.5em}\hyperref[app:retrieval-details]{\ref*{app:retrieval-details}\quad\nameref*{app:retrieval-details}} & \pageref{app:retrieval-details} \\
 & \hspace{1.5em}\hyperref[app:host-request-gate]{\ref*{app:host-request-gate}\quad\nameref*{app:host-request-gate}} & \pageref{app:host-request-gate} \\
 & \hspace{1.5em}\hyperref[app:host-packet-gate]{\ref*{app:host-packet-gate}\quad\nameref*{app:host-packet-gate}} & \pageref{app:host-packet-gate} \\
 & \hspace{1.5em}\hyperref[app:finalization-details]{\ref*{app:finalization-details}\quad\nameref*{app:finalization-details}} & \pageref{app:finalization-details} \\
\textbf{\ref{app:tool-registries}} & \textbf{\nameref{app:tool-registries}} & \pageref{app:tool-registries} \\
 & \hspace{1.5em}\hyperref[app:general-tools]{\ref*{app:general-tools}\quad\nameref*{app:general-tools}} & \pageref{app:general-tools} \\
 & \hspace{1.5em}\hyperref[app:smoothing-tools]{\ref*{app:smoothing-tools}\quad\nameref*{app:smoothing-tools}} & \pageref{app:smoothing-tools} \\
 & \hspace{1.5em}\hyperref[app:channel-selection]{\ref*{app:channel-selection}\quad\nameref*{app:channel-selection}} & \pageref{app:channel-selection} \\
 & \hspace{1.5em}\hyperref[app:selectable-quantitative-tools]{\ref*{app:selectable-quantitative-tools}\quad\nameref*{app:selectable-quantitative-tools}} & \pageref{app:selectable-quantitative-tools} \\

 & \hspace{1.5em}\hyperref[app:quantitative-tool-examples]{\ref*{app:quantitative-tool-examples}\quad\nameref*{app:quantitative-tool-examples}} & \pageref{app:quantitative-tool-examples} \\

\textbf{\ref{app:agents-prompts}} & \textbf{\nameref{app:agents-prompts}} & \pageref{app:agents-prompts} \\
 & \hspace{1.5em}\hyperref[app:agents-evidence-prep]{\ref*{app:agents-evidence-prep}\quad\nameref*{app:agents-evidence-prep}} & \pageref{app:agents-evidence-prep} \\
 & \hspace{1.5em}\hyperref[app:agents-comparators]{\ref*{app:agents-comparators}\quad\nameref*{app:agents-comparators}} & \pageref{app:agents-comparators} \\
 & \hspace{1.5em}\hyperref[app:agents-judge-host]{\ref*{app:agents-judge-host}\quad\nameref*{app:agents-judge-host}} & \pageref{app:agents-judge-host} \\

\addlinespace[4pt]
\multicolumn{3}{@{}l}{\textit{Part IV --- Worked examples}} \\[-2pt]
\textbf{\ref{app:agent-workflow-trace}} & \textbf{\nameref{app:agent-workflow-trace}} & \pageref{app:agent-workflow-trace} \\
\textbf{\ref{app:agent-dialogue-side-by-side}} & \textbf{\nameref{app:agent-dialogue-side-by-side}} & \pageref{app:agent-dialogue-side-by-side} \\
\end{tabular*}
\endgroup
\clearpage

\section{Context Management for Multi-Turn Diagnosis}
\label{app:context-memory}

During development, repeated image inputs and an expanding conversation history caused the \agent{Judge}'s context to reach the model's token limit. Both sources accumulate as an investigation proceeds: new measurements introduce additional images, while subsequent turns add interpretations and decisions. We therefore manage image inputs and conversation history separately, retaining recent material in the active context and compact records of earlier material.

\begin{enumerate}[leftmargin=*,itemsep=4pt,topsep=4pt]
\item \textbf{Image retention.} We retain $n_{\mathrm{img}}$ images in the \agent{Judge}'s active context and allow an extension of up to $n_{\mathrm{ext}}$ additional images from the most recent image history. Earlier images are stored as external artifacts rather than repeatedly included as image inputs. Each stored image has a structured record containing its file path, a description of its diagnostic relevance, a timestamp, the originating agent, and the associated \newred{sample} identifier. These records preserve the link between an earlier observation and its source artifact while reducing repeated image input.

\item \textbf{Conversation summarization.} We retain the initial conversation turn and the most recent $n_{\mathrm{turn}}$ turns, replacing the intervening turns with a structured summary generated by the \agent{Judge}. The summary captures the key observations and decisions from those turns. Retaining the opening turn preserves the initial task context, while retaining recent turns preserves the immediate reasoning that leads to the next action.
\end{enumerate}

\paragraph{Illustrative examples.}
The following examples use $n_{\mathrm{img}}=3$, $n_{\mathrm{ext}}=2$, and $n_{\mathrm{turn}}=2$ to make the retention rules concrete. These values, artifact paths, and diagnostic observations are illustrative; they are not reported experimental settings or results.

\noindent\textbf{Example 1: retaining images without repeating the full image history.}
Suppose an investigation has produced seven images, $I_1,\ldots,I_7$, in chronological order. The active context contains the three most recent images, $I_5,I_6,I_7$. An extension can include $I_3$ and $I_4$, increasing the image input to five rather than all seven. Earlier images remain stored with descriptive records. For example, the record for $I_2$ could be:
\begin{promptlisting}{Illustrative record for an archived image}
{
  "image_id": "I2",
  "path": "artifacts/window_123/waveform_overlay.png",
  "description": "The query has isolated peaks not visible in the normal reference; their cause remains unresolved.",
  "timestamp": "2026-09-30T10:02:00Z",
  "source": "(*@ABAgent@*)",
  "related_window": "window_123"
}
\end{promptlisting}
\par\smallskip
This record identifies the observation and its source without presenting the earlier waveform image again. The description also preserves the uncertainty in the observation rather than treating the peaks as a confirmed fault.

\noindent\textbf{Example 2: summarizing an extended conversation.}
Suppose the conversation contains eight turns, $T_1,\ldots,T_8$. With $n_{\mathrm{turn}}=2$, the active context retains the opening turn $T_1$ and the latest turns $T_7,T_8$, while turns $T_2$--$T_6$ are replaced by a summary. If those middle turns discussed the peaks in Example~1, a compact summary could be:
\begin{promptlisting}{Illustrative summary of turns 2--6}
{
  "related_window": "window_123",
  "summarized_turns": [2, 6],
  "observations": ["Isolated peaks occur in the query (image I2)."],
  "open_question": "Do the peaks reflect a fault or a transient disturbance?",
  "decision": "Inspect an additional measurement before choosing a class."
}
\end{promptlisting}
\par\smallskip
The next turn therefore has the initial task, the recent discussion, and a compact account of the earlier evidence and unresolved question. This makes explicit what the summary needs to preserve for the investigation to continue coherently.

Together, these mechanisms reduce the amount of historical material included in each request while maintaining a compact record of the investigation. Their usefulness depends on the image descriptions and conversation summaries retaining the information needed for subsequent diagnostic decisions.

\FloatBarrier
\Needspace{6\baselineskip}
\section{Preventing Label Leakage through Client--Server Data Isolation}
\label{app:information-isolation}

A diagnostic agent should infer the query's health status from the permitted evidence. Original dataset paths can undermine this requirement: a directory named after a fault class, or a filename containing a specimen's condition, can reveal the answer before the signal is inspected. We therefore separate the source-data namespace from the files supplied to the agents. The client reads the source file, while the server creates a new analysis copy with a neutral filename in a job-specific workspace.

\paragraph{From source paths to neutral analysis paths.}
The implementation applies this separation at three points:
\begin{enumerate}[leftmargin=*,itemsep=4pt,topsep=4pt]
\item \textbf{Client submission.} The evaluation client uploads the query under the fixed multipart filename \texttt{input.csv}. The original query filename and its parent directories are therefore not used as the uploaded file's identity. Dataset context and optional operating-condition information are supplied as separate fields so that schema resolution and reference selection do not require the original query path.

\item \textbf{Server staging.} The server first writes the upload to a temporary \texttt{input.csv}, then moves it to \texttt{inputs/sample.csv} inside the job workspace. By default, the workspace is named using a randomly generated UUID; a caller may also provide an artifact directory. This relocation replaces the original dataset hierarchy with the server's analysis namespace. Before inference, normalization drops the query's \texttt{label} column and retains only the configured time and signal columns.

\item \textbf{Reference materialization.} Retrieved reference signals are also written into the workspace, under paths such as \texttt{inputs/references/class\_0.csv}. The agent receives these aligned copies rather than the original training-file paths, which may contain additional specimen or fault metadata. The class identifier is retained because the reference is an intentionally labeled example; it does not identify the query's true class.
\end{enumerate}

\paragraph{Separating inference from scoring.}
The API accepts the query's true label as evaluation metadata. The server keeps it in the request record and uses it to compute correctness after the diagnostic workflow produces a prediction. During inference, the agents receive the sanitized query, labeled references, and dataset context; the query's true label is excluded from these inputs. The scoring process can therefore compare the prediction with the ground truth while keeping the evaluation label separate from the evidence used for diagnosis.

\paragraph{Illustrative path transformation.}
For a hypothetical query whose source directory reveals a fault category, the inference path becomes:
\begin{promptlisting}{Illustrative query-path transformation (not an experimental sample)}
Client source:   /data/bearing/fault_class_2/specimen_A/window_17.csv
Upload filename: input.csv
Server input:    /analysis/<job_uuid>/inputs/sample.csv
Inference CSV:   time_point + configured signal channels
Scoring record:  true_label = 2 (kept separately from the inference CSV)
\end{promptlisting}
\par\smallskip
The path supplied for signal analysis no longer contains the source fault category or specimen name. This mechanism is path pseudonymization and data relocation. Dataset-level context, labeled references, and baseline predictions remain explicit permitted inputs. The separation addresses accidental disclosure through source paths and query columns; its scope is the inference data flow, with evaluation metadata and diagnostic logs retained by the trusted \host{}. Filesystem access control and cryptographic confidentiality require their own enforcement.

\FloatBarrier
\Needspace{6\baselineskip}
\section{Extended Experimental Evaluation}

\label{app:complete-challenge-results}

\subsection{Performance on BFH, GRS, and Bel}
\label{app:performance_bfh_and_grs}
Table~\ref{tab:challenge-results-combined} reports per-dataset results for the standard configuration and baselines on the public BFH/GRS benchmarks, together with full-test-set Bel scores. The training-free comparison is summarized in the following subsection. Chronos-2~Small \method{} variants are reported separately in Appendix~\ref{app:backbone-substitution}.
\newcommand{\agentbaselineresultrows}{%
 & ZARA \citep{li2026zara}
& 45.00 & 39.30 & 39.00 & 40.31 & 35.00 & 24.90
& 63.43 & 65.26 & 41.94 & \boldnum{43.23} & 47.66 & 44.22
& 39.67 & 34.84 & 51.01 & 50.90  & 44.24 & 44.17 \\
\multirow{-2}{*}{Agent} & FETA \citep{sui2025feta}
& 34.00 & 32.32 & 36.00 & 33.84 & 17.00 & 17.65
& 53.73 & 53.46 & 35.21 & 30.79 & 38.02 & 36.90
& 29.00 & 27.94 & 42.32 & 40.38  & 52.12 & 51.48 \\
\midrule
\rowcolor{blue!8}
\cellcolor{white}\textbf{Ours} & \method{}
& \boldnum{62.00} & \boldnum{61.93} & \boldnum{44.00} & 36.88 & \boldnum{45.00} & \boldnum{43.60}
& \boldnum{86.37} & \boldnum{86.12} & \boldnum{47.24} & 38.54 & 52.95 & 52.93
& \boldnum{50.33} & \boldnum{47.47} & \boldnum{62.19} & \boldnum{59.20}  & \boldnum{74.55} & \boldnum{73.95} \\
}

\newlength{\challengemetricwidth}
\newlength{\challengecustomerwidth}
\newlength{\challengeavgcustomerwidth}
\newcommand{\challengeresultscombinedtable}{%
\begin{table}[H]
\caption{Performance comparison of Agentic4PDM against deep-learning, time-series foundation-model, and agentic baselines across three evaluation settings: BFH and GRS on the public benchmarks (UO, PU, and MCC5), and the full customer test set (Bel). ACC and F1-M denote accuracy and macro-F1, respectively; AVG is the unweighted mean across public datasets. Bold indicates the best results in each column. }
\label{tab:challenge-results-combined}
\label{tab:agent-multiclass-bfh}
\centering\scriptsize
\setlength{\tabcolsep}{2.1pt}
\renewcommand{\arraystretch}{1.12}
\settowidth{\challengemetricwidth}{\textbf{00.00}}
\setlength{\challengecustomerwidth}{\dimexpr2\challengemetricwidth+4\tabcolsep\relax}
\setlength{\challengeavgcustomerwidth}{\dimexpr6\challengemetricwidth+12\tabcolsep\relax}
\resizebox{\textwidth}{!}{%
\begin{tikzpicture}[baseline=(challengegrid.base)]
\node[inner sep=0pt,outer sep=0pt] (challengegrid) {%
\begin{tabular}{@{}ll*{18}{w{r}{\challengemetricwidth}}}
\toprule
\multirow{3}{*}[-1.5ex]{Group} & \multirow{3}{*}[-1.5ex]{Method} & \multicolumn{6}{c}{\textbf{\makecell{Baseline-Failure\\Hard (BFH)}}} & \multicolumn{6}{c}{\textbf{\makecell{Geometry-Stratified\\Response-Stability (GRS)}}} & \multicolumn{4}{c}{\textbf{\makecell{AVG challenge\\tests}}} & \multicolumn{2}{c}{\parbox[c]{\dimexpr2\challengemetricwidth\relax}{\centering\bfseries Our\\customer\\test}} \\
\cmidrule(lr){3-8}\cmidrule(lr){9-14}\cmidrule(lr){15-18}\cmidrule(lr){19-20}
& & \multicolumn{2}{c}{UO} & \multicolumn{2}{c}{PU} & \multicolumn{2}{c}{\newred{MCC5}} & \multicolumn{2}{c}{UO} & \multicolumn{2}{c}{PU} & \multicolumn{2}{c}{\newred{MCC5}} & \multicolumn{2}{c}{BFH} & \multicolumn{2}{c}{GRS} & \multicolumn{2}{c}{Bel} \\
\cmidrule(lr){3-4}\cmidrule(lr){5-6}\cmidrule(lr){7-8}\cmidrule(lr){9-10}
\cmidrule(lr){11-12}\cmidrule(lr){13-14}\cmidrule(lr){15-16}\cmidrule(lr){17-18}\cmidrule(lr){19-20}
& & \multicolumn{1}{c}{ACC} & \multicolumn{1}{c}{F1-M} & \multicolumn{1}{c}{ACC} & \multicolumn{1}{c}{F1-M} & \multicolumn{1}{c}{ACC} & \multicolumn{1}{c}{F1-M} & \multicolumn{1}{c}{ACC} & \multicolumn{1}{c}{F1-M} & \multicolumn{1}{c}{ACC} & \multicolumn{1}{c}{F1-M} & \multicolumn{1}{c}{ACC} & \multicolumn{1}{c}{F1-M} & \multicolumn{1}{c}{ACC} & \multicolumn{1}{c}{F1-M} & \multicolumn{1}{c}{ACC} & \multicolumn{1}{c}{F1-M} & \multicolumn{1}{c}{ACC} & \multicolumn{1}{c}{F1-M} \\
\midrule
 & CNN-TCC~\citep{eldele2021tstcc} & 21.00 & 15.26 & 22.00 & 19.99 & 28.00 & 23.74 & 27.13 & 27.06 & 23.73 & 19.86 & 50.35 & 48.89 & 23.67 & 19.66 & 33.74 & 31.94 & 46.97 & 44.08 \\
 & QCNN~\citep{QCNN2022} & 1.00 & 0.94 & 30.00 & 29.06 & 8.00 & 7.60 & 27.92 & 27.07 & 46.03 & 37.19 & 53.65 & \boldnum{53.68} & 13.00 & 12.53 & 42.53 & 39.31  & 53.94 & 49.07 \\
\multirow{-3}{*}{\makecell[l]{Deep\\learning}} & WDCNN~\citep{zhang2017wdcnn} & 4.00 & 3.73 & 26.00 & 24.04 & 10.00 & 10.75 & 26.87 & 25.53 & 44.26 & 35.65 & \boldnum{53.99} & 53.61 & 13.33 & 12.84 & 41.71 & 38.26  & 61.21 & 59.16 \\
\midrule
 & Moirai 1.1 R Large~KNN & 27.00 & 28.20 & 36.00 & 23.64 & 9.00 & 7.41 & 71.30 & 71.97 & 31.68 & 25.12 & 42.97 & 41.27 & 24.00 & 19.75 & 48.65 & 46.12  & 52.73 & 47.06 \\
\rowcolor{yellow!15} \cellcolor{white} & Moirai 1.1 R Large~\evidencestream{XGBoost} & 22.00 & 25.53 & 38.00 & 27.15 & 34.00 & 29.58 & 73.13 & 73.42 & 30.91 & 24.11 & 49.74 & 49.86 & 31.33 & 27.42 & 51.26 & 49.13  & 50.30 & 50.08 \\
\rowcolor{yellow!15} \cellcolor{white} & Moirai 1.1 R Large~\evidencestream{SVM} & 27.00 & 28.90 & 31.00 & 23.60 & 11.00 & 7.05 & 82.44 & 82.79 & 34.33 & 27.08 & 45.83 & 45.26 & 23.00 & 19.85 & 54.20 & 51.71  & 61.52 & 57.54 \\
\cmidrule(l){2-20}
 & Moirai 2 Small~KNN & 19.00 & 20.21 & 37.00 & 27.83 & 18.00 & 18.38 & 47.97 & 47.78 & 26.05 & 20.37 & 31.94 & 29.90 & 24.67 & 22.14 & 35.32 & 32.69  & 49.39 & 42.94 \\
 & Moirai 2 Small~XGBoost & 20.00 & 20.69 & 35.00 & 25.50 & 17.00 & 17.32 & 55.18 & 54.84 & 28.48 & 22.44 & 38.80 & 38.91 & 24.00 & 21.17 & 40.82 & 38.73  & 55.76 & 55.76 \\
 & Moirai 2 Small~SVM & 18.00 & 20.39 & 38.00 & 27.92 & 1.00 & 0.98 & 61.73 & 59.99 & 28.92 & 22.68 & 37.93 & 36.51 & 19.00 & 16.43 & 42.86 & 39.73  & 61.52 & 58.78 \\
\cmidrule(l){2-20}
 & Chronos 2~KNN & 25.00 & 20.94 & 13.00 & 13.31 & 24.00 & 18.83 & 64.09 & 64.50 & 20.42 & 17.67 & 30.99 & 29.20 & 20.67 & 17.69 & 38.50 & 37.12  & 59.09 & 58.95 \\
 & Chronos 2~XGBoost & 21.00 & 19.80 & 11.00 & 10.98 & 12.00 & 12.11 & 59.63 & 59.84 & 18.76 & 19.78 & 31.86 & 31.69 & 14.67 & 14.29 & 36.75 & 37.11  & 54.24 & 53.21 \\
 & Chronos 2~SVM & 29.00 & 29.26 & 28.00 & 24.64 & 4.00 & 3.39 & 75.62 & 74.94 & 28.04 & 22.64 & 34.29 & 32.78 & 20.33 & 19.09 & 45.98 & 43.45  & 56.67 & 52.19 \\
\cmidrule(l){2-20}
 & Chronos 2 Small~KNN & 18.00 & 16.65 & 22.00 & 22.29 & 23.00 & 18.46 & 52.16 & 52.61 & 16.11 & 13.76 & 24.39 & 21.24 & 21.00 & 19.13 & 30.89 & 29.20  & 52.12 & 50.00 \\
 & Chronos 2 Small~XGBoost & 14.00 & 15.00 & 22.00 & 17.01 & 12.00 & 11.89 & 59.37 & 59.59 & 16.67 & 15.61 & 28.99 & 27.88 & 16.00 & 14.63 & 35.01 & 34.36  & 58.48 & 58.42 \\
\multirow{-12}{*}{\shortstack[l]{Time-series\\foundation\\models}} & Chronos 2 Small~SVM & 13.00 & 12.70 & 10.00 & 9.76 & 12.00 & 9.15 & 65.01 & 63.76 & 17.77 & 15.14 & 30.03 & 28.06 & 11.67 & 10.54 & 37.60 & 35.65  & 61.21 & 60.89 \\
\midrule
\agentbaselineresultrows
\bottomrule
\end{tabular}%
};
\draw[line width=0.35pt,dash pattern=on 2pt off 1.5pt]
  ([xshift=-\challengecustomerwidth,yshift=-0.8pt]challengegrid.north east) --
  ([xshift=-\challengecustomerwidth,yshift=0.8pt]challengegrid.south east);
\draw[line width=0.35pt,dash pattern=on 2pt off 1.5pt]
  ([xshift=-\challengeavgcustomerwidth,yshift=-0.8pt]challengegrid.north east) --
  ([xshift=-\challengeavgcustomerwidth,yshift=0.8pt]challengegrid.south east);
\end{tikzpicture}%
}
\end{table}
}

\challengeresultscombinedtable
\FloatBarrier

\subsection{Training-Free Comparison with ZARA and FETA}
\label{par:training-free-comparison}

\begin{table}[H]
\caption{Performance comparison of training-free (TF) Agentic4PDM against ZARA and FETA across two public benchmarks, BFH and GRS, on UO, PU, and MCC5. ACC and F1-M denote accuracy and macro-F1, respectively; AVG is the unweighted mean across datasets. TF uses no fitted-model predictions. Bold indicates the best results in each column, including ties.}
\label{tab:training-free-agentic-comparison}
\centering\scriptsize
\setlength{\tabcolsep}{2.1pt}
\renewcommand{\arraystretch}{1.12}
\settowidth{\challengemetricwidth}{\textbf{00.00}}
\resizebox{\textwidth}{!}{%
\begin{tabular}{@{}ll*{6}{w{r}{\challengemetricwidth}}|*{6}{w{r}{\challengemetricwidth}}|*{4}{w{r}{\challengemetricwidth}}@{}}
\toprule
\multirow{3}{*}{Group} & \multirow{3}{*}{Method} & \multicolumn{6}{c}{\textbf{\shortstack{Baseline-Failure\\Hard (BFH)}}} & \multicolumn{6}{c}{\textbf{\shortstack{Geometry-Stratified\\Response-Stability (GRS)}}} & \multicolumn{4}{c}{\textbf{AVG}} \\
\cmidrule(lr){3-8}\cmidrule(lr){9-14}\cmidrule(lr){15-18}
& & \multicolumn{2}{c}{UO} & \multicolumn{2}{c}{PU} & \multicolumn{2}{c}{\newred{MCC5}} & \multicolumn{2}{c}{UO} & \multicolumn{2}{c}{PU} & \multicolumn{2}{c}{\newred{MCC5}} & \multicolumn{2}{c}{BFH} & \multicolumn{2}{c}{GRS} \\
\cmidrule(lr){3-4}\cmidrule(lr){5-6}\cmidrule(lr){7-8}\cmidrule(lr){9-10}
\cmidrule(lr){11-12}\cmidrule(lr){13-14}\cmidrule(lr){15-16}\cmidrule(lr){17-18}
& & ACC & F1-M & ACC & F1-M & ACC & F1-M & ACC & F1-M & ACC & F1-M & ACC & F1-M & ACC & F1-M & ACC & F1-M \\
\midrule
 & ZARA~\citep{li2026zara}
& 45.00 & 39.30 & 39.00 & 40.31 & 35.00 & 24.90
& 63.43 & 65.26 & 41.94 & \boldnum{43.23} & 47.66 & 44.22
& 39.67 & 34.84 & 51.01 & 50.90 \\
\multirow{-2}{*}{Agent} & FETA~\citep{sui2025feta}
& 34.00 & 32.32 & 36.00 & 33.84 & 17.00 & 17.65
& 53.73 & 53.46 & 35.21 & 30.79 & 38.02 & 36.90
& 29.00 & 27.94 & 42.32 & 40.38 \\
\midrule
\rowcolor{green!10}
\cellcolor{white}\textbf{Ours} & Agentic4PDM (TF)
& \boldnum{63.00} & \boldnum{62.12} & \boldnum{44.00} & \boldnum{42.30} & \boldnum{42.00} & \boldnum{40.73}
& \boldnum{81.52} & \boldnum{81.40} & \boldnum{47.68} & 40.78 & \boldnum{50.69} & \boldnum{50.14}
& \boldnum{49.67} & \boldnum{48.38} & \boldnum{59.96} & \boldnum{57.44} \\
\bottomrule
\end{tabular}%
}
\end{table}

Following ZARA's notion of \emph{training-free inference} \citep{li2026zara}, the training-free (TF) variant uses no task-specific trained classifier to predict test labels and does not fine-tune its pretrained models. For \method{}, this means omitting the fitted SVM and XGBoost predictions. Table~\ref{tab:training-free-agentic-comparison} reports per-dataset BFH and GRS scores for TF, ZARA, and FETA, together with unweighted means over UO, PU, and MCC5.

Even without fitted-model predictions, TF outperforms ZARA and FETA on all four challenge-average measures, showing that \method{} remains effective when these predictions are unavailable. Compared with TF, standard mode (Table~\ref{tab:challenge-results-combined}) improves mean GRS accuracy and macro-F1 by 2.23 and 1.76 percentage points, respectively, and mean BFH accuracy by 0.66 points. TF nevertheless achieves higher mean BFH macro-F1 (48.38\% versus 47.47\%). These results suggest that fitted-model predictions can provide useful candidate-class anchors for the \agent{Judge}'s reasoning and subsequent tool-based investigation, particularly on GRS, although their benefit varies across challenges and metrics.

\subsection{Performance Across Three Runs}
\label{app:run-variability}

\begin{figure*}[h]
\centering
\includegraphics[width=\textwidth]{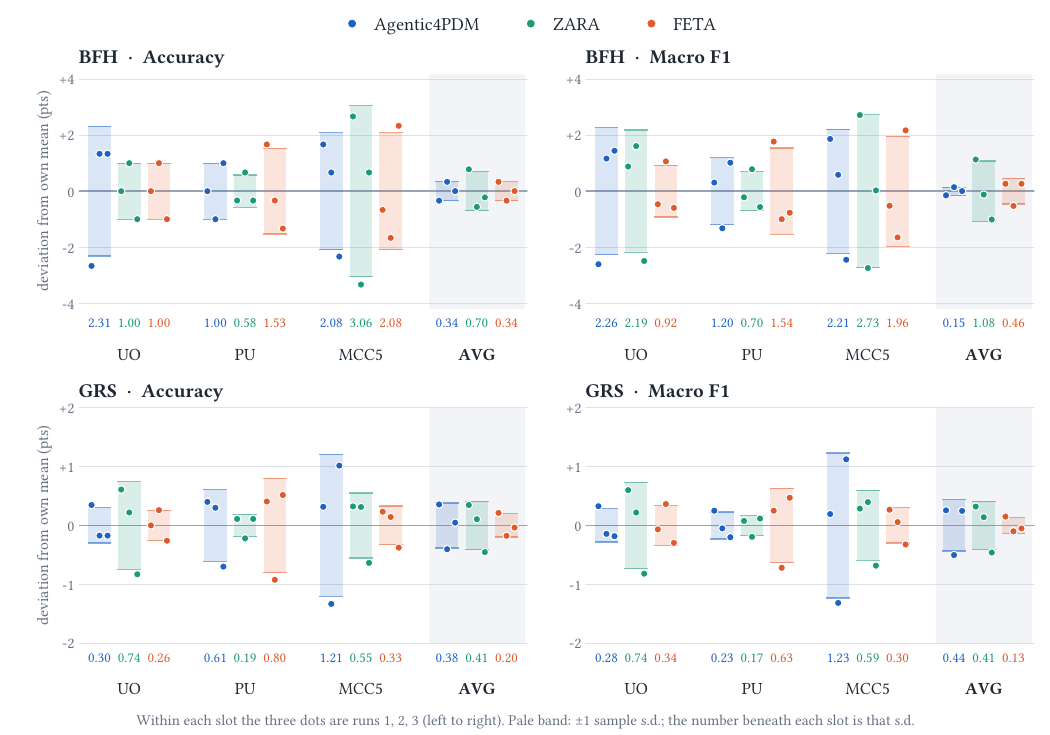}
\Description{Four panels for BFH and GRS, each for accuracy and macro F1. For UO, PU, MCC5 Motor and their average, three dots per method show how far each of three executions of Agentic4PDM, ZARA and FETA lies from that method's own mean, with a pale band for one standard deviation and the deviation printed beneath.}
\caption{Variation across three executions of standard \method{}, ZARA, and FETA on BFH and GRS. For each method, dataset, and metric, scores are centered on their mean across the three executions. The dots therefore show deviations from that mean in percentage points, with runs 1--3 ordered from left to right within each group. Shaded bands span $\pm1$ sample standard deviation ($n=3$); the corresponding standard deviation appears below each group. AVG denotes the unweighted mean of the three dataset scores in each execution. BFH and GRS use different vertical scales.}
\label{fig:run-variability}
\end{figure*}

Figure~\ref{fig:run-variability} summarizes the observed variation between executions. On GRS, the scores averaged over the three datasets vary little for all three methods. On BFH, standard \method{} has a smaller spread in the average macro F1 than ZARA. Individual datasets show greater variation on BFH than on GRS. Each BFH dataset contains 100 windows, so changing the correctness of one prediction changes accuracy by one percentage point. These comparisons describe the three observed executions; they do not establish the variability expected over a larger number of runs.

\subsection{Agentic4PDM versus Ensembles of Its Evidence Sources}
\label{app:evidence-ensemble-comparison}

To assess whether a direct combination of the available source judgments can account for the diagnostic performance, we compare \method{} with \emph{Flat source plurality}. This ensemble combines the current-\newred{sample} \agent{NAAgent} and \agent{ABAgent} outputs, the final \agent{LBAgents} conclusion, and predictions from the SVM and XGBoost baselines. It provides a fixed aggregation of these sources against which to compare the full diagnostic workflow.

\paragraph{Flat source plurality.}
Each valid \agent{Analyst} and the final \agent{LBAgents} conclusion contributes one count to every class it selects, with duplicate selections removed within each source. SVM and XGBoost contribute one vote each for their predicted class. The class with a unique highest total is the ensemble prediction; a tie for the highest total or the absence of valid votes produces an abstention. The aggregation uses the saved source outputs and does not learn source weights.

\paragraph{Comparison at matched coverage.}
For each dataset, we first select the samples on which Flat source plurality produces a prediction, using predictions only, before consulting the true labels. We then score both methods on exactly those samples. \newred{Macro-F1} (F1-M) is computed on this covered set by averaging the per-class F1 scores equally over all dataset classes, with undefined class F1 set to zero. Coverage is the fraction of the available successful samples included in the comparison. Table~\ref{tab:evidence-source-ensemble} reports F1-M at the ensemble's coverage.

\begin{table}[H]
\caption{\newred{Macro-F1} (F1-M, \%) of Flat source plurality and \method{} on the same covered samples in each dataset. Bold marks the higher score.}
\label{tab:evidence-source-ensemble}
\centering\small
\setlength{\tabcolsep}{12pt}
\renewcommand{\arraystretch}{1.25}
\begin{tabular}{lccc}
\toprule
\rowcolor{black!5}
Method & MCC5 & PU & UO \\
\midrule
Flat source plurality & 48.26 & 28.45 & 85.00 \\
\rowcolor{ourslight}
\textbf{\method{}} & \boldnum{54.09} & \boldnum{37.90} & \boldnum{87.38} \\
\bottomrule
\end{tabular}
\par\smallskip
\begin{minipage}{0.85\linewidth}
\footnotesize\emph{Evaluation note.} Matched coverage is 77.69\% on MCC5, 84.11\% on PU, and 89.52\% on UO. Agentic4PDM uses saved classes, including diagnostic guesses on unresolved samples. Both rows use the same covered \newred{sample} identifiers within each dataset.
\end{minipage}
\end{table}

\paragraph{Result analysis.}
\method{} achieves higher \newred{macro-F1} than Flat source plurality on all three matched sets, with margins of 5.83 percentage points on MCC5, 9.45 on PU, and 2.38 on UO. The largest difference occurs on PU, where direct aggregation reaches 28.45\% \newred{macro-F1} and the full workflow reaches 37.90\%. These results show that the full workflow's saved predictions are not reproduced by simply counting the classes supported by its \agent{Analysts}, \agent{LBAgents} conclusion, and two baseline models. They are consistent with additional value from the workflow's investigation and evidence resolution, although this comparison does not isolate the contribution of any individual mechanism.

\subsection{Evidence Guidance}
\label{app:gate-saved-log-audit}
We examine whether the reference preflight provides a useful distinction between corroborative and advisory guidance to the \agent{Judge}. We measure how often each role is available and compare its correctness on the same query samples. This analysis reuses saved measurement records from the GRS evaluations on MCC5, PU, and UO described in Section~\ref{sec:exp-settings}.

For each dataset, we use the GRS query samples in these saved runs. For each query, we take the packets recorded in its ledger and count each measurement once. Each packet carries its role and its winning class (Section~\ref{sec:request-gate}).

\textit{Coverage.} Table~\ref{tab:gate-audit-1} reports, for each role, the percentage of GRS query samples that have at least one packet with that role. Every GRS query sample in the saved runs is counted in the denominator, including those with no packet of that role. A query can have packets of both roles, so the two percentages need not sum to 100\%.

\textit{Correctness.} Table~\ref{tab:gate-audit-2} compares the two roles on the same queries: it uses only the queries that have at least one corroborative packet and at least one advisory packet. For each of these queries and each role, we compute the fraction of packets with that role whose winning class is the true class of the query. We then average these fractions over the queries, so that each query has equal weight regardless of how many packets it has. A packet with no unique winning class is still counted under the role of its tool and is scored as incorrect. The true classes are used only to score the saved outputs after inference.

\begin{table}[H]
\centering\small
\setlength{\tabcolsep}{4pt}
\renewcommand{\arraystretch}{1.15}
\begin{minipage}[t]{0.48\linewidth}
\vspace{0pt}
\caption{\newred{Influence-role} coverage (\%). Percentage of GRS query samples with at least one packet of each role in the \newred{ledger}.}
\label{tab:gate-audit-1}
\end{minipage}\hfill
\begin{minipage}[t]{0.48\linewidth}
\vspace{0pt}
\caption{One-packet guidance on the same GRS query samples containing both roles. Scores are percentages; difference is corroborative minus advisory (percentage points).}
\label{tab:gate-audit-2}
\end{minipage}
\par
\begin{minipage}[t]{0.48\linewidth}
\centering
\begin{tabular}{@{}lrr@{}}
\toprule
Dataset & Corroborative coverage & Advisory coverage\\
\midrule
MCC5 & 30.30\% & 83.68\%\\
PU & 64.35\% & 86.53\%\\
UO & 68.15\% & 96.85\%\\
\bottomrule
\end{tabular}
\end{minipage}\hfill
\begin{minipage}[t]{0.48\linewidth}
\centering
\begin{tabular}{@{}lrrr@{}}
\toprule
Dataset & Corroborative & Advisory & Difference\\
\midrule
MCC5 & 55.56\% & 34.26\% & +21.30 pp\\
PU & 48.06\% & 47.86\% & +0.20 pp\\
UO & 82.24\% & 64.64\% & +17.60 pp\\
\bottomrule
\end{tabular}
\end{minipage}
\end{table}

Results in Tables~\ref{tab:gate-audit-1} and~\ref{tab:gate-audit-2} support two findings:

\paragraph{Finding 1: corroborative evidence appears less often.}
Table~\ref{tab:gate-audit-1} shows lower corroborative coverage on every dataset: 30.30\% versus 83.68\% on MCC5, 64.35\% versus 86.53\% on PU, and 68.15\% versus 96.85\% on UO. Thus, reference-checked guidance is available in fewer samples than advisory guidance.

\paragraph{Finding 2: higher corroborative correctness on matched UO and MCC5 samples.}
Table~\ref{tab:gate-audit-2} shows higher corroborative correctness on UO (82.24\% versus 64.64\%) and MCC5 (55.56\% versus 34.26\%). PU is nearly equal (48.06\% versus 47.86\%). These results support the hypothesis on the matched UO and MCC5 sets, with the clearest descriptive support on UO. Together, the two findings show that corroborative evidence appears less often, yet provides stronger guidance on most datasets (UO and MCC5).

\section{Can the Best Single Tool on Each Dataset Replace Agentic4PDM?}
\label{app:single-tool-transfer}
A fixed measurement provides a useful comparison for the full diagnostic workflow, but the comparison must use tools that the agent can actually select. The current production selector exposes 25 default ToolCards, one per family, with fixed parameters (Table~\ref{tab:quantitative-tool-catalog}). We therefore select the highest-macro-F1 default on each GRS dataset and evaluate that same configuration across all three datasets. Tool selection uses these evaluated samples, so the results describe the strongest defaults on these sets rather than performance estimated on a separate selection split.

\paragraph{Measurement and scoring.}
Each tool compares each numeric channel with its class references. A channel votes only when its winning class is unique; a unique largest class-vote count determines the \newred{sample} prediction. A top tie or the absence of a valid vote is recorded as unknown. Accuracy includes every \newred{sample}, and \newred{macro-F1} averages over the true dataset classes, with unknown predictions contributing false negatives. This uniform offline comparison does not override the ToolCards' inference permissions: for example, the residual-level tool supports fault existence, not fault-type adjudication.

\begin{table}[H]
\caption{\newred{Macro-F1} (\%) of the dataset-winning tools among the 25 selectable defaults and the full \method{} in the updated GRS audit. Agentic4PDM uses saved-class scoring. Bold marks the highest displayed score in each column.}
\label{tab:single-tool-transfer}
\centering\footnotesize
\setlength{\tabcolsep}{3pt}
\renewcommand{\arraystretch}{1.25}
\begin{tabular*}{\textwidth}{@{\extracolsep{\fill}}>{\raggedright\arraybackslash}p{0.27\textwidth}>{\raggedright\arraybackslash}p{0.35\textwidth}rrrr@{}}
\toprule
\rowcolor{black!5}
Tool (selected on) / method & ToolCard / scoring & MCC5 & PU & UO & Mean \\
\midrule
Windowed profile (MCC5) & \nolinkurl{quant_windowed_profile_windows_16_amplitude_l1} & \cellcolor{yellow!18}36.15 & 35.22 & 30.17 & 33.85 \\
Crossing-run statistics (PU) & \nolinkurl{quant_crossing_run_statistics_q50_js} & 22.89 & \cellcolor{yellow!18}37.94 & 16.73 & 25.85 \\
Distribution KS (UO) & \nolinkurl{quant_distribution_ks_raw} & 31.78 & 32.78 & \cellcolor{yellow!18}60.02 & 41.53 \\
\midrule
\rowcolor{ourslight}
\textbf{\method{}} & Saved class & \boldnum{52.93} & \boldnum{38.54} & \boldnum{86.12} & \boldnum{59.20} \\
\bottomrule
\end{tabular*}
\par\smallskip
\noindent\raggedright\footnotesize\emph{Design note.} The three fixed tools were selected by \newred{macro-F1} within the 25 defaults on the dataset in parentheses, then scored unchanged on the other datasets. Pale yellow marks the selection dataset; pale blue marks the full workflow. Mean weights the three datasets equally. Saved-class scoring includes the agent's best guesses on unresolved samples.
\end{table}

\paragraph{Dataset dependence and full-workflow performance.}
The winning default changes with the dataset. Crossing-run statistics ranks first on PU but fifteenth on MCC5 and twentieth on UO, whereas the windowed amplitude profile is more consistent, ranking first, fourth, and fifth on MCC5, PU, and UO. Thus, the defaults include both dataset-specific and more broadly useful measurements. Distribution KS has the highest equal-dataset mean \newred{macro-F1} among all 25 defaults (41.53\%). The full workflow reaches 59.20\% under saved-class scoring. Under saved-class scoring, its margins over each dataset's best default are 16.78, 0.60, and 26.10 percentage points on MCC5, PU, and UO, respectively.

\paragraph{Scope of the comparison.}
The audit reuses saved offline tool predictions; it does not rerun tool or LLM inference. Historical agent implementation hashes differ from the current registry for the 11 tool identifiers observed in the archived runs. The full workflow also uses retrieval, multiple measurements, baseline evidence, and \agent{Judge} reasoning. These rows therefore provide configuration-level comparisons and full-agent context, rather than a controlled ablation isolating adaptive tool selection. The agent scores here follow the updated audit and its scoring conventions; they should be read within that scope rather than substituted for the main experiment's aggregate results.
\FloatBarrier

\FloatBarrier
\Needspace{6\baselineskip}

\section{Backbone Substitution: What the Agent Adds to a Representation}
\label{app:backbone-substitution}

The main results use Moirai~1.1~R~Large (M1.1R). An archived Chronos-2~Small (C2S) experiment substitutes the frozen retrieval encoder and its embedding cache. In standard mode, its SVM and XGBoost heads are fitted to C2S rather than M1.1R embeddings; in TF, neither backbone supplies fitted-model predictions. The LLM, decoding, prompts, budgets, thresholds, and BFH/GRS manifests are shared. These are different backbones, not repeat executions of one configuration.

\begin{figure}[H]
\centering
\includegraphics[width=\linewidth]{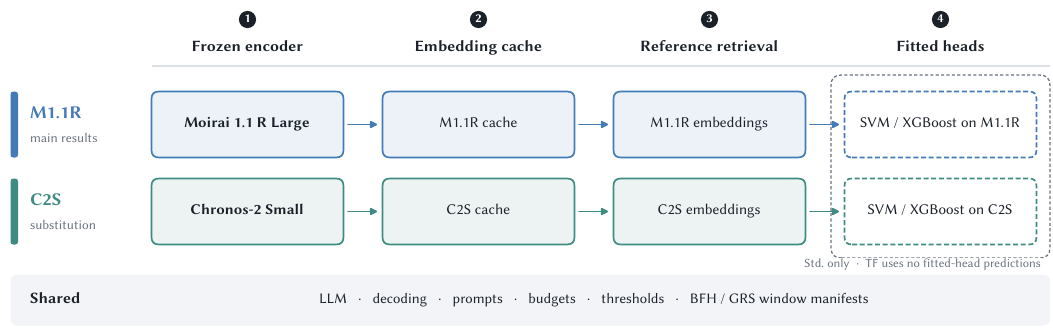}
\Description{Two parallel pipelines compare M1.1R and C2S. Each has a frozen encoder, its own embedding cache and retrieval representation, and SVM and XGBoost heads fitted on that backbone. TF uses no fitted-head predictions; the LLM, prompts, budgets, thresholds, and window manifests are shared.}
\caption{Backbone substitution changes the frozen encoder, embedding cache, retrieval representation, and standard fitted heads. TF uses no fitted-head predictions. All other settings and BFH/GRS windows are shared.}
\label{fig:backbone-config-diff}
\end{figure}

\begin{figure}[H]
\centering
\includegraphics[width=\linewidth]{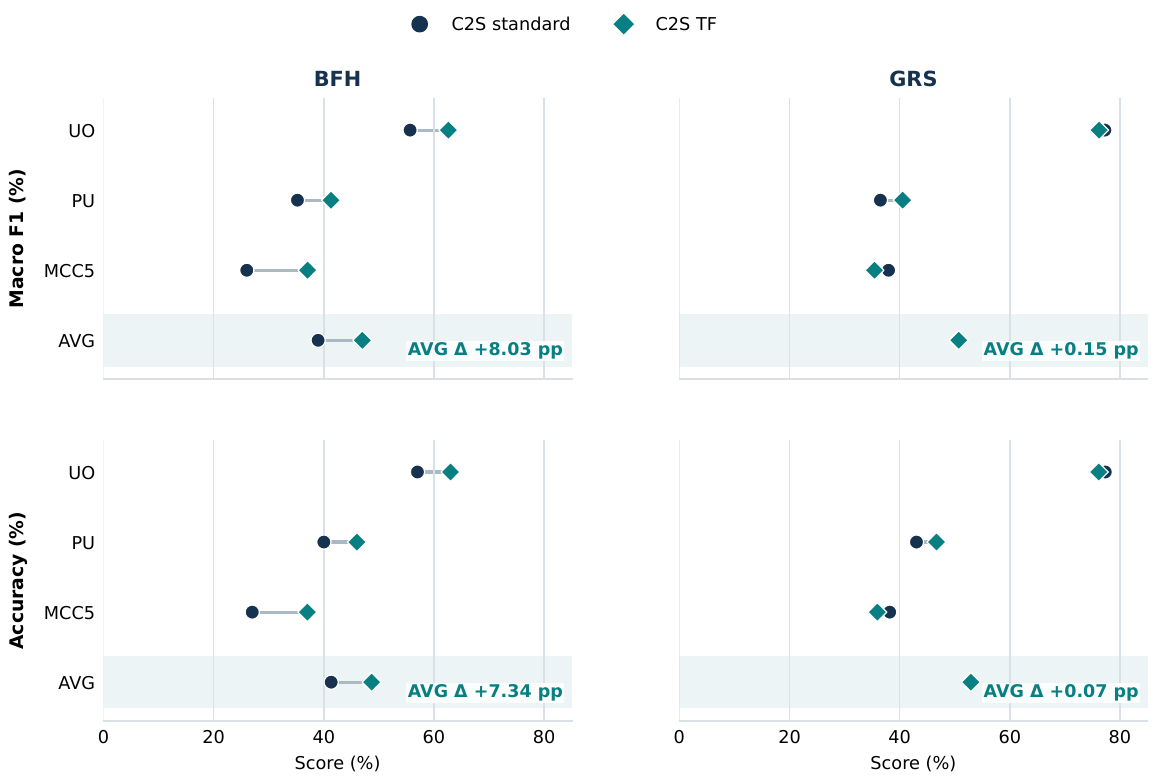}
\Description{Four paired dot plots compare C2S standard with C2S TF for BFH and GRS, each measured by macro F1 and accuracy on UO, PU, MCC5, and their mean. The mean BFH changes are plus 8.03 and plus 7.34 percentage points; GRS changes are plus 0.15 and plus 0.07 points.}
\caption{C2S \method{} standard and TF on the same frozen BFH and GRS windows. Circles and diamonds show \protect\rev{the reported scores}; connectors expose the direction and size of each change. ACC and macro F1 are percentages. AVG is the unweighted mean of UO, PU, and MCC5, and its annotation gives TF minus standard in percentage points.}
\label{fig:c2s-agent-results}
\end{figure}

\begin{figure}[H]
\centering
\includegraphics[width=\linewidth]{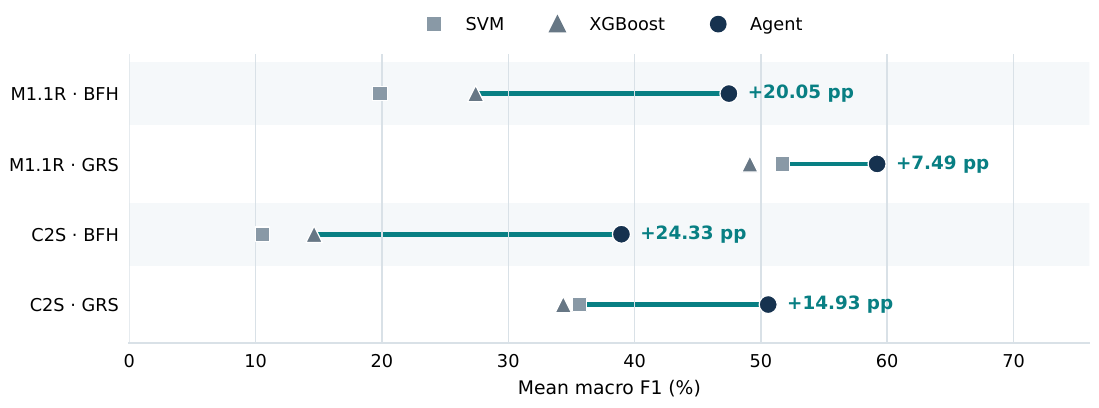}
\Description{Four rows compare SVM, XGBoost, and agent mean macro F1 for M1.1R and C2S on BFH and GRS. The agent exceeds the stronger fitted head by 20.05, 9.93, 24.33, and 14.93 percentage points, respectively.}
\caption{Mean macro F1 (\%) for each standard agent and the two fitted heads built on the same backbone. The labeled gain is over the stronger fitted head in each row. \protect\rev{These comparisons are descriptive, not causal.}}
\label{fig:same-backbone-gains}
\end{figure}

The agent exceeds both fitted heads for both backbones and both challenges. The C2S gain over its stronger head is larger, although its absolute macro F1 is lower. This separates the workflow's improvement over a given representation from the quality of the representation itself. The comparison is descriptive: it does not isolate the causal effect of the host gates or quantify run-to-run variability.

\FloatBarrier
\Needspace{6\baselineskip}

\section{Empirical Time and Token Use}
\label{app:resource-complexity}

\begin{table}[H]
\centering\small
\caption{Mean recorded resource use per query sample. Token columns are in thousands (k); time is in seconds. Recorded timer scopes differ across methods, as described below.}
\label{tab:resource-complexity}
\setlength{\tabcolsep}{7pt}
\renewcommand{\arraystretch}{1.15}
\begin{tabular*}{\textwidth}{@{\extracolsep{\fill}}lrrrrr@{}}
\toprule
\textbf{Method} & \textbf{Time (s)} & \textbf{Input (k)} & \textbf{Output (k)} & \textbf{Total (k)} & \textbf{LLM calls} \\
\midrule
FETA & 1{,}078.59 & 68.05 & 4.62 & 72.67 & 8.0 \\
ZARA & 98.20 & 5.11 & 2.72 & 7.83 & 4.0 \\
\method{} (standard) & 1{,}645.79 & 89.88 & 12.48 & 102.36 & 17.4 \\
\bottomrule
\end{tabular*}
\end{table}

\begin{figure}[H]
\centering
\includegraphics[width=\textwidth]{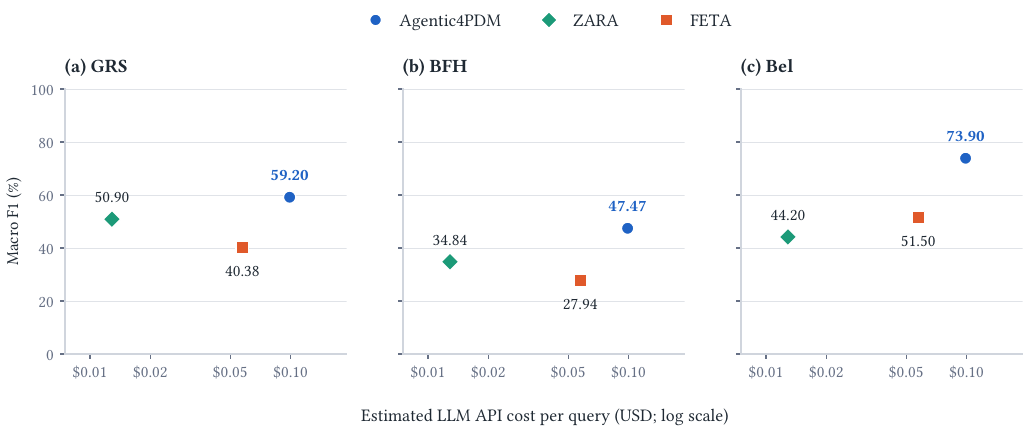}
\caption{Diagnostic \newred{macro-F1} versus estimated Qwen3.6-27B API cost per query at Alibaba Cloud's Singapore/International list prices. Costs follow Eq.~\ref{eq:qwen-api-cost}, using separate input and output token rates. Each method's mean cost estimate is reused across GRS, BFH, and Bel. These values estimate token-based API charges rather than observed API bills or total deployment costs.}
\label{fig:performance-qwen-api-cost}
\end{figure}

We next examine what the additional diagnostic reasoning of \method{} costs in practice. Table~\ref{tab:resource-complexity} reports recorded processing time, token use, and LLM calls. The time column includes the stages recorded by each implementation, including retrieval, feature preparation, tool execution, and LLM inference. The timer scopes differ: FETA measures per-worker processing, whereas ZARA and \method{} measure their full workflows. The time values should therefore be interpreted as descriptive implementation measurements rather than as a strictly controlled latency benchmark. The token counts provide a more direct basis for comparing hosted-LLM usage. We convert the mean input and output tokens into an estimated API charge using Qwen3.6-27B prices of USD\,0.60 per million input tokens and USD\,3.60 per million output tokens. This calculation prices the recorded token usage only and does not rerun the diagnostic workflows on the hosted service.

For method $m$, let $\bar u_m$ and $\bar v_m$ denote the reported mean input and output tokens per query, respectively. The estimated LLM API charge is
\begin{equation}
\widehat c_{\mathrm{API},m}
=
\frac{0.60\,\bar u_m+3.60\,\bar v_m}{10^6}
\quad\text{USD/query}.
\label{eq:qwen-api-cost}
\end{equation}
Using the unrounded token means underlying Table~\ref{tab:resource-complexity} gives estimated costs of USD\,0.0575 for FETA, USD\,0.0129 for ZARA, and USD\,0.0989 for standard \method{} per query.

The three methods spend these resources in substantially different ways. FETA records 1{,}078.59\,s per query, of which 1{,}044.51\,s comes from retrieval and only 34.08\,s from subsequent inference. Its moderate token consumption therefore does not translate into short processing time: the dominant bottleneck is its per-channel DTW retrieval, which repeatedly aligns the query with labeled reference sequences. FETA illustrates an important distinction between LLM cost and end-to-end computational cost---reducing language-model usage does not necessarily make the complete diagnostic workflow inexpensive.

ZARA lies at the other extreme. It requires only 98.20\,s, 7.83k tokens, and four LLM calls per query. Much of its discriminative structure is prepared before inference, allowing the online workflow to operate with relatively little computation. This makes ZARA substantially cheaper at inference time, but also leaves less opportunity to construct new evidence specifically for an ambiguity encountered in the current case.

\method{} intentionally makes the opposite design choice. It records 1{,}645.79\,s, 102.36k tokens, and 17.4 LLM calls per query. Relative to FETA, this corresponds to approximately $1.53\times$ the recorded processing time and $1.41\times$ the total token use, rather than a twofold increase. The additional computation arises from repeated evidence acquisition and adjudication: the \agent{Judge} may identify a remaining ambiguity, request a targeted measurement, inspect the returned evidence, reconcile conflicting results, and continue the investigation before committing to a class. In other words, the additional resource use is not incidental overhead; it is the computational cost of allowing the investigation itself to adapt to the current case.

The relevant question is therefore not whether \method{} is the cheapest workflow, but what diagnostic improvement accompanies this additional expenditure. Figure~\ref{fig:performance-qwen-api-cost} makes this trade-off explicit. Standard \method{} costs an estimated USD\,0.0989 per query, compared with USD\,0.0575 for FETA, or approximately $1.72\times$ as much. In absolute terms, however, the difference is about USD\,0.0414 per query. Against this additional estimated API charge, \method{} improves three-dataset mean \newred{macro-F1} over FETA by 18.82 percentage points on GRS (59.20\% versus 40.38\%) and 19.53 points on BFH (47.47\% versus 27.94\%). The additional inference therefore accompanies substantial gains both across the coverage-oriented GRS regimes and on the concentrated failure cases represented by BFH.

The comparison with ZARA exposes the other side of the trade-off. ZARA has the lowest estimated API cost, at USD\,0.0129 per query, but also achieves lower \newred{macro-F1} than \method{} in all three evaluation settings shown in Figure~\ref{fig:performance-qwen-api-cost}. The gaps are 8.30 percentage points on GRS and 12.63 points on BFH; more importantly, on the unfiltered Bel test set collected from our customer, the gap widens to 29.70 points (73.90\% versus 44.20\%). Thus, the cheapest online workflow is not necessarily the most effective one, particularly when evaluated on data outside the public benchmark settings.

Taken together, these results show a clear resource--performance trade-off. \method{} pays more at inference time because it performs more sample-specific investigation before deciding, whereas FETA concentrates cost in retrieval and ZARA minimizes online computation through stronger offline preparation. Under the pricing scenario considered here, the additional LLM cost of \method{} relative to FETA is only about four cents per query, while the corresponding macro-F1 gains approach twenty percentage points on both GRS and BFH. This does not by itself establish that the method is economically preferable in every industrial deployment. Rather, it makes the relevant deployment question more precise: whether the incremental cost of investigating an escalated case is smaller than the operational value of the diagnostic errors that the additional investigation prevents.

\section{How consistent is performance across operating conditions?}
\label{par:condition-accuracy}

\begin{figure}[h]
\centering
\includegraphics[width=0.5\linewidth]{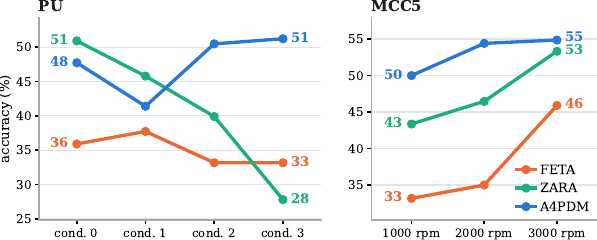}
\caption{Accuracy of FETA, ZARA, and \method{} across the displayed PU operating conditions and MCC5 motor speeds.}
\label{fig:condition-accuracy}
\end{figure}

Figure~\ref{fig:condition-accuracy} examines whether the aggregate advantage of \method{} is concentrated in particular operating conditions. Across both datasets, \method{} achieves higher mean accuracy than FETA and ZARA (Table~\ref{tab:challenge-results-combined}) while exhibiting comparatively limited variation across the displayed regimes. On PU, its accuracy spans 10 percentage points across conditions, compared with 23 points for ZARA. On MCC5, the corresponding ranges are 5 points for \method{} and 10 for ZARA. FETA shows relatively small variation, particularly on PU, but at a substantially lower accuracy level on both datasets. These results indicate that the higher average performance of \method{} is accompanied by comparatively consistent behavior across the evaluated operating conditions rather than being attributable to a single favorable regime.

\section{Experimental Setup: Data, Protocol, and Per-Method Configurations}
\label{app:experimental-setup}

\subsection{Datasets, Signal Modalities, and Split Boundaries}
\label{app:datasets}
Table~\ref{tab:dataset-summary} summarizes the four diagnostic datasets, including their class counts, labeled training budgets, fault settings, operating conditions, and available sampling rates. PU and UO cover three-class bearing diagnosis, MCC5 covers four-class motor diagnosis, and Bel provides a separate binary bearing evaluation on data supplied by our customer (Section~\newred{\ref{sec:exp-settings}}). Table~\ref{tab:challenge-manifests} reports the total number of test samples and per-class support in the frozen BFH and GRS sets for PU, UO, and MCC5, together with the complete Bel test set (330 samples: 150 normal and 180 abnormal). BFH uses balanced or nearly balanced class quotas to test recovery on concentrated baseline failures, while GRS retains representatives across signal geometry and model-response patterns, yielding class counts determined by the coverage construction. Details of the BFH and GRS test-set selection rules and manifest construction are provided in Appendix~\ref{sec:dual-challenge-benchmarks}, \emph{Benchmark Construction: How the BFH and GRS Challenge Sets Were Built}.
\paragraph{Split boundaries.} For PU, UO, MCC5 and BEL, the labeled training samples and the test samples come from different recordings: no recording contributes samples to both sets, so a test sample and its references never come from the same recording. The preceding sample $x^{\mathrm{prev}}$ used by the \agent{LBAgents} is the sample just before the query in its own recording, and it is never part of the labeled training set.

\begin{table}[H]
\caption{Acquisition characteristics and retained-task statistics for PU, UO, MCC5, and Bel. Bel is evaluated separately from BFH and GRS; a dash indicates an unreported sampling rate.}
\label{tab:dataset-summary}
\centering\footnotesize
\setlength{\tabcolsep}{2pt}
\renewcommand{\arraystretch}{1.12}
\begin{tabular*}{\textwidth}{@{\extracolsep{\fill}}>{\raggedright\arraybackslash}p{0.08\textwidth}>{\centering\arraybackslash}p{0.075\textwidth}>{\centering\arraybackslash}p{0.08\textwidth}>{\raggedright\arraybackslash}p{0.20\textwidth}>{\raggedright\arraybackslash}p{0.445\textwidth}>{\raggedleft\arraybackslash}p{0.07\textwidth}@{}}
\toprule
\textbf{Dataset} & \textbf{Classes} & \textbf{Train} & \textbf{Fault generation / setting} & \textbf{Operating conditions} & \textbf{Rate (Hz)} \\
\midrule
\textbf{PU}~\citep{Lessmeier2016PU} & 3 & 150 & Real bearing damage from accelerated lifetime tests & Condition 0: 900\,rpm, 0.7\,Nm, 1000\,N. Conditions 1--3: 1500\,rpm with torque/radial-load pairs 0.1\,Nm/1000\,N, 0.7\,Nm/400\,N, and 0.7\,Nm/1000\,N. & 64{,}000 \\
\midrule
\textbf{UO}~\citep{HuangBaddour2018UOBearing} & 3 & 150 & Inner-/outer-race defect specimens; fabrication method not specified in the source description & Time-varying speed: A, increasing; B, decreasing; C, increasing then decreasing; D, decreasing then increasing. & 200{,}000 \\
\midrule
\textbf{MCC5}~\citep{chen2026motor} & 4 & 200 & Laboratory motor measurements with rotor, electrical, and bearing fault configurations & Frozen scored pool: 20\,Nm at 1000, 2000, and 3000\,rpm (Conditions 0--2). & 12{,}800 \\
\midrule
\textbf{Bel} & 2 & 20 & Bearing vibration supplied by our customer; binary normal/abnormal labels, with damage type unspecified &  - & -- \\
\bottomrule
\end{tabular*}
\end{table}

\subsubsection{Frozen Manifests and Per-Class Support}
\label{app:benchmark-manifests}
\label{app:challenge-evaluation-rationale}

Table~\ref{tab:challenge-manifests} lists the frozen BFH and GRS test-set sizes and per-class sample counts, alongside the Bel test set, which is scored in full without BFH or GRS selection.

\begin{table}[H]
\caption{Test-set sizes and per-class support for the frozen BFH/GRS manifests and the complete Bel test set. Each class count is a number of samples; class counts sum to the total $N$ in each row.}
\label{tab:challenge-manifests}
\centering\small
\setlength{\tabcolsep}{6pt}
\renewcommand{\arraystretch}{1.18}
\begin{tabular*}{0.90\linewidth}{@{\extracolsep{\fill}}llrrrr@{}}
\multicolumn{6}{@{}l}{\textbf{(a) Bearing diagnosis}} \\
\addlinespace[3pt]
\toprule
\textbf{Dataset} & \textbf{Challenge} & \textbf{Total} $N$ & \textbf{Healthy} & \textbf{Inner-race} & \textbf{Outer-race} \\
\midrule
\multirow{2}{*}{PU} & BFH & 100 & 34 & 33 & 33 \\
& GRS & 906 & 205 & 487 & 214 \\
\addlinespace[4pt]
\multirow{2}{*}{UO} & BFH & 100 & 34 & 33 & 33 \\
& GRS & 763 & 197 & 221 & 345 \\
\bottomrule
\end{tabular*}

\medskip
\begin{tabular*}{0.90\linewidth}{@{\extracolsep{\fill}}lrrrrr@{}}
\multicolumn{6}{@{}l}{\textbf{(b) Motor diagnosis: MCC5}} \\
\addlinespace[3pt]
\toprule
\textbf{Challenge} & \textbf{Total} $N$ & \textbf{Normal} & \textbf{Rotor} & \textbf{Electrical} & \textbf{Bearing} \\
\midrule
BFH & 100 & 25 & 25 & 25 & 25 \\
GRS & 1{,}152 & 233 & 341 & 175 & 403 \\
\bottomrule
\end{tabular*}

\medskip
\begin{tabular*}{0.90\linewidth}{@{\extracolsep{\fill}}lrrr@{}}
\multicolumn{4}{@{}l}{\textbf{(c) Binary bearing diagnosis: Bel}} \\
\addlinespace[3pt]
\toprule
\textbf{Evaluation set} & \textbf{Total} $N$ & \textbf{Normal} & \textbf{Abnormal} \\
\midrule
Complete test set & 330 & 150 & 180 \\
\bottomrule
\end{tabular*}
\end{table}

\FloatBarrier

\subsection{Shared Evaluation Protocol and Inventory of Compared Methods}
\label{app:datasets-baselines}
Table~\ref{tab:baseline-overview} groups the comparison methods into deep-learning, foundation-model, and agentic families. The evaluated non-agent baselines include three neural classifiers and 12 combinations of foundation-model representations with classical heads. The latter pair each of four backbones with $k$-nearest neighbors (KNN), a support vector machine (SVM), or extreme gradient boosting (XGBoost), allowing comparison across representations and classifiers. These non-agent methods produce diagnostic predictions without adaptive evidence construction or \host{}-governed adjudication.

Using a fixed or pretrained representation with a separately fitted classifier follows an established evaluation paradigm in time-series learning, rather than a task-specific architectural modification. Franceschi et al.\ learn transferable unsupervised representations and evaluate them with an RBF-kernel SVM \citep{franceschi2019unsupervised}. TS2Vec likewise evaluates learned representations with an RBF-kernel SVM across the UCR and UEA archives~\citep{yue2022ts2vec}, and MOMENT evaluates pretrained time-series representations using an SVM without task-specific backbone fine-tuning~\citep{Goswami2024MOMENT}. More directly, StarEmbed evaluates fixed Moirai and Chronos embeddings with KNN, a linear probe, random forest, and an MLP~\citep{li2026starembed}. A broader empirical comparison of time-series embeddings also evaluates KNN, SVM, and XGBoost as downstream classifiers~\citep{irani2025embedding}. These precedents motivate the general representation-plus-classifier setting; our particular four-backbone, three-head grid is a controlled comparison within that setting, not a claim that every exact pairing has already been standardized. Varying the head allows us to assess each representation without relying on a single classifier choice.

\begin{table}[H]
\caption{Baseline inventory grouped by model family. Each foundation-model backbone is paired with KNN, SVM, and XGBoost.}
\label{tab:baseline-overview}
\centering\small
\setlength{\tabcolsep}{6pt}
\renewcommand{\arraystretch}{1.18}
\begin{tabular}{@{}>{\raggedright\arraybackslash}p{0.32\linewidth}>{\raggedright\arraybackslash}p{\dimexpr0.68\linewidth-2\tabcolsep\relax}@{}}
\toprule
\textbf{Model / backbone} & \textbf{Prediction / reasoning approach} \\
\midrule
\rowcolor{gray!10} \multicolumn{2}{@{}l@{}}{\strut\textbf{(a) Deep learning}\quad\textit{Evaluated baselines}} \\
\makecell[tl]{CNN-TCC~\citep{eldele2021tstcc}\\QCNN~\citep{QCNN2022}\\WDCNN~\citep{zhang2017wdcnn}} & Neural classifiers \\
\addlinespace[5pt]
\rowcolor{gray!10} \multicolumn{2}{@{}l@{}}{\strut\textbf{(b) Foundation models}\quad\textit{Evaluated baselines}} \\
\makecell[tl]{Moirai 1.1 R Large~\citep{woo2024moirai}\\Moirai 2 Small~\citep{liu2026moirai20timeseries}\\Chronos-2~\citep{ansari2025chronos2}\\Chronos-2 Small~\citep{ansari2025chronos2}} & Fixed representations with KNN, SVM, or XGBoost (12 combinations), following MOMENT's representation-plus-SVM evaluation~\citep{Goswami2024MOMENT}, StarEmbed's fixed-embedding evaluation~\citep{li2026starembed}, and the broader embedding comparison~\citep{irani2025embedding}. \\
\addlinespace[5pt]
\rowcolor{gray!10} \multicolumn{2}{@{}l@{}}{\strut\textbf{(c) Agentic methods}\quad\textit{Evaluated baselines}} \\
\textbf{FETA}~\citep{sui2025feta} & Training-set channel ranking, DTW exemplar retrieval, and channel-level LLM reasoning with confidence-based aggregation. \\
\addlinespace[3pt]
\textbf{ZARA}~\citep{li2026zara} & Offline pairwise feature knowledge; placement-specific, class-wise retrieval and hierarchical feature selection, candidate pruning, and decision reasoning. \\
\bottomrule
\end{tabular}
\par\smallskip
\begin{minipage}{\linewidth}
\footnotesize\textit{Evaluation scope.} Table~\ref{tab:challenge-results-combined} groups the deep-learning, foundation-model, and agent results. Implementation settings for FETA and ZARA are given in Appendix~\ref{app:agentic-baseline-boundaries}.
\end{minipage}
\end{table}

\begin{table}[H]
\caption{Shared labeled-reference and query-\newred{sample} settings for the evaluated methods; BFH/GRS manifest sizes and per-class support are reported in Table~\ref{tab:challenge-manifests}.}
\label{tab:shared-protocol}
\centering\footnotesize
\setlength{\tabcolsep}{5pt}
\renewcommand{\arraystretch}{1.2}
\begin{tabularx}{\linewidth}{@{}>{\raggedright\arraybackslash}p{0.28\linewidth}>{\raggedright\arraybackslash}X@{}}
\toprule
\textbf{Protocol item} & \textbf{Setting} \\
\midrule
Labeled reference pool & 50 samples per class: 150 each for PU and UO (3 classes), and 200 for MCC5 (4 classes) \\
Query \newred{sample} & 2048 \newred{observations}, non-overlapping, using all available dataset channels \\
\bottomrule
\end{tabularx}
\end{table}

These values describe the labeled reference pool and each query, not the scored challenge sets. The shared reference budget does not imply identical pretraining resources or computation; Appendix~\ref{app:benchmark-manifests} reports the separate BFH/GRS manifest sizes and per-class support.

\subsection{Configuration of Agentic4PDM}
\label{app:experimental-protocol-details}

To support reproducibility, we document the fixed inference and retrieval settings used for the reported executions.

\paragraph{Operational settings and sensitivity.}
Challenge construction uses three priority-rule steps, a GRS geometry target of $G{=}250$ for all three datasets, and response-stability thresholds $(0.70,0.50)$ throughout. These are fixed operating choices, not learned parameters or claims of optimality. Increasing $G$ raises geometric resolution and agent-run count, while higher stability thresholds produce smaller, more homogeneous tiers.

\begin{table}[H]
\caption{Main Moirai~1.1~R~Large \method{} configuration. One setting applies to standard and TF on both challenges and all three datasets; TF omits fitted-model outputs. Appendix~\ref{app:backbone-substitution} specifies the Chronos-2~Small substitution.}
\label{tab:agent-config}
\centering\footnotesize
\setlength{\tabcolsep}{6pt}
\renewcommand{\arraystretch}{1.15}
\begin{tabular}{@{}>{\raggedright\arraybackslash}p{0.15\linewidth}>{\raggedright\arraybackslash}p{0.27\linewidth}>{\raggedright\arraybackslash}p{0.52\linewidth}@{}}
\toprule
\textbf{Component} & \textbf{Setting} & \textbf{Value} \\
\midrule
\rowcolor{gray!10}\multicolumn{3}{@{}l@{}}{\strut\textbf{(a) Language model}} \\
& Model and access & Qwen3.6-27B vision-language model, served locally through an OpenAI-compatible endpoint; the same model serves every agent role \\
& Decoding & Temperature 0, top-$p$ 1, top-$k$ $-1$, presence penalty 1.5, seed 0 \\
& Extended thinking & Enabled for the analysis and synthesis roles, 512-token budget \\
& Output limit & 38{,}048 tokens per request \\
& Images & At most 10 per request, longest side 1400\,px \\
\addlinespace[3pt]
\rowcolor{gray!10}\multicolumn{3}{@{}l@{}}{\strut\textbf{(b) Reference retrieval}} \\
& Encoder & Frozen Moirai~1.1~R~Large, mean pooling over all channels, per-dimension feature scaling \\
& Embedding settings & Context length 2048, patch size 32, batch size 100 \\
& Reference selection & Uses the shared labeled-reference pool in Table~\ref{tab:shared-protocol}; retains the highest-scoring reference per class by cosine similarity, with no operating-condition filter \\
\addlinespace[3pt]
\rowcolor{gray!10}\multicolumn{3}{@{}l@{}}{\strut\textbf{(c) Evidence construction}} \\
& Smoothing registry & Five label-free families (Savitzky--Golay, median, exponential weighted mean, median--mean--EWM, low-pass); the agent selects the family per channel and the \host{} fixes the parameters \\
& Channel priorities & Ordinal levels: high, medium, low (or removed) \\
& Verification rounds & 2 for the \agent{Analyst} reports and 2 for the smoothing review \\
& \agent{LBAgents} & \agent{NAAgent} and \agent{ABAgent} comparisons repeated on the previous \newred{sample} only, when a current-\newred{sample} comparison remains unresolved \\
\addlinespace[3pt]
\rowcolor{gray!10}\multicolumn{3}{@{}l@{}}{\strut\textbf{(d) \agent{Judge} and \host{} budgets}} \\
& \agent{Judge} steps & At most $N_J=10$ per \newred{sample}, with no minimum; identical in every ablation variant \\
& \agent{Judge} images & 6 anchor and 8 active \\
& Request gate & At most 5 tool families offered and 3 tools recommended per round (Section~\ref{sec:adjudication}) \\
& Request timeouts & \newred{PU} archived configuration: 1800\,s for stage 1 and 3600\,s for stage 2 \\
& Packet gate & At most 10 tool executions per round, 60 per \newred{sample}, and 10 per registered family; the execution budget is enforced by the \host{} and is identical in every ablation variant, including those in which this gate is disabled \\
\addlinespace[3pt]
\rowcolor{gray!10}\multicolumn{3}{@{}l@{}}{\strut\textbf{(e) Fitted-model evidence (standard only)}} \\
& Revealed predictions & XGBoost and SVM fitted on the same reference samples and on the run's backbone representation \\
& Reveal order & Disclosed only after the \agent{Analyst} reports are frozen; individual model identities stay hidden \\
\bottomrule
\end{tabular}
\end{table}

\subsection{Training Settings of the Deep-Learning Baselines}
\label{app:deep-baseline-settings}

The deep-learning comparison includes CNN-TCC, QCNN, and WDCNN. Table~\ref{tab:deep-config} groups their shared optimizer and schedule settings, then separates the input-dependent convolution settings for bearing and motor data.

\begin{table*}[t]
\caption{Training settings shared by the deep-learning baselines (CNN-TCC, QCNN, and WDCNN).}
\label{tab:deep-config}
\centering\footnotesize
\setlength{\tabcolsep}{6pt}
\renewcommand{\arraystretch}{1.2}
\begin{tabular*}{\textwidth}{@{\extracolsep{\fill}}@{}c c c c c c >{\raggedright\arraybackslash}p{2.15cm} >{\raggedright\arraybackslash}p{2.15cm} >{\raggedright\arraybackslash}p{2.2cm}@{}}
\toprule
\multicolumn{4}{c}{\textbf{Optimizer}} & \multicolumn{2}{c}{\textbf{Schedule}} & \multicolumn{3}{c}{\textbf{Architecture}} \\
\cmidrule(lr){1-4}\cmidrule(lr){5-6}\cmidrule(l){7-9}
Algorithm & $\beta_1$ & $\beta_2$ & Learning rate & Epochs & Batch size & Bearing (3 channels) & Motor (8 channels) & Common \\
\midrule
Adam & 0.9 & 0.95 & $10^{-3}$ & 100 & 128 & First convolution: kernel 16, stride 2 & First convolution: kernel 8, stride 1 & 64 final channels; dropout 0.35 \\
\bottomrule
\end{tabular*}
\end{table*}

\subsection{Training Settings of the Foundation-Model Baselines}
\label{app:foundation-baseline-settings}

The foundation-model baselines pair four frozen encoders---Moirai~1.1~R~Large, Moirai~2~Small, Chronos-2, and Chronos-2~Small---with KNN, SVM, and XGBoost heads. Table~\ref{tab:fm-config} summarizes the shared embedding procedure and each head's settings.

\begin{table}[H]
\caption{Training settings of the foundation-model baselines and their classical heads.}
\label{tab:fm-config}
\centering\footnotesize
\setlength{\tabcolsep}{6pt}
\renewcommand{\arraystretch}{1.15}
\begin{tabular}{@{}>{\raggedright\arraybackslash}p{0.19\linewidth}>{\raggedright\arraybackslash}p{0.23\linewidth}>{\raggedright\arraybackslash}p{0.52\linewidth}@{}}
\toprule
\textbf{Group} & \textbf{Setting} & \textbf{Value} \\
\midrule
\rowcolor{gray!10}\multicolumn{3}{@{}l@{}}{\strut\textbf{(a) Foundation-model representations}} \\
& Backbones & Moirai~1.1~R~Large, Moirai~2~Small, Chronos-2, Chronos-2~Small, all frozen \\
& Embedding & Mean pooling, per-dimension feature scaling, context length 2048, patch size 32, batch size 100 \\
\addlinespace[3pt]
\rowcolor{gray!10}\multicolumn{3}{@{}l@{}}{\strut\textbf{(b) Classical heads on those representations}} \\
& KNN & $k=5$, uniform weights \\
& SVM & RBF kernel, $C=1.0$, $\gamma=\text{scale}$ \\
& XGBoost & 100 trees, maximum depth 6, learning rate 0.1 \\
\bottomrule
\end{tabular}
\end{table}

\subsection{Implementation Settings of the Agentic Baselines (FETA and ZARA)}
\label{app:agentic-baseline-boundaries}

This subsection documents the implementations of FETA and ZARA used in the BFH and GRS evaluations in Section~\ref{sec:exp-settings}. Both use Qwen3.6-27B with deterministic decoding, frozen challenge manifests, a class-balanced reference bank, and common resource limits. Each preserves its own evidence workflow: FETA ranks channels and retrieves DTW exemplars before per-channel reasoning and confidence aggregation, whereas ZARA combines frozen-encoder retrieval with pairwise feature knowledge, candidate pruning, and decision synthesis. The comparison therefore holds these shared evaluation conditions fixed while comparing how the workflows construct and use evidence.

\paragraph{FETA: retrieval and channel-level reasoning.}
FETA ranks channels using class-separation and approximate nearest-neighbor accuracy scores computed on labeled training data. For each selected channel, it retrieves similar labeled sequences using dynamic time warping (DTW). Independent LLM reasoners compare the query with these examples and return a label and self-assessed confidence; a deterministic aggregator combines their outputs. Its training-free setting means no task-specific parameter training, not the absence of labeled reference data \citep{sui2025feta}.

\paragraph{ZARA: statistical knowledge and hierarchical reasoning.}
ZARA addresses human activity recognition from motion sensors. Offline, it constructs activity-pair feature-importance knowledge using statistical features and permutation-based ranking with AutoGluon. At inference, a frozen time-series encoder supports sensor-placement-specific, class-wise retrieval. A Feature Selector identifies coarse discriminative features; an Evidence Pruning agent removes mismatched activities; feature selection is repeated for the surviving candidates; and a Decision Insight agent produces the final label and rationale. Training-free inference therefore coexists with an offline knowledge-construction stage \citep{li2026zara}.

Both methods are evaluated on BFH and GRS (Table~\ref{tab:challenge-results-combined}) and are compared architecturally in the baseline inventory (Table~\ref{tab:baseline-overview}).

\begin{table}[H]
\caption{Implementation settings for FETA and ZARA in the BFH and GRS evaluations. The table summarizes their method-specific workflows under common evaluation conditions.}
\label{tab:agentic-baseline-boundaries}
\centering\scriptsize
\setlength{\tabcolsep}{3.2pt}
\renewcommand{\arraystretch}{1.12}
\begin{tabular}{@{}>{\raggedright\arraybackslash}p{0.105\textwidth}>{\raggedright\arraybackslash}p{0.185\textwidth}>{\raggedright\arraybackslash}p{0.235\textwidth}>{\raggedright\arraybackslash}p{0.215\textwidth}>{\raggedright\arraybackslash}p{0.205\textwidth}@{}}
\toprule
System & Method-specific workflow & Retrieval and prompting & Common configuration & Reported comparison \\
\midrule
\textbf{FETA} \citep{sui2025feta} & Channel-level reasoning: ranks channels, compares retrieved exemplars, and aggregates per-channel judgments with confidence scores. & Channel ranking from the labeled reference bank; DTW exemplar retrieval; a per-channel comparison prompt; and deterministic confidence aggregation. & Qwen3.6-27B; deterministic decoding; frozen manifests; class-balanced reference bank; common resource limits. & Evaluated on the common BFH/GRS task, split, manifests, reference data, and scoring procedure. \\
\midrule
\textbf{ZARA} \citep{li2026zara} & Feature-guided reasoning: combines pairwise feature knowledge with hierarchical selection, candidate pruning, and decision synthesis. & Frozen-encoder, class-wise retrieval; pairwise feature-importance knowledge; feature-selection and decision prompts; and candidate pruning. & Qwen3.6-27B; deterministic decoding; frozen manifests; class-balanced reference bank; common resource limits. & Evaluated on the common BFH/GRS task, split, manifests, reference data, and scoring procedure. \\
\bottomrule
\end{tabular}
\end{table}

\FloatBarrier

\FloatBarrier
\Needspace{6\baselineskip}

\section{Benchmark Construction: How the BFH and GRS Challenge Sets Were Built}
\label{app:benchmark-provenance}
\label{sec:dual-challenge-benchmarks}

Running the complete agentic workflow on every held-out \newred{sample} is materially more expensive than scoring a conventional classifier in a batched forward pass.  Let \(\mathcal D_{\mathrm{test}}=\{(x_i,y_i,d_i)\}_{i=1}^{N_{\mathrm{test}}}\) be a held-out pool, where \(d_i\) is the operating condition, and let \(q_i\) denote the number of LLM and tool calls consumed by \newred{sample} \(x_i\).  Full-pool evaluation costs
\begin{equation}
Q(\mathcal D_{\mathrm{test}})=\sum_{i=1}^{N_{\mathrm{test}}}q_i .
\label{eq:grs-full-cost}
\end{equation}
For agentic systems, reducing \(N_{\mathrm{test}}\) is therefore an evaluation-design requirement rather than a convenience.  A single reduction rule is insufficient: selecting only the samples with the highest baseline error can spend the evaluation budget on many near-duplicate examples of one failure pattern, whereas selecting only geometric centroids can omit difficult tails. We therefore use two complementary challenge benchmarks:
\begin{enumerate}[noitemsep,topsep=2pt,leftmargin=*]
\item the \emph{Baseline-Failure Hard (BFH) challenge set}, which concentrates the budget on failure \emph{severity}; and
\item the \emph{Geometry-Stratified Response-Stability (GRS) diversity set}, which spreads the budget over representative \emph{coverage} across class, condition, signal geometry, and frozen-model response regimes.
\end{enumerate}
Each uses a frozen selection registry that excludes the evaluated agent.  We write the BFH and GRS registries as \(\mathcal{F}_{\mathrm B}=\{f_m\}_{m=1}^{M_{\mathrm B}}\) and \(\mathcal{F}_{\mathrm G}=\{f_m\}_{m=1}^{M_{\mathrm G}}\), respectively.  The registry sizes and members are fixed in the released manifests.  These registries are challenge-construction devices, not ensembles or claims that every member is a competitive baseline.  What counts as hard in BFH therefore depends on which models sit in the registry: a model that is wrong almost everywhere marks almost every \newred{sample} as a failure. GRS uses frozen-model response patterns to group samples and retains one representative from every nonempty group. Its construction is driven by representative coverage across signal and response regimes.

The same-backbone non-agent streams used in the empirical comparison are members of the relevant construction registries. BFH maximizes aggregate registry error, so registry-member scores and agent--registry gaps are affected by the sampling rule. An apparent gain can therefore partly be a selection artifact, whose contribution is not isolated by these comparisons. They describe recovery on samples chosen for registry failure and cannot establish unbiased superiority on the full pool. FETA and ZARA are excluded from BFH construction, so we emphasize their matched-\newred{sample} comparisons while retaining registry scores as descriptive diagnostics. These agent comparisons remain conditional on the BFH manifest. GRS retains one representative per nonempty signal/response group to cover the regimes present in the held-out pool.

\subsection{\texorpdfstring{\newred{Challenge Set I}}{Challenge Set I} --- BFH: A Failure-Severity Stress Test}
\label{app:benchmark-bfh}
\label{sec:bfh-challenge-set}

For each held-out \newred{sample}, define the frozen-model correct count and error rate
\begin{equation}
N_i^{\mathrm{correct}}=\sum_{m=1}^{M_{\mathrm B}}\mathbf{1}\!\left[f_m(x_i)=y_i\right],
\qquad
h_i=1-\frac{N_i^{\mathrm{correct}}}{M_{\mathrm B}}.
\label{eq:bfh-hardness}
\end{equation}
Thus \(h_i=0\) means that every selection model is correct, while larger \(h_i\) means that a greater fraction of the frozen \newred{registry models} fails. For class-specific quotas \(\{n_y\}_{y\in\mathcal{Y}}\), with \(\sum_y n_y=N_{\mathrm{BFH}}\), the BFH set is
\begin{equation}
\mathcal{T}_{\mathrm{BFH}}
=
\bigcup_{y\in\mathcal{Y}}
\operatorname{Top}_{n_y}
\left\{(x_i,y_i,d_i):y_i=y,\ N_i^{\mathrm{correct}}\geq1;\ h_i\right\},
\label{eq:bfh-set}
\end{equation}
where \(\operatorname{Top}_{n_y}\{\cdot;h_i\}\) returns the \(n_y\) largest hardness values and uses a fixed seed only to break exact ties.  The current export assigns nearly equal class quotas and excludes \(N_i^{\mathrm{correct}}=0\) samples as a conservative label-quality guard.  This exclusion is a protocol choice, not proof of mislabeling; its count and sensitivity should be reported.

BFH answers a narrow question: \emph{can the agent recover samples on which a large fraction of the frozen \newred{registry models} fails?}  It is deliberately severe, but top-\(h_i\) selection may choose many samples that are hard for the same reason. Consequently, BFH can over-sample one recurring signal or model-response pattern and under-cover other classes, conditions, and presentation patterns already present in the held-out pool.

\subsection{\texorpdfstring{\newred{Challenge Set II}}{Challenge Set II} --- GRS: A Representative-Coverage Diversity Test}
\label{app:benchmark-grs}
\label{sec:grs-challenge-set}

GRS tests agent performance across the diversity of the held-out pool. Its construction represents operating conditions, signal patterns, and frozen-model response regimes through explicit grouping and representative selection.

\begin{figure*}[t]
\centering
\includegraphics[width=\textwidth]{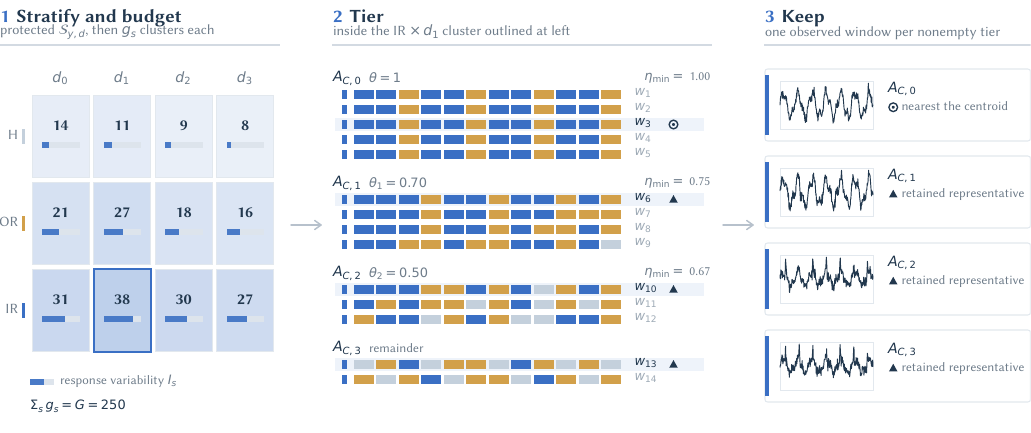}
\caption{GRS construction: (1) allocate geometry clusters within class--condition strata; (2) group frozen-model responses into stability tiers; (3) retain one observed representative per nonempty tier. Bars show response variability $I_s$; counts show cluster allocations $g_s$ (illustrative total $G{=}250$). Stability is measured within each model across samples, not between models.}
\Description{A three-stage figure showing allocation of geometry clusters within class-condition strata, grouping of frozen-model prediction patterns within a cluster, and selection of one real signal sample per response tier.}
\label{fig:grs-construction}
\end{figure*}

Sampling by class and condition ensures coverage of those categories, but does not explicitly account for variation within them. samples from the same class and condition can differ in signal structure or in the predictions they elicit from the frozen models. GRS therefore groups the held-out pool in three stages: by class and condition, by similarity of cached signal embeddings, and by similarity of frozen-model responses. It retains one observed \newred{sample} from each resulting group (Figure~\ref{fig:grs-construction}). This makes within-category variation an explicit part of benchmark construction and limits repeated representation of the same signal and response group.

\paragraph{Step 1: group similar signals within each class and condition.}
Each nonempty class--condition stratum is clustered separately using k-means on cached embeddings $\boldsymbol{\xi}_i=\phi(x_i)$. A global target $G$ controls the total number of geometry clusters. Every nonempty stratum receives at least one cluster; the remaining clusters are allocated according to stratum size and frozen-model response variability. The allocation priority combines the stratum size $n_s$ with a response-variability measure $I_s$, so larger or more response-variable strata receive finer geometric coverage. Bounded largest-remainder apportionment converts these priorities into integer cluster counts, breaking ties by class--condition key and capping each count at the stratum size. This assigns finer geometric coverage to larger or more response-variable strata while retaining every observed class--condition combination.

\paragraph{Step 2: group similar model responses within each cluster.}
A \newred{sample}'s response signature is the ordered list of predictions from the $M_{\mathrm G}$ frozen registry models. Within each geometry cluster, samples with identical signatures form initial groups. The largest such group becomes the \emph{exact tier}: each model gives a constant prediction across its samples, although different models may disagree with one another. The remaining groups are merged into broader tiers using a per-model stability threshold:
\begin{equation}
\eta_m(A)=\max_{k\in\mathcal{Y}}
\frac{|\{i\in A:f_m(x_i)=k\}|}{|A|}
\geq\theta
\quad\text{for every registry model }m.
\label{eq:grs-model-stability}
\end{equation}
Here $A$ is a candidate tier. For example, $\theta=0.70$ requires each model to repeat its own most frequent prediction on at least 70\% of the tier's samples; it does not require agreement between models. The construction processes thresholds $0.70$ and $0.50$ in order. At each threshold, it scans remaining signature groups by decreasing size, breaking ties by signature, and adds a group when the merged tier still satisfies the threshold. Any groups left after the final threshold form a remainder tier. This is a deterministic grouping rule, not an optimal clustering claim.

\paragraph{Step 3: retain one observed representative per tier.}
Every nonempty tier contributes exactly one observed \newred{sample} to the manifest. For the exact tier, the representative is the \newred{sample} nearest the tier centroid in embedding space, with ties resolved by the fixed \newred{sample} identifier. The threshold and remainder tiers also contribute one observed representative each. All retained samples come from the held-out pool, and the evaluated agent does not participate in their selection. This gives each nonempty signal/response regime a place in the benchmark.

\paragraph{Subset size.}
With an exact tier, two threshold tiers, and a possible remainder tier, each geometry cluster contributes between one and four representative samples. The geometry target controls the resolution of signal grouping, while the final subset size depends on the response groups present in the data. The released manifests use fixed geometry targets and report the resulting number of representatives for each evaluation setting (Table~\ref{tab:challenge-manifests}).

\paragraph{MCC5 healthy class.} Under the global target, the healthy class of MCC5 received too few samples in the first construction. We therefore removed its healthy samples and rebuilt the healthy class on its own with $G{=}100$, using the same steps, registry, and thresholds; the three fault classes were kept unchanged. The MCC5 manifest thus combines the three fault classes from the $G{=}250$ construction with the healthy class from the $G{=}100$ construction, which gives at most 350 geometry clusters in total and 1{,}152 samples, 233 of them healthy (Table~\ref{tab:challenge-manifests}). This change depended only on class composition, not on any evaluation score, and all methods are scored on the same final manifest.

\paragraph{Configuration and interpretation.}
The construction uses frozen embeddings, a frozen prediction registry, and fixed selection settings that do not use any evaluated agent's outputs. The artifact records the seed choices and validation checks for unique real samples, complete assignment of source samples to tiers, and one representative per nonempty tier.

GRS provides a practical test of agent performance across stratified signal and response patterns. Its groups are defined by embeddings and frozen-model predictions, so they do not guarantee identical agent behaviour within a group. Scores give equal weight to selected samples, without weighting by the number of source samples they represent; class balance is not enforced. Every nonempty tier is kept and every tier counts once, so a GRS score describes performance across the response patterns found in the held-out pool rather than a stream in which common patterns would dominate. Together, BFH and GRS test concentrated failure recovery and broader case coverage.

\paragraph{Evaluation protocol.}
BFH and GRS are evaluated using the same frozen agent configuration. BFH tests recovery on concentrated baseline failures, while GRS tests performance across the stratified signal and response patterns. The selection registry and benchmark manifests do not depend on any evaluated agent's outputs, and all compared methods are scored on the same selected samples within each challenge. Results are reported separately for the two challenges to distinguish failure recovery from broader case coverage.
\FloatBarrier

\subsection{Evaluation Scale and Class Composition}
\label{app:benchmark-budget-rationale}

\paragraph{What the evaluation covers.}
Each variant evaluated on both challenges is scored on 3121 challenge rows: 300 BFH samples (100 per dataset) and 2821 GRS samples (906 on PU, 763 on UO, 1152 on MCC5). No manifest row is chosen using the outputs of any evaluated agent, and every compared method deep-learning baselines, foundation-model baselines, FETA, ZARA, and both \method{} variants is scored on the same rows of the final manifests under the same LLM, the same labeled reference budget, and the same resource limits. The small reference budget defines the few-shot setting; it is not the test-set size. The row count assumes no disjointness between BFH and GRS.

\paragraph{Why the matched design carries the comparison.}
Every method is scored on the same frozen samples, permitting paired comparisons on each challenge. Matched samples do not remove BFH selection bias: the manifest is chosen using registry errors, including errors of the fitted heads used in the comparison. Choosing rows without agent outputs prevents agent-dependent selection, but does not make the test neutral toward registry members. FETA and ZARA do not define the manifest; their comparisons address relative performance on the same failure-focused challenge. GRS instead retains a representative from each nonempty signal/response group, so its comparison is organized around coverage of these regimes. Both challenges give descriptive margins rather than an unbiased estimate of accuracy on the full pool or in deployment; Appendix~\ref{app:run-variability} reports how far the \method{} scores move between executions.

\paragraph{Class counts follow the construction, not a sampling choice.}
Class imbalance is routine in predictive-maintenance diagnosis, where normal-operation records are abundant while some faults are rare or costly to collect \citep{wang2020imbalanced,jalayer2022imbalanced}. The GRS imbalance has a different origin. Retaining one representative per nonempty geometry/response group necessarily yields more cases for classes that occupy more groups, so the counts are a consequence of the coverage rule rather than a sampling decision taken afterwards. Forcing equal counts would require discarding selected groups or altering how they are represented---that is, removing the coverage the manifest exists to provide. The residual imbalance is bounded: every retained class has at least 175 samples, and the largest within-dataset majority-to-minority ratio is $2.38{:}1$. Macro-F1 is reported alongside accuracy so that each class contributes equally, the composite weights datasets and challenges equally rather than letting the largest manifest dominate (Section~\ref{sec:exp-settings}), and BFH is separately balanced or near-balanced. Minority-class performance is therefore visible in the headline numbers rather than absorbed by a majority class.

\paragraph{Why BFH is deliberately small.}
BFH answers one question---can the agent recover cases that many frozen registry models misclassify?---and 100 samples per dataset is the instrument for that question rather than a concession to cost. By Eq.~\ref{eq:bfh-hardness} the set takes the hardest $n_y$ samples per class by frozen-model error rate, so enlarging it would descend the hardness ranking and admit progressively easier samples, diluting the very contrast the challenge isolates. Breadth is supplied instead by GRS, which is roughly an order of magnitude larger and spans classes, operating conditions, signal geometry, and frozen-model response regimes. The two challenges are reported separately for this reason: one measures recovery on concentrated failures, the other coverage across diverse regimes, and neither substitutes for the other.

\paragraph{Precedent for bounded evaluation subsets.}
Scoring agents on a specified subset rather than every available case is standard where per-case inference cost is high: SWE-bench Lite scores 300 of 2294 tasks \citep{jimenez2024swebenchlite}, and WebRL evaluates 165 of WebArena's 812 tasks, explicitly citing testing cost \citep{qi2025webrl}. We cite these as established practice, not as license, and we claim no statistical equivalence between software tasks and sensor samples. The case here rests on frozen manifests, matched evaluation across three datasets, and two challenges that isolate complementary failure modes.

\FloatBarrier
\Needspace{6\baselineskip}

\section{Summary of Notation for the Method and Inference Path}
\label{app:method-notation}

Table~\ref{tab:method-notation} collects the symbols used in the protocol and \newred{sample-specific} inference path. It includes only notation needed to read the main equations; symbols used only in detailed appendix constructions are defined locally. Benchmark-construction notation for BFH and GRS is likewise defined in Section~\ref{sec:dual-challenge-benchmarks}.

\raggedbottom
\begingroup
\let\vss\vfil
\footnotesize
\setlength{\tabcolsep}{5pt}
\renewcommand{\arraystretch}{1.02}
\begin{longtable}{@{}>{\raggedright\arraybackslash}p{0.23\linewidth}>{\raggedright\arraybackslash}p{0.70\linewidth}@{}}
\caption{Summary of main notation for the \method{} protocol and inference path.} \label{tab:method-notation}\\
\toprule Symbol & Meaning \\ \midrule
\endfirsthead
\toprule Symbol & Meaning \\ \midrule
\endhead
\bottomrule
\endfoot \multicolumn{2}{@{}l}{\textit{Inputs and reference retrieval}} \\
$x^\star$ & Query multivariate time-series \newred{sample} supplied for diagnosis. \\
$x^{\mathrm{prev}}$ & Immediate predecessor of the query \newred{sample}, available only through one-step temporal context. \\
$T,C$ & Numbers of \newred{observations} and sensor channels in one \newred{sample}, respectively. \\
$\mathcal Y,\newred{Y},y$ & Diagnosis-label set, its number of classes, and a generic candidate class. \\
$\mathcal D_{\mathrm{ref}},N$ & Labeled training set (the reference pool) and its number of samples. \\
$\mathcal K,F$ & Read-only knowledge context (operating condition of the input: machine type, rotational speed, and recorded channels), and the complete inference procedure of Sections~\ref{sec:stage2} and~\ref{sec:adjudication}. \\
$\mathcal R_y,\mathcal R^\star,r_y$ & Training samples labeled $y$ in $\mathcal D_{\mathrm{ref}}$, and the class-balanced reference bank $\mathcal R^\star=\{r_0,\ldots,r_{\newred{Y}-1}\}$, where $r_y\in\mathcal R_y$ is the retrieved class-$y$ reference (Section~\newred{\ref{sec:problem}}). \\
\newred{$\mathcal R^\star(q)$} & \newred{Reference bank of Eq.~\ref{eq:ret} for a sample $q$: for each class $y$, the class-$y$ training sample most similar to $q$, excluding $q$ itself; $\mathcal R^\star=\mathcal R^\star(x^\star)$.} \\
$\phi_{\mathrm{emb}},s_{\mathrm{cos}}$ & Frozen retrieval embedding and cosine-similarity score. \\
\midrule
\multicolumn{2}{@{}l}{\textit{Request gate and governed follow-up}} \\
$\mathcal E_t,\mathcal H_t$ & \agent{Judge}-visible evidence ledger of packets returned before request round $t$, updated as $\mathcal E_{t+1}=\mathcal E_t\cup\{p_\tau\}_{\tau\in\mathcal T_t}$; $\mathcal E$ denotes the ledger at finalization\newred{. $\mathcal H_t$ is the} \host{}-fixed live candidate set at round $t$, \newred{built from the sources of $\mathcal H_1$ and the ledger $\mathcal E_t$}. \\
$\Phi_t,\Phi_{\mathrm{reg}},\Phi(\mathcal E_t)$ & Requested properties, registered property vocabulary, and properties already \newred{measured in the ledger (closed)}. \\
\newred{$\mathcal S_t,\mathcal T_t$} & \newred{ToolCards offered for the declared properties, and the batch the \agent{Judge} commits to (Section~\ref{sec:request-gate}).} \\
$\tau,p_\tau$ & One selected immutable ToolCard and the measurement packet it returns. \\
\newred{$\mathcal P_1,\mathcal P_2,\mathcal P_3$} & \newred{Channel priority tiers after the \host{} guards: \texttt{high}, \texttt{medium}, and \texttt{low}.} A removed channel is in no tier (Section~\ref{sec:stage2}). \\
$\mathcal P^\star$ & \newred{The highest nonempty tier; only its channels decide the winning class (Section~\ref{sec:request-gate}).} \\
\newred{$\Phi(\tau)$} & \newred{Properties measured by ToolCard $\tau$.} \\
\midrule
\multicolumn{2}{@{}l}{\textit{Reference preflight and \newred{finalization}}} \\
\newred{$r_y^{\mathrm{ctrl}}$} & Labeled class-\newred{$y$} pseudo-query (control), equal to the retrieved reference \newred{$r_y$}. \\
$\mathcal R^{\mathrm{alt}}(q)$ & \newred{Alternate bank for a control $q$: $\mathcal R^{\mathrm{alt}}(q)=\mathcal R^\star(q)$, which never contains $q$ itself.} \\
$\hat y_\tau(\cdot),g_{\mathrm{ref}}(\tau)$ & \newred{Winning class of} ToolCard $\tau$ (\newred{Section~\ref{sec:request-gate}}): $\hat y_\tau(x^\star;\mathcal R^\star)$ on the query and \newred{$\hat y_\tau(r_y^{\mathrm{ctrl}};\mathcal R^{\mathrm{alt}}(r_y^{\mathrm{ctrl}}))$} on a control ($\bot$ if none; Appendix~\ref{app:host-packet-gate}), and the \newred{reference preflight, which requires} correct recovery of all $\newred{Y}$ classes. \\
$\tilde y$ & Admitted \agent{Judge} proposal; $\tilde y=\bot$ if none was admitted within $N_J$ steps (Section~\ref{sec:finalization}). \\
$N_J$ & \agent{Judge} step budget per \newred{sample} ($N_J=10$; Table~\ref{tab:agent-config}). \\
$b_1,b_2$ & Fitted-model \newred{predictions} given to the \agent{Judge}. \\
$\hat y$ & Class the protocol reports for the \newred{sample}: $\tilde y$ if admitted, otherwise the \agent{Judge}'s last-step class in $\mathcal H_1$, marked as not admitted. \\
$\mathcal A$ & \newred{Record of all actions and results}, published with $\hat y$. \\
\end{longtable}
\endgroup
\raggedbottom

\FloatBarrier
\Needspace{6\baselineskip}

\section{\texorpdfstring{Agentic4PDM Inference Protocol: Retrieval, \newred{Gates, and Finalization}}{Agentic4PDM Inference Protocol: Retrieval, Gates, and Finalization}}
\label{app:method-details}

\subsection{Retrieval Protocol: Class-Balanced Reference Selection}
\label{app:retrieval-details}
This subsection formalizes the class-balanced retrieval procedure summarized in Section~\newred{\ref{sec:problem}}. Retrieval ranks over the entire labeled reference pool of each class; no operating-condition filter is applied, and the labeled bank is assumed to cover every retained class, $\mathcal R_y\neq\emptyset$ for all $y\in\mathcal Y$. A mean-pooled Moirai~1.1~R~Large embedding maps the query and the candidates to a shared representation, and a per-dimension scaler fitted on eligible training embeddings is applied to both. The system directly ranks each class pool by cosine similarity and retains its $k$ highest-scoring references:
\begin{equation}
\begin{split}
\mathcal{R}^\star_y &=
\operatorname{TopK}_{r\in\mathcal R_y}\!
\left(s_{\mathrm{cos}}(x^\star,r),k\right),\\
s_{\mathrm{cos}}(x,r) &=
\frac{\phi_{\mathrm{emb}}(x)\cdot\phi_{\mathrm{emb}}(r)}
{\lVert\phi_{\mathrm{emb}}(x)\rVert\,
\lVert\phi_{\mathrm{emb}}(r)\rVert}.
\end{split}
\label{eq:stage2-retrieval}
\end{equation}
Selecting one pool per class keeps the \newred{reference} bank class-balanced by construction. Because ranking is by embedding cosine similarity alone, a retrieved reference is the highest-scoring available exemplar of its class but is not guaranteed to share the query's operating condition. The \newred{reference} bank $\mathcal R^\star=\{\mathcal R^\star_y\}_{y\in\mathcal Y}$, which with $k=1$ is $\{r_y\}_{y\in\mathcal Y}$ as in Section~\ref{sec:problem}, is then used for downstream preprocessing, segmentation, and visual comparison. The reported configuration retains the single highest-scoring reference per class; in that case Eq.~\ref{eq:stage2-retrieval} \newred{gives the reference bank $\mathcal R^\star(x^\star)$ of Eq.~\ref{eq:ret}}, which the reference preflight reuses with a control as the query. This retrieval step does not fit or apply a KNN classifier and does not aggregate neighbor labels.

\FloatBarrier
\Needspace{6\baselineskip}

\subsection{Request Gate: Property Registry}
\label{app:why-request-gate}
\label{app:host-request-gate}

\makeatletter
\providecommand{\gate@breakunderscore}{\char`\_\penalty100\hskip0pt\relax}
\providecommand{\gatereason}[1]{{\ttfamily\scriptsize
  \let\_\gate@breakunderscore #1}}
\providecommand{\gateid}[1]{{\ttfamily\let\_\gate@breakunderscore #1}}
\makeatother

\newred{Section~\ref{sec:request-gate} defines the request gate, and Appendix~\ref{app:workflow-request-gate} shows one complete round.}

\Needspace{18\baselineskip}
\paragraph{Property registry.}
Table~\ref{tab:request-gate-property-registry} lists the 48 registered
properties. Each describes an aspect of the signal that may remain uncertain,
is shared across datasets, and never refers to a class. \newred{The \host{} maps each property to the ToolCards that measure it; a property therefore only determines which cards may be offered, and neither reveals scores nor chooses a card on the \agent{Judge}'s behalf.}

{\small
\setlength{\tabcolsep}{5pt}
\renewcommand{\arraystretch}{1.08}
\begin{longtable}{@{}>{\centering\arraybackslash}p{0.04\linewidth}
                   >{\raggedright\arraybackslash}p{0.33\linewidth}
                   >{\raggedright\arraybackslash}p{0.57\linewidth}@{}}
\caption{Registered signal properties. Each property names an aspect of the
signal that may remain unresolved; none refers to a class.}
\label{tab:request-gate-property-registry}\\
\toprule
\# & Property & What it describes \\
\midrule
\endfirsthead
\toprule
\# & Property & What it describes \\
\midrule
\endhead
\bottomrule
\endfoot
1 & \path{amplitude_distribution_match} & How the signal's amplitude values are spread, i.e., the shape of their histogram. \\
2 & \path{severity_or_offset_distribution_shift} & Whether the amplitude values are shifted, widened, or offset as a whole. \\
3 & \path{global_summary_match} & Whole-\newred{sample} summary statistics such as level, spread, and peakiness. \\
4 & \path{global_severity_or_impulsiveness} & How strong and how spiky the signal is over the whole \newred{sample}. \\
5 & \path{local_ordering_dynamics} & The order in which neighbouring values rise and fall. \\
6 & \path{scale_invariant_local_shape} & Local rise-and-fall patterns, regardless of amplitude and time scale. \\
7 & \path{local_energy_bursts} & Short bursts of energy that stand out from their surroundings. \\
8 & \path{impulsive_energy_distribution} & How impulsive energy is spread across the \newred{sample}. \\
9 & \path{short_memory_linear_dynamics} & How well each value can be predicted from the few values just before it. \\
10 & \path{innovation_distribution} & The spread of what that short-term prediction fails to explain. \\
11 & \path{unexplained_residual_spectrum} & The frequency content of the part the short-term prediction misses. \\
12 & \path{roughness_across_scales} & How rough or irregular the signal looks at coarser and finer scales. \\
13 & \path{scale_dependent_structure} & Structure that appears only at particular time scales. \\
14 & \path{harmonic_spacing} & The spacing between repeated, evenly spaced frequency peaks. \\
15 & \path{modulation_spacing} & How often the slow amplitude envelope repeats. \\
16 & \path{recurrence_match} & How the signal repeats itself across time lags. \\
17 & \path{periodic_structure} & Whether the signal contains regular periodic patterns, and at what period. \\
18 & \path{state_recurrence} & How often the signal returns to states it has visited before. \\
19 & \path{thresholded_recurrence_profile} & Returns to earlier states, counted only when they come within a fixed distance. \\
20 & \path{global_spectral_shape} & The overall shape of the frequency spectrum. \\
21 & \path{broad_band_energy_distribution} & How energy is shared among wide frequency bands. \\
22 & \path{nonstationary_profile} & How the signal's behaviour changes over the course of the \newred{sample}. \\
23 & \path{temporal_severity_evolution} & How signal strength changes over time within the \newred{sample}. \\
24 & \path{temporal_impulsiveness_evolution} & How spikiness changes over time within the \newred{sample}. \\
25 & \path{multiscale_transient_energy} & Energy of short transients measured at several time scales. \\
26 & \path{dyadic_morphology} & Signal shape viewed at successively halved time scales. \\
27 & \path{amplitude_modulation_shape} & The shape of the slow amplitude variation (envelope) carried by the signal. \\
28 & \path{burst_repetition_spectrum} & How regularly bursts repeat, seen as peaks in the envelope spectrum. \\
29 & \path{state_persistence} & How long the signal stays above or below a level before switching. \\
30 & \path{run_length_distribution} & The spread of those stay durations. \\
31 & \path{complexity_across_scales} & How complex or unpredictable the signal is at several time scales. \\
32 & \path{nonlinear_dynamics} & Irregular patterns that a simple linear model does not capture. \\
33 & \path{symbolic_state_occupancy} & How much time the signal spends in each coarse amplitude level. \\
34 & \path{symbolic_state_transitions} & How the signal moves between coarse amplitude levels. \\
35 & \path{modulation_dynamics} & How the signal's instantaneous frequency and envelope vary over time. \\
36 & \path{instantaneous_frequency_distribution} & The spread of the signal's instantaneous frequency. \\
37 & \path{point_aligned_waveform} & The waveform shape around aligned reference points such as peaks. \\
38 & \path{point_aligned_derivative_shape} & The rate of change around those aligned points. \\
39 & \path{normalized_spectral_peaks} & The positions and relative heights of the spectral peaks. \\
40 & \path{dominant_peak_morphology} & The height, width, and sharpness of the strongest spectral peak. \\
41 & \path{time_localized_spectrum} & Frequency content measured over short time segments. \\
42 & \path{spectral_event_timing} & When particular frequency events occur within the \newred{sample}. \\
43 & \path{extrema_morphology} & The shape and spacing of local peaks and troughs. \\
44 & \path{oscillation_and_transient_profile} & The balance between steady oscillation and short transients. \\
45 & \path{scale_persistent_morphology} & Shape features that survive as the signal is progressively smoothed. \\
46 & \path{smoothed_shape} & The signal's shape after fine detail is smoothed away. \\
47 & \path{detrended_shape} & The signal's shape after slow drift is removed. \\
48 & \path{local_morphology_without_drift} & Local peak-and-trough shape after slow drift is removed. \\
\end{longtable}
}

\FloatBarrier

\subsection{Packet and Influence Gate: From Measurement to Influence Role}
\label{app:host-packet-gate}

\newred{Section~\ref{par:evidence-admission} defines the packet and influence gate. This appendix shows its two steps on actual packets and adds the points that the main text leaves out.} The complete exchange for one round appears in Appendix~\ref{app:workflow-packet-influence-gate}.

Each step is illustrated with examples from the archived \newred{PU} run in
Appendix~\ref{app:agent-workflow-trace}, a three-class problem with
classes~0, 1, and~2. \newred{Samples are identified by their number}. The gate
processes each round in two steps, using only \host{}-owned facts; consequently,
\agent{Judge} text cannot change a packet role. \newred{Step~1 follows the Execute step of Section~\ref{sec:request-gate}, and Step~2 follows the reference preflight of Section~\ref{par:evidence-admission}.}

\paragraph{Step 1: Is the result usable?}
\newred{As in Section~\ref{sec:request-gate}, a packet is \emph{excluded} if the tool fails or the channels of the highest nonempty tier $\mathcal P^\star$ do not agree on a winning class, and \emph{usable} otherwise.} An excluded packet
cannot support any class. The packet nonetheless stays in the \newred{ledger} and must still be
covered by the \agent{Judge}'s round report (Section~\ref{par:evidence-admission}), so an unhelpful result cannot disappear.

\textit{Example.} Table~\ref{tab:ex-step1} shows the three packets of round~1
for \newred{sample}~107665. Its channel priorities rank \texttt{vibration\_1} and
\texttt{phase\_current\_1} as high and \texttt{phase\_current\_2} as medium.
For the fractal-scaling and local-extrema packets, both high-priority channels
name the same class, so that class becomes the \newred{winning class} and the
disagreeing medium channel is not consulted. For the autocorrelation packet,
the two high-priority channels name Classes~0 and~2; the medium channel cannot
break this disagreement, so the packet has no \newred{winning class} and is excluded.

\begin{table}[H]
\caption{Step~1 on the round-1 packets of \newred{sample}~107665: per-channel winners
and the resulting \newred{winning class}.}
\label{tab:ex-step1}
\centering\small
\renewcommand{\arraystretch}{1.15}
\begin{tabular}{@{}lcccll@{}}
\toprule
\newred{ToolCard} & \makecell{\texttt{vibration\_1}\\(high)} &
\makecell{\texttt{phase\_current\_1}\\(high)} &
\makecell{\texttt{phase\_current\_2}\\(medium)} & \newred{Winning class} & Step~1 \\
\midrule
Fractal scaling & 0 & 0 & 2 & Class~0 & Usable \\
Local extrema & 2 & 2 & 0 & Class~2 & Usable \\
Autocorrelation & 0 & 2 & 1 & None & Excluded \\
\bottomrule
\end{tabular}
\end{table}

\paragraph{Step 2: What may this result support on its own?}
\newred{Every usable packet then receives its influence role by Eq.~\ref{eq:packet-role}: corroborative if its tool passes the reference preflight, and advisory otherwise. An excluded packet gets no role, whatever its preflight result.}

\newred{The preflight $g_{\mathrm{ref}}(\tau)$ of Eq.~\ref{eq:reference-sanity}, shown as ``Preflight $g_{\mathrm{ref}}$'' in Figure~\ref{fig:stage2-inference}, is set to~0 by a wrong winning class, no winning class, or a failed run on any control.} The controls cover every class in $\mathcal Y$
rather than only the live candidates, so a tool cannot pass by
never having to recognize a class that the current round has set aside.
The selectable default quantitative tools are listed in Table~\ref{tab:quantitative-tool-catalog},
and Appendix~\ref{app:workflow-packet-influence-gate} shows the preflight record for an
actual round.

\textit{Example.} Table~\ref{tab:ex-step2} shows two preflight records. For
\newred{sample}~106619, the autocorrelation ToolCard recovers every control, so
$g_{\mathrm{ref}}(\tau)=1$ and its packet is corroborative. For
\newred{sample}~107665, the fractal-scaling ToolCard recovers the Class~0 control, but
on the Class~1 and Class~2 controls it names no class under the
channel-priority rule of Step~1; $g_{\mathrm{ref}}(\tau)=0$ and its packet is advisory. The
preflight is repeated for every \newred{sample} because it uses that \newred{sample}'s
references and channel priorities, so the same ToolCard can pass for one
\newred{sample} and fail for another.

\begin{table}[H]
\caption{Reference preflight records: the class named by the ToolCard for
each control, and the resulting role.}
\label{tab:ex-step2}
\centering\small
\renewcommand{\arraystretch}{1.15}
\begin{tabular}{@{}llccccl@{}}
\toprule
\newred{Sample} & \newred{ToolCard} & Class-0 control & Class-1 control & Class-2 control
& $g_{\mathrm{ref}}(\tau)$ & Role \\
\midrule
106619 & Autocorrelation & 0 & 1 & 2 & 1 & Corroborative \\
107665 & Fractal scaling & 0 & none & none & 0 & Advisory \\
\bottomrule
\end{tabular}
\end{table}

\FloatBarrier

\subsection{Publication: What the \host{} Serializes and Reports}
\label{app:finalization-details}

The \host{} publishes $\hat y=\tilde y$ when a proposal is admitted; otherwise it publishes the \agent{Judge}'s last-step class, which lies in $\mathcal H_1$. It serializes the \newred{admission status} (admitted or not admitted) and the record with $\hat y$, so a class that was not admitted is distinguishable from an admitted decision.

\FloatBarrier
\Needspace{6\baselineskip}

\section{Tool Registries: What the Agents May Execute}
\label{app:tool-registries}

The quantitative-tool catalog in Table~\ref{tab:quantitative-tool-catalog} lists the 25 default ToolCards exposed by the production selector, with their exact identifiers and fixed parameter settings. Input capabilities, requested properties, question scope, and budget may further restrict the eligible set for a given sample.

\paragraph{How the quantitative tools are authored.}
The tools in Table~\ref{tab:quantitative-tool-catalog} are written once, offline, by a stronger model (GPT-5.6) rather than by the inference-time agent. We separate the two roles because writing a numerical measurement that is correct and robust is a different task from choosing among measurements: the smaller model that runs \method{} (Qwen3.6-27B) only selects from the fixed registry and never writes code. Authoring is property-first. GPT-5.6 receives labeled training samples of PU only, retrieves non-overlapping samples from each class, and inspects them both as raw series and as class-overlaid figures. From what it observes, it first names signal properties that appear to vary (for example periodic recurrence, burst repetition, or amplitude distribution; Table~\ref{tab:request-gate-property-registry}) and only then writes a quantitative tool that measures each property. A tool is therefore defined by what it measures, not by how well it separates the training classes. We choose this over accuracy-driven selection because a tool that separates the training classes may still not measure the property needed to resolve a particular ambiguity, which is the failure mode that motivates \method{}. No sample of UO, MCC5, or Bel, and no test sample of any dataset, was shown to GPT-5.6, so on these three datasets the tools are applied to data they were not written for. The description above is sufficient to reproduce the results. Authoring does not certify a tool for use. A registered card contains no class identity, label, or performance statistic, and at inference the \host{} treats a tool's result as checked evidence only if the tool recovers the known class of every reference control for that sample (the reference preflight in Eq.~\ref{eq:reference-sanity}; Appendix~\ref{app:host-packet-gate}).

\subsection{General-Purpose Inspection Tools}
\label{app:general-tools}

Table~\ref{tab:general-tools} records every \host{}-facing action the \newred{\agent{Judge}} may request. The \host{} checks syntax, scope, authorization, and quantitative bindings before dispatch.

\begin{table}[H]
\caption{All seven general tool actions available to the \newred{\agent{Judge}}.}
\label{tab:general-tools}
\centering\footnotesize
\renewcommand{\arraystretch}{1.10}
\setlength{\tabcolsep}{3pt}
\begin{tabular}{@{}>{\raggedright\arraybackslash}p{0.32\linewidth}>{\raggedright\arraybackslash}p{0.60\linewidth}@{}}
\toprule
Action & Purpose and constrained request interface \\
\midrule
\makecell[l]{\texttt{compare\_normal}\\\texttt{compare\_abnormal}} & Re-check the normal-region overlay or the abnormal pairwise evidence. Request \texttt{task\_objective} and \texttt{judge\_feedback}; the abnormal action may additionally bound \texttt{target\_classes}. \\
\texttt{view\_image} & Load one previously supplied figure for next-turn close inspection. Request \texttt{image\_path}. \\
\texttt{run\_sandbox} & Optional same-transform, full-sample comparative measurement. Request \texttt{task\_objective}, \texttt{sandbox\_request}, and \texttt{metric\_spec}; configurations that isolate the registered quantitative library disable this action. \\
\texttt{select\_quantitative\_tools} & \newred{Declare the unresolved properties $\Phi_t$; the \host{} returns the offer $\mathcal S_t$ from the 25 default ToolCards (Declare and Offer, Section~\ref{sec:request-gate}).} \\
\texttt{run\_quantitative\_tools} & \newred{Commit a batch $\mathcal T_t$ from the offer; the \host{} checks the request conditions and then runs it (Commit and Execute, Section~\ref{sec:request-gate}).} \\
\texttt{finalize} & Propose one terminal class decision. It must be standalone and include the \host{}-required fault-screen contract. \\
\bottomrule
\end{tabular}
\end{table}

\begin{promptlisting}{\host{}-facing tool-call format (runtime template)}
STRICT JSON REQUEST
{
  "reasoning": "<live confusion, discriminating question, and observable>",
  "tool_calls": [
    {
      "action": "<one valid action from the tool interface>",
      "task_objective": "<bounded question>"
    }
  ]
}

ACTION-SPECIFIC FIELDS
Comparator actions additionally use "judge_feedback"; "compare_abnormal" may include "target_classes". "view_image" requires "image_path". "run_sandbox" requires "sandbox_request" and "metric_spec". "select_quantitative_tools" supplies a "question_spec"; "run_quantitative_tools" uses the host-returned selection identifier and precommitted batch; and "finalize" carries the required decision and evidence contracts.

CONSTRAINT
Never combine "finalize" with a non-final action. The host validates syntax, authorization, scope, and the registered quantitative bindings before execution.
\end{promptlisting}

\FloatBarrier
\Needspace{6\baselineskip}

\subsection{Smoothing Families and \host{}-Fixed Parameters}
\label{app:smoothing-tools}

The \agent{SAgent} chooses exactly one label-blind family per channel from the fixed registry in Table~\ref{tab:smoothing-tools}. The \host{} tunes the reviewed parameter grid and does not accept an LLM-proposed parameter value.

\begin{table}[H]
\caption{Complete Stage-2 smoothing-family registry.}
\label{tab:smoothing-tools}
\centering\footnotesize
\renewcommand{\arraystretch}{1.10}
\setlength{\tabcolsep}{3pt}
\begin{tabular}{@{}p{0.32\linewidth}p{0.60\linewidth}@{}}
\toprule
Family & Registered purpose \\
\midrule
\texttt{savgol} & Light polynomial denoising that preserves local waveform shape. \\
\texttt{median} & Suppress isolated impulses without averaging their amplitude into neighbors. \\
\texttt{ewm} & Track a smooth causal-style local level with bounded lag. \\
\texttt{median\_mean\_ewm} & Extract a slow robust trend from mixed impulsive and broadband noise. \\
\texttt{lowpass} & Remove high-frequency content while avoiding phase shift. \\
\bottomrule
\end{tabular}
\end{table}

\subsection{\texorpdfstring{\newred{Channel Priority Tiers} and \host{} Guards}{Channel Priority Tiers and Host Guards}}
\label{app:channel-selection}
The \agent{CSAgent} reads the smoothed, segmented views of all channels and assigns each channel one of the four \newred{options} in Table~\ref{tab:channel-levels}\newred{: a priority tier (\texttt{high}, \texttt{medium}, or \texttt{low}) or removal}. When uncertain, it keeps a channel at \texttt{medium}. It may remove a channel only if the channel is a uniform oscillation with no visible modulation or deviation and adds no context against the references.

\begin{table}[H]
\caption{Channel \newred{priority tiers} assigned by the \agent{CSAgent}.}
\label{tab:channel-levels}
\centering\footnotesize
\renewcommand{\arraystretch}{1.10}
\setlength{\tabcolsep}{3pt}
\begin{tabular}{@{}p{0.16\linewidth}p{0.76\linewidth}@{}}
\toprule
\newred{Tier} & Assigned when \\
\midrule
\texttt{high} & The channel is a primary discriminator: it shows transient bursts or impulses, a clear envelope change, a region marked as deviant, or class-separating structure whose removal would change the class ranking. \\
\texttt{medium} & The channel is useful but secondary: it gives operating context, such as load or phase-current behavior, or a weak deviation that supports but cannot settle the class. This is the default \newred{tier} for a kept channel. \\
\texttt{low} & The channel carries weak but real class information and serves only as tie-break support. \\
removed & The channel shows no visible class information. \\
\bottomrule
\end{tabular}
\end{table}

Two deterministic \host{} guards then act on the agent's output, and the agent's original selection is kept beside the corrected one in $\mathcal A$. First, if the agent keeps a single channel while two or more are available, the \host{} restores all channels. Second, the \host{} scores each channel by the selection score of its strongest diagnostic deviant region in the segmentation. Let $s_{\max}$ be the highest channel score. Every channel scoring at least $\max(2.5,\,0.7\,s_{\max})$ is restored if it was removed, and it is set to \texttt{high} if it attains $s_{\max}$ or its deviant region is a burst train; otherwise it keeps its \newred{tier}, with \texttt{medium} for a restored channel. The agent must keep at least one channel. After the guards, every kept channel has exactly one \newred{tier}, and the \newred{tiers} \texttt{high}, \texttt{medium}, and \texttt{low} \newred{are} the priority tiers \newred{$\mathcal P_1$, $\mathcal P_2$, and $\mathcal P_3$} of Section~\ref{sec:stage2}, in this order.

\subsection{Quantitative Tools Available to the \agent{Judge}}

\label{app:selectable-quantitative-tools}
Only the default configuration of each family can enter the production shortlist. Table~\ref{tab:quantitative-tool-catalog} lists all 25 such cards, including defaults not invoked in the archived runs. Availability is determined by the selector and the request constraints, rather than invocation frequency. Each card has fixed parameters; the agent selects a card without proposing alternative parameter values. The residual-level card is restricted to fault-existence questions.

\raggedbottom
\begingroup
\footnotesize
\setlength{\tabcolsep}{3pt}
\renewcommand{\arraystretch}{1.15}
\begin{longtable}{@{}>{\raggedright\arraybackslash}p{0.04\linewidth}>{\raggedright\arraybackslash}p{0.37\linewidth}>{\raggedright\arraybackslash}p{0.24\linewidth}>{\raggedright\arraybackslash}p{0.29\linewidth}@{}}
\caption{The 25 default quantitative ToolCards available to the agent. Each entry gives its exact identifier, fixed configuration, and measurement purpose. Eligibility remains subject to the input, requested properties, question scope, and budget.}
\label{tab:quantitative-tool-catalog}\\
\toprule
ID & Selectable ToolCard & Fixed configuration & Measurement purpose \\
\midrule
\endfirsthead
\multicolumn{4}{@{}l}{\footnotesize\tablename~\thetable{} -- continued}\\
\toprule
ID & Selectable ToolCard & Fixed configuration & Measurement purpose \\
\midrule
\endhead
\midrule
\multicolumn{4}{r@{}}{\footnotesize Continued on next page}\\
\endfoot
\bottomrule
\endlastfoot
T01 & \nolinkurl{quant_autocorrelation_lags_128_pearson} & \texttt{max\_\allowbreak lag=128; similarity=pearson} & Compare periodic recurrence across lags without requiring cycle-phase alignment. \\
\addlinespace[3pt]
T02 & \nolinkurl{quant_autoregressive_coefficients_order_8_l1} & \texttt{order=8; similarity=l1} & Compare short-memory linear dynamics using regularized autoregressive coefficients. \\
\addlinespace[3pt]
T03 & \nolinkurl{quant_cepstral_shape_quefrency_128_pearson} & \texttt{max\_\allowbreak quefrency=128; similarity=pearson} & Compare repeated spectral spacing over normalized quefrency bins. \\
\addlinespace[3pt]
T04 & \nolinkurl{quant_crossing_run_statistics_q50_js} & \texttt{similarity=js; threshold\_\allowbreak quantile=0.5} & Compare threshold crossings, run lengths, and state persistence. \\
\addlinespace[3pt]
T05 & \nolinkurl{quant_detrended_morphology_linear_residual_pearson} & \texttt{representation=linear\_\allowbreak residual; similarity=pearson} & Compare local residual shape after removing a declared slow trend. \\
\addlinespace[3pt]
T06 & \nolinkurl{quant_distribution_ks_raw} & \texttt{kind=ks; standardize=false} & Compare complete-sample amplitude distributions without requiring point alignment. \\
\addlinespace[3pt]
T07 & \nolinkurl{quant_envelope_spectrum_log_cosine} & \texttt{domain=envelope; kind=periodogram; similarity=cosine; transform=log} & Compare modulation or burst-repetition spectrum over normalized bins. \\
\addlinespace[3pt]
T08 & \nolinkurl{quant_fractal_scaling_lags_128_pearson} & \texttt{max\_\allowbreak lag=128; similarity=pearson} & Compare how signal roughness changes across lag scales. \\
\addlinespace[3pt]
T09 & \nolinkurl{quant_gaussian_scale_space_sigma_8_pearson} & \texttt{sigma=8; similarity=pearson} & Compare morphology after a declared smoothing scale. \\
\addlinespace[3pt]
T10 & \nolinkurl{quant_haar_multiscale_level_3_energy_l1} & \texttt{distance=l1; level=3; summary=energy} & Compare Haar detail-coefficient energy across dyadic scales. \\
\addlinespace[3pt]
T11 & \nolinkurl{quant_instantaneous_frequency_sigma_2_histogram_js} & \texttt{comparison=histogram\_\allowbreak js; sigma=2} & Compare histograms of instantaneous frequency after fixed smoothing. \\
\addlinespace[3pt]
T12 & \nolinkurl{quant_local_extrema_profile_windows_32_l1} & \texttt{distance=l1; window\_\allowbreak count=32} & Compare peak/trough density, prominence, and spacing over time. \\
\addlinespace[3pt]
T13 & \nolinkurl{quant_multiscale_complexity_scales_16_l1} & \texttt{distance=l1; max\_\allowbreak scale=16} & Compare nonlinear complexity under progressive coarse graining. \\
\addlinespace[3pt]
T14 & \nolinkurl{quant_ordinal_patterns_order_4_js} & \texttt{order=4; similarity=js} & Compare local up/down ordering dynamics independently of amplitude scale. \\
\addlinespace[3pt]
T15 & \nolinkurl{quant_prediction_residual_order_8_ks} & \texttt{comparison=ks; order=8} & Compare innovations left after a short autoregressive model. \\
\addlinespace[3pt]
T16 & \nolinkurl{quant_recurrence_profile_threshold_0_5_pearson} & \texttt{max\_\allowbreak lag=128; similarity=pearson; threshold=0.5} & Compare lag-dependent recurrence rates after scale normalization. \\
\addlinespace[3pt]
T17 & \nolinkurl{quant_residual_level_carrier_removed_rms_band_2} & \texttt{band\_\allowbreak bins=2; measure=carrier\_\allowbreak removed\_\allowbreak rms} & Compare residual signal level after removing the carrier; fault-existence scope only. \\
\addlinespace[3pt]
T18 & \nolinkurl{quant_sax_symbolic_dynamics_alphabet_6_transition} & \texttt{alphabet\_\allowbreak size=6; representation=transition} & Compare transitions between symbolic amplitude states. \\
\addlinespace[3pt]
T19 & \nolinkurl{quant_scalar_features_combined_l1} & \texttt{distance=l1; feature\_\allowbreak set=combined} & Compare explainable complete-sample robust, impulsive, and spectral summaries. \\
\addlinespace[3pt]
T20 & \nolinkurl{quant_spectral_peak_morphology_top_16_l1} & \texttt{peak\_\allowbreak count=16; similarity=l1} & Compare normalized locations and heights of dominant spectral peaks. \\
\addlinespace[3pt]
T21 & \nolinkurl{quant_spectrum_shape_log_js} & \texttt{domain=signal; kind=periodogram; similarity=js; transform=log} & Compare normalized log-periodogram shape over the complete sample. \\
\addlinespace[3pt]
T22 & \nolinkurl{quant_stft_time_frequency_windows_16_cosine} & \texttt{similarity=cosine; window\_\allowbreak count=16} & Compare time-localized spectral-energy maps. \\
\addlinespace[3pt]
T23 & \nolinkurl{quant_teager_energy_sigma_2_histogram_js} & \texttt{comparison=histogram\_\allowbreak js; sigma=2} & Compare local oscillatory energy and transient-burst distributions. \\
\addlinespace[3pt]
T24 & \nolinkurl{quant_waveform_shape_raw_pearson} & \texttt{representation=raw; similarity=pearson} & Compare raw waveform shape at matching sample positions. \\
\addlinespace[3pt]
T25 & \nolinkurl{quant_windowed_profile_windows_16_amplitude_l1} & \texttt{distance=l1; feature\_\allowbreak set=amplitude; window\_\allowbreak count=16} & Compare amplitude-summary profiles across 16 consecutive windows. \\
\addlinespace[3pt]
\end{longtable}
\endgroup

\FloatBarrier
\Needspace{6\baselineskip}
\raggedbottom

\FloatBarrier
\Needspace{6\baselineskip}

\raggedbottom

\subsection{Representative Quantitative Tool Pseudocode}
\label{app:quantitative-tool-examples}

\subsubsection{Tool 1: Threshold-Crossing Run Statistics (T04)}
\label{app:tool-example-crossing-runs}
\nolinkurl{quant_crossing_run_statistics_q50_js}

This tool turns each signal into a binary sequence using its own median, then measures the lengths of consecutive runs in each state. It compares whether the query and reference sustain excursions for similar durations, while discarding their absolute positions. Run lengths of 32 samples or more share the last histogram bin.
\begin{promptlisting}{T04 -- state-persistence scoring core}
# Fixed parameters: threshold_quantile=0.5, similarity="js"
def run_profile(signal):
    states = signal > median(signal)
    runs = run_length_encode(states)  # (state, length) for each contiguous run
    histograms = []
    for state in [False, True]:
        lengths = [clip(length, 1, 32) for s, length in runs if s == state]
        histograms.append(count_lengths(lengths, bins=1..32))
    return normalize(concatenate(histograms))

def similarity(query, reference):
    p = run_profile(query)
    q = run_profile(reference)
    distance = jensen_shannon_distance(normalize(p), normalize(q), base=2)
    return clip(1 - distance, 0, 1) if is_finite(distance) else 0
\end{promptlisting}

\subsubsection{Tool 2: Envelope-Spectrum Similarity (T07)}
\label{app:tool-example-envelope-spectrum}
\nolinkurl{quant_envelope_spectrum_log_cosine}

This tool extracts the amplitude envelope using the Hilbert transform, then compares the shape of its log-scaled power spectrum. It measures modulation and burst-repetition structure with reduced dependence on waveform phase. Frequencies are represented by normalized FFT bins; the tool does not identify physical defect frequencies in Hz.
\begin{promptlisting}{T07 -- amplitude-modulation scoring core}
# Fixed parameters: domain="envelope", kind="periodogram",
#                   transform="log", similarity="cosine"
def envelope_spectrum(signal):
    centered = signal - mean(signal)
    envelope = abs(hilbert(centered))
    envelope = envelope - mean(envelope)
    power = abs(rfft(envelope * hann_window(len(signal)))) ** 2
    power = power[1:]  # omit the DC bin
    positive = power[power > 0]
    scale = median(positive) if len(positive) > 0 else 1
    return normalize(log1p(power / max(scale, 1e-12)))

def similarity(query, reference):
    p = envelope_spectrum(query)
    q = envelope_spectrum(reference)
    denominator = norm(p) * norm(q)
    if denominator <= 1e-12:
        return 0.5
    return clip((dot(p, q) / denominator + 1) / 2, 0, 1)
\end{promptlisting}

\FloatBarrier
\Needspace{6\baselineskip}

\raggedbottom

\section{Prompt Templates for Each Agent Role}
\label{app:agents-prompts}

The templates below include active role contracts and, where explicitly marked, an inactive historical template. Runtime-injected figures, class labels, and file paths are represented by angle-bracket placeholders; the \host{} interface, \newred{acceptance} rules, and required output fields are part of each contract.

\subsection{Evidence-Preparation Agents}
\label{app:agents-evidence-prep}

\begin{promptlisting}{\agent{SAgent} (runtime template)}
SYSTEM
You are reviewing RAW multi-channel time-series visualizations for the input sample and its references. Choose the best smoothing method PER CHANNEL for downstream fault inference, then assess confidence in that choice.

INPUT
Available channels: <all_channels>
Sample length: <sample_length>

CONSTRAINTS
Every channel must receive one enabled registered family. Decide per channel from the raw visualization; preserve discriminative transients and do not return parameters or tool names. The host tunes a reviewed parameter grid.

JSON OUTPUT
{
  "reasoning": "<per-channel rationale>",
  "confidence": "<level>",
  "channels": {
    "<channel>": {
      "enabled": true,
      "family": "<family>"
    }
  }
}
\end{promptlisting}

\begin{promptlisting}{\agent{CSAgent} (runtime template)}
SYSTEM
You are reviewing a smoothed multi-channel time-series visualization. Select channels informative for fault discrimination and assign each a high, medium, or low importance level.

CONSTRAINTS
Remove a channel only when it is uniform cyclic/AC, has no modulation or deviation, and adds no reference-comparison context. If uncertain, keep it at medium. High priority requires a visible primary discriminator.

JSON OUTPUT
{
  "selected_channels": ["<channel>"],
  "removed_channels": ["<channel>"],
  "channel_weights": {
    "<channel>": "high"
  },
  "reasoning": "<visible cue>"
}
\end{promptlisting}

\noindent\textit{Inactive historical template.} The following Area-segmentation prompt is retained for provenance only. It is not called on the reported inference path; the \host{} fixes stable and deviant regions deterministically.
\begin{promptlisting}{Area-segmentation agent (inactive historical template)}
SYSTEM
You are reviewing one smoothed signal sample and its highlighted channel views.

INPUT
The full signal length is <sample_length>. Orange dashed markers indicate 25

TASK
Return one representative normal region and zero or more abnormal regions. The normal region is a stable, representative baseline; abnormal regions contain clear deviations, bursts, or rhythmic anomalies rather than isolated noise.

CONSTRAINTS
Express every boundary as a fraction of the total signal length. Evaluate the proposal from scratch when a verification image is supplied. Do not call a prominent fault burst normal merely because it is long, and return an empty abnormal list when no clear deviation is visible.

JSON OUTPUT
{
  "normal": {
    "start_pct": 0.05,
    "end_pct": 0.55
  },
  "abnormal": [
    {
      "start_pct": 0.60,
      "end_pct": 0.90
    }
  ],
  "reasoning": "visible cue and marker-based boundary rationale"
}
\end{promptlisting}

\subsection{\agent{Analysts} and \agent{LBAgents}}
\label{app:agents-comparators}

The listings below reproduce the archived runtime role contracts. Their \texttt{normal-region comparator} and \texttt{abnormal-region comparator} correspond to \agent{NAAgent} and \agent{ABAgent}; the lookback comparator templates are grouped as \agent{LBAgents}, reapplying the \agent{NAAgent} and \agent{ABAgent} comparisons to the preceding \newred{sample}, respectively. The archived \texttt{Decision Synthesis Agent} is the \agent{Judge}. These runtime names are retained inside the quoted templates. A \texttt{predicted\_class} field in a \agent{NAAgent} report is actionable only when its \texttt{classification\_status} is \texttt{decided}; an ambiguous or unknown report names candidate classes but no decided class.

\begin{promptlisting}{\agent{NAAgent} (runtime template)}
SYSTEM
You are the normal-region visual comparator for a multi-class time-series diagnosis.

INPUT
Read the supplied input-versus-reference figures first. Compare the stable baseline region only, using the allowed channels and their host-provided priority weights.

CONSTRAINTS
Assign an absolute similarity band (high, medium, or low) to every candidate class. Use only visible structure: baseline texture, quiet-region morphology, and recurrence within the displayed region. Do not infer from class order, textbook expectations, or numeric thresholds. If the top band is shared, return an ambiguous status and no class vote. If no candidate is answerable, return unknown.

JSON OUTPUT
{
  "classification_status": "decided|ambiguous|unknown|failed",
  "predicted_class": 0,
  "ambiguous_classes": [],
  "similarity_bands": {
    "0": "high",
    "1": "low"
  },
  "reasoning": "sample-specific visible comparison",
  "skeptical_questions": [],
  "needed_investigations": []
}
\end{promptlisting}

\begin{promptlisting}{\agent{ABAgent} (runtime template)}
SYSTEM
You are the abnormal-region visual comparator for a multi-class time-series diagnosis.

INPUT
Read the focused deviation figures and every input-versus-reference figure. Compare burst morphology, active fraction, event duration, and temporal recurrence on the host-selected channels.

CONSTRAINTS
Do not assume that the input is faulty: Class 0 remains eligible. Do not use absolute amplitude thresholds, hidden labels, baseline-model output, or a class named by the caller as a reason to choose. Assign a band to every candidate. A shared top band is ambiguous and casts no vote; all-low evidence is unknown.

JSON OUTPUT
{
  "classification_status": "decided|ambiguous|unknown|failed",
  "predicted_class": null,
  "ambiguous_classes": [1, 2],
  "similarity_bands": {
    "0": "low",
    "1": "medium",
    "2": "medium"
  },
  "reasoning": "named observable that separates or ties candidates",
  "skeptical_questions": [],
  "needed_investigations": []
}
\end{promptlisting}

\noindent\textit{Scope of the \agent{Analyst} null.} The \texttt{predicted\_class=null} value above belongs only to an intermediate \agent{Analyst} report: it means that \agent{Analyst} casts no class vote. It is not the public Stage-2 \texttt{predicted\_class}. The admitted proposal, or else the \agent{Judge}'s last-step class in $\mathcal H_1$, supplies $\hat y$, so the serialized Stage-2 result always contains a class.

\begin{promptlisting}{\agent{LBAgents}: \agent{NAAgent} on the preceding \newred{sample} (runtime template)}
SYSTEM
You are reviewing the preceding window as non-voting temporal context.

TASK
Compare the preceding window's stable region with the same resolved class references and the same effective channel priorities used for the current window.

CONSTRAINTS
Preserve explicit ties and unavailable figures. A lookback result cannot override a current-window comparator or host-authorized measurement, and it cannot create a class vote. Report whether it is available, tied, internally inconsistent, or informative.

JSON OUTPUT
{
  "status": "available|tied|unavailable|failed",
  "classification_status": "decided|ambiguous|unknown|failed",
  "predicted_class": null,
  "ranked_classes": [],
  "supports_normal": false,
  "supports_fault_classes": [],
  "reasoning": "temporal-context comparison and limitations"
}
\end{promptlisting}

\begin{promptlisting}{\agent{LBAgents}: \agent{ABAgent} on the preceding \newred{sample} (runtime template)}
SYSTEM
You are reviewing the preceding window's focused deviation region as non-voting temporal context.

TASK
Compare the previous window's event morphology and recurrence with every resolved reference class using the active channel set. State whether the pattern is tied or class-discriminating without turning that result into an authorization.

CONSTRAINTS
Do not use the current prediction as an input. Do not infer fault existence from a lookback alone. Preserve ambiguity, unavailable regions, and any conflict with the current window in structured fields.

JSON OUTPUT
{
  "status": "available|tied|unavailable|failed",
  "classification_status": "decided|ambiguous|unknown|failed",
  "predicted_class": null,
  "ranked_classes": [],
  "supports_normal": false,
  "supports_fault_classes": [],
  "reasoning": "previous-window morphology and explicit non-voting status"
}
\end{promptlisting}

\subsection{\agent{Judge} and \host{} Contract}
\label{app:agents-judge-host}

\begin{promptlisting}{\agent{Judge} (runtime template)}
SYSTEM
You are the Decision Synthesis Agent for multi-channel time-series fault analysis.

MISSION
Decide the final class; the dataset registry designates one class as normal and all other classes as fault types. Do not finalize while material evidence is unresolved. Name the live confusion and the smallest discriminating question.

TOOLS
Return valid JSON: "reasoning" and "tool_calls". Results arrive next turn.
Never combine "finalize" with a non-final call; evidence does not choose a class.

USER
[HOST-SUPPLIED EVIDENCE NARRATIVE]
- class semantics, reference provenance, and selected view/region artifacts
- normal and abnormal comparator states, including ambiguities
- baseline arbitration, lookback context, and authorized quantitative records

WORKSPACE
Root: <workspace_dir>
Use the tool-ready paths in the decision brief; the full audit is in full_judge_context.json.

INSTRUCTIONS
1. Name the candidates, question, and observable answer that separates them.
2. Request only the focused comparison, image inspection, or registered action needed to answer that question; reconcile disagreements before finalizing.
3. A blocked, unavailable, or unpreflighted result cannot support a final class.

EXAMPLE RESPONSE SHAPE (investigation turn)
{
  "reasoning": "<named question and observable>",
  "tool_calls": [{
    "action": "compare_abnormal",
    "task_objective": "<question>"
  }]
}
HOST INTERFACE
The host validates JSON, dispatches action, and admits "finalize" only after authorization and budget checks, followed by the host finalization check.
\end{promptlisting}

\noindent\textit{Runtime meaning of ``do not finalize.''} Here \texttt{finalize} is a \agent{Judge} request rather than something the \host{} publishes. The same instruction is present in the ungoverned prompt; the governed configuration additionally requires the \host{} to admit a \texttt{finalize} request only after the finalization check. If no admissible call arrives before the step limit, the \host{} publishes the \agent{Judge}'s last-step class, which must lie in $\mathcal H_1$, and marks it as not admitted.

\begin{promptlisting}{\newred{Registered-measurement and \host{}-finalization contract}}
JUDGE REQUEST
Declare the unresolved signal properties to measure, chosen from the registered property list. Then commit to a batch of ToolCards from the host's offer.

HOST MEASUREMENT
The host offers only ToolCards that measure a declared property. Before running a committed batch, it checks that the batch uses only offered cards, covers every declared property, and declares no property measured in an earlier round. A returned packet reports the per-channel rankings, the winning class, and the tool's stated limitations; a packet whose tool failed, or that has no winning class, is excluded and supports no class.

FINALIZATION
The host admits the proposed class only if it is one of the starting candidate classes and at least one measurement round is complete. It does not validate free-text causal claims. It then serializes the reported class, its record, confidence, reasoning, and the recoverable evidence/action trace.
\end{promptlisting}

\section{\texorpdfstring{End-to-End Trace of One \newred{Sample}, Stage by Stage}{End-to-End Trace of One Sample, Stage by Stage}}
\label{app:agent-workflow-trace}

This section reproduces a five-part workflow trace from the frozen artifact. All five parts follow the same query \newred{sample} from one archived run. The trace parts are condensed for readability; agent statements, identifiers, tool descriptions, scores, and stated limitations are kept as recorded. Appendix~\ref{app:workflow-packet-influence-gate} shows one corroborative packet and one advisory packet for Class~2; a third packet is excluded. Appendix~\ref{app:workflow-finalization} shows the \host{} publishing the \agent{Judge}'s admitted Class~2 together with the record behind it. The trace illustrates the protocol; it is not an additional aggregate evaluation result.

\begin{table}[H]
\caption{Roadmap for the complete workflow trace.}
\label{tab:agent-workflow-roadmap}
\centering\small
\setlength{\tabcolsep}{5pt}
\renewcommand{\arraystretch}{1.12}
\begin{tabular}{@{}cp{0.84\linewidth}@{}}
\toprule
Appendix & Protocol event shown \\
\midrule
\ref{app:workflow-evidence-construction} & Agent selection of transform families and channel priorities, followed by \host{} parameter tuning, deterministic segmentation, and channel guards. \\
\ref{app:workflow-comparative-reasoning} & Frozen \agent{NAAgent} and \agent{ABAgent} comparisons that reject the healthy class but leave the two fault classes live. \\
\ref{app:workflow-request-gate} & Uncertainty declaration, score-blind ToolCard offer, pre-result batch commitment, and \host{}-controlled execution. \\
\ref{app:workflow-packet-influence-gate} & Packet provenance checks, reference preflight, influence assignment, and the round report. \\
\ref{app:workflow-finalization} & \agent{Judge} proposal, \host{} recheck of the earlier records, and the \newred{admitted} Class~2 record. \\
\bottomrule
\end{tabular}
\end{table}

The sequence makes the authority boundary observable across time. The agent may shape a view, compare references, state what remains uncertain, select from a sealed offer, and propose a diagnosis. The \host{} alone fixes transform parameters and regions, controls when measurements execute, assigns what their packets may support, and decides whether the proposed class is admitted before writing the terminal record.

\begingroup
\setlength{\parindent}{0pt}
\FloatBarrier
\Needspace{6\baselineskip}
\clearpage
\refstepcounter{subsection}\label{app:workflow-evidence-construction}

{\large\bfseries \tracesec{1} --- Evidence Construction: building a view both sides must live with}\\[4pt]
{\small Taken from one archived run (trace steps \textnormal{\ttfamily 1--2}), figures included. Before anything is compared or diagnosed, the system has to decide \emph{what the signal looks like}. That choice is normally made by whoever is also drawing the conclusion. Here it is split in two, and frozen before any comparison begins.}

\vspace{5pt}
\seriesstrip{\here{\tracesec{1} Evidence Construction}\ $\rightarrow$ \ \steptwo\ $\rightarrow$ \ \stepthree\ $\rightarrow$ \ \stepfour\ $\rightarrow$ \ \stepfive}

\vspace{5pt}
\begin{tcolorbox}[colback=black!2,colframe=black!30,boxrule=0.4pt,arc=2pt,
  left=7pt,right=7pt,top=6pt,bottom=6pt,breakable]

\note{The \newred{sample} as it arrives.}\ Three channels, 2048 \newred{observations}: one vibration channel and two phase currents. The currents are clean periodic sinusoids; the vibration channel is noisy and non-periodic, and whatever fault evidence exists is in its texture rather than its level.

\vspace{4pt}
\begin{center}\includegraphics[width=0.93\linewidth]{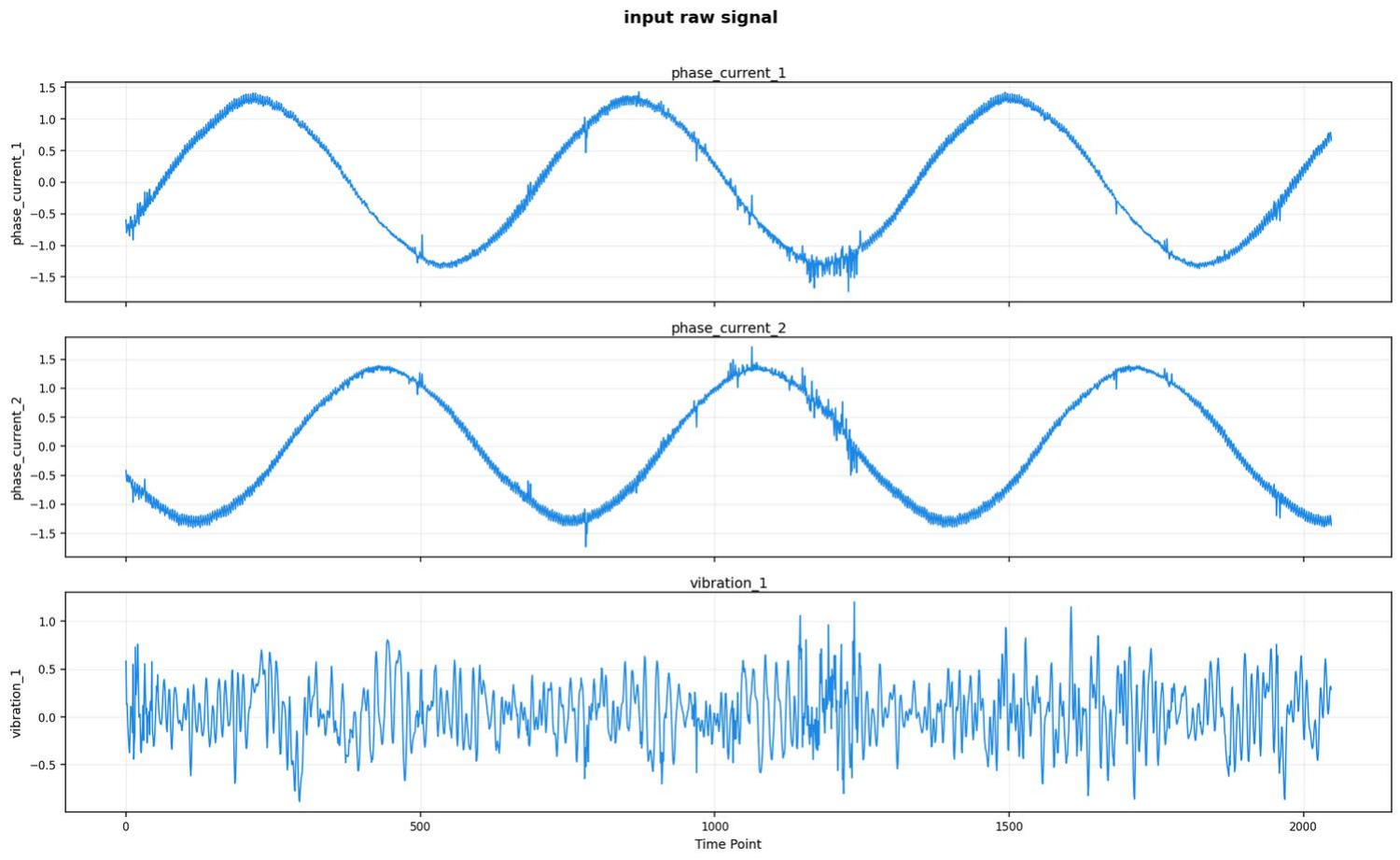}\end{center}

\hsep

\jturn{Step 1}{--- the agent chooses a transform \emph{family}, and only that}\\[4pt]
The smoothing agent is shown the raw signal and a registry of transform families, each with a purpose, its strengths, and the cases where it should be avoided. It returns one family per channel and no parameters at all:\\[3pt]
\hspace*{1.2em}\begin{tabular}{@{}>{\ttfamily}l@{\quad}l@{}}
phase\_current\_1 & \textnormal{\ttfamily savgol} \lab{Savitzky--Golay}\\
phase\_current\_2 & \textnormal{\ttfamily savgol}\\
vibration\_1 & \textnormal{\ttfamily median\_mean\_ewm} \lab{rolling median + EWM}\\
\end{tabular}\\[4pt]
\note{Its reasoning, in its own words.}\ \emph{``For \textnormal{\ttfamily phase\_current\_1} and \textnormal{\ttfamily phase\_current\_2} the signals are clearly periodic sinusoids with high-frequency noise; Savitzky--Golay smooths the noise while preserving the underlying shape and avoiding the phase lag of causal filters. For \textnormal{\ttfamily vibration\_1} the signal is non-stationary with significant impulsive content and bursts. A simple low-pass filter might blur these critical transient spikes; \textnormal{\ttfamily median\_mean\_ewm} combines robustness against isolated impulses with trend tracking, extracting the envelope without destroying the diagnostic spike structure.''}\\[4pt]
\vspace{4pt}
\begin{center}\includegraphics[width=0.97\linewidth]{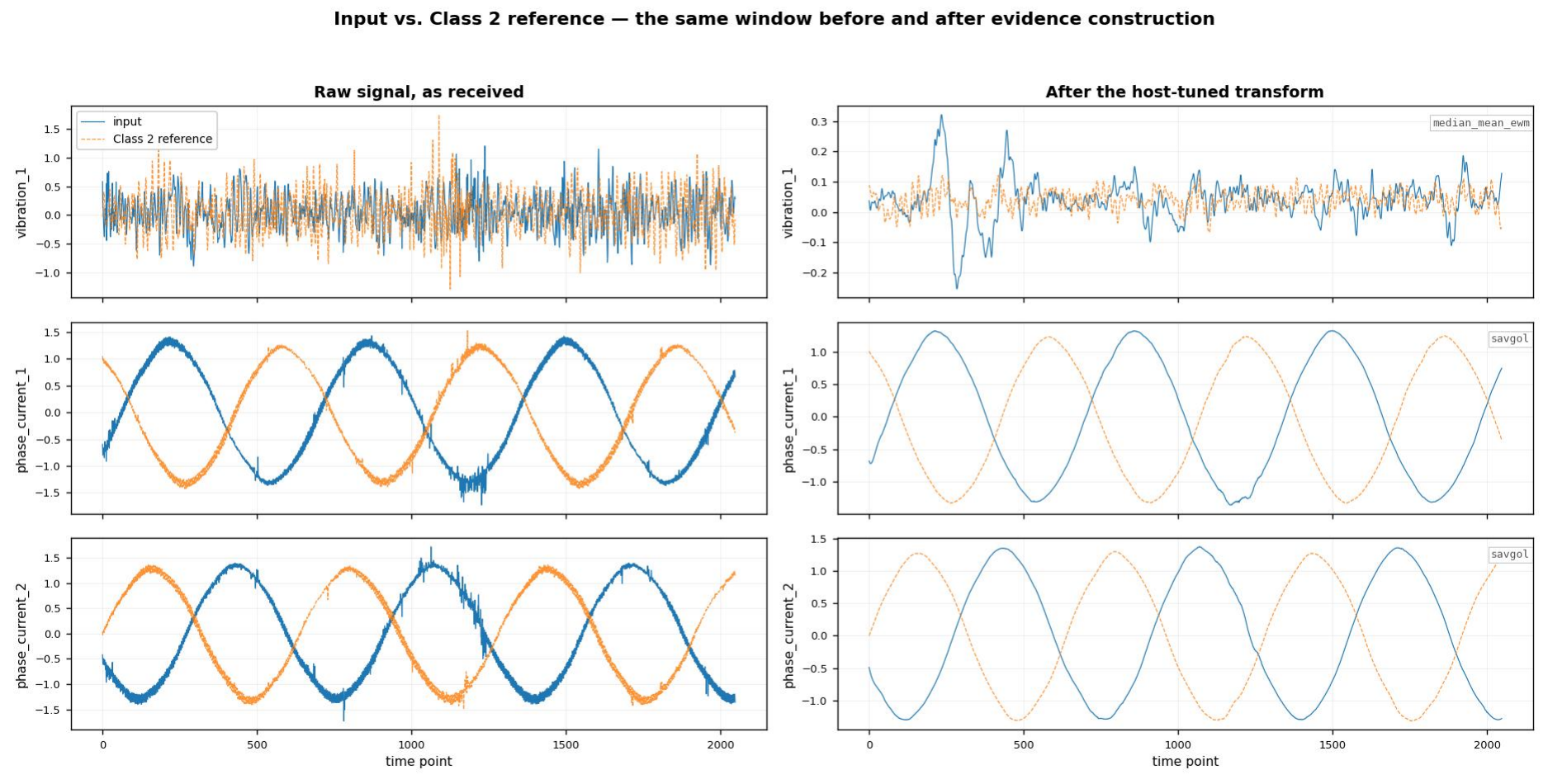}\end{center}
\vspace{1pt}
\note{What the transform actually did.}\ The left column is the \newred{sample} as it arrived, overlaid on the Class~2 reference it will later be compared against; the right column is the same pair after the \host{}-tuned transform. Two things are worth reading off it.\\[3pt]
\hspace*{1.2em}\begin{tabular}{@{}p{0.035\linewidth}p{0.90\linewidth}@{}}
\textbf{1.} & On \textnormal{\ttfamily vibration\_1} the raw traces are \emph{indistinguishable} --- two clouds of dense noise occupying the same $\pm1.5$ band. After \textnormal{\ttfamily median\_mean\_ewm} they separate cleanly: the reference settles around $+0.05$ and the input around $-0.02$, on an axis two orders of magnitude finer. The difference the \newred{Analysts} will argue over does not exist in the raw view. It is produced here.\\[3pt]
\textbf{2.} & On the two phase currents the transform removes the noise and leaves a visible phase offset between input and reference --- which is exactly what the Abnormal Analyst reports in Appendix~\tracesec{2} as a \emph{``sinusoidal match with amplitude/phase offset''}.\\
\end{tabular}\\[4pt]
\note{Why the split matters.}\ Choosing how hard to smooth is exactly the knob that can make a fault appear or disappear. The registry records the policy in one line --- \textnormal{\ttfamily selection\_policy: llm\_selects\_family\_host\_tunes\_parameters}, \textnormal{\ttfamily label\_blind: true} --- so the agent may say \emph{what kind} of view it wants, while the actual parameters come from a finite \host{}-run search under a fidelity floor, with no class labels in scope. The agent never proposes a window length, and the \host{} never accepts one.

\hsep

\hturn{Step 1b}{--- the \host{} segments the result, deterministically}\\[4pt]
With the views fixed, an all-channel segmentor marks stable and deviant regions on every signal --- input and references alike --- before any agent is asked which channels matter. The regions are computed, not proposed.

\vspace{4pt}
\begin{center}\includegraphics[width=0.93\linewidth]{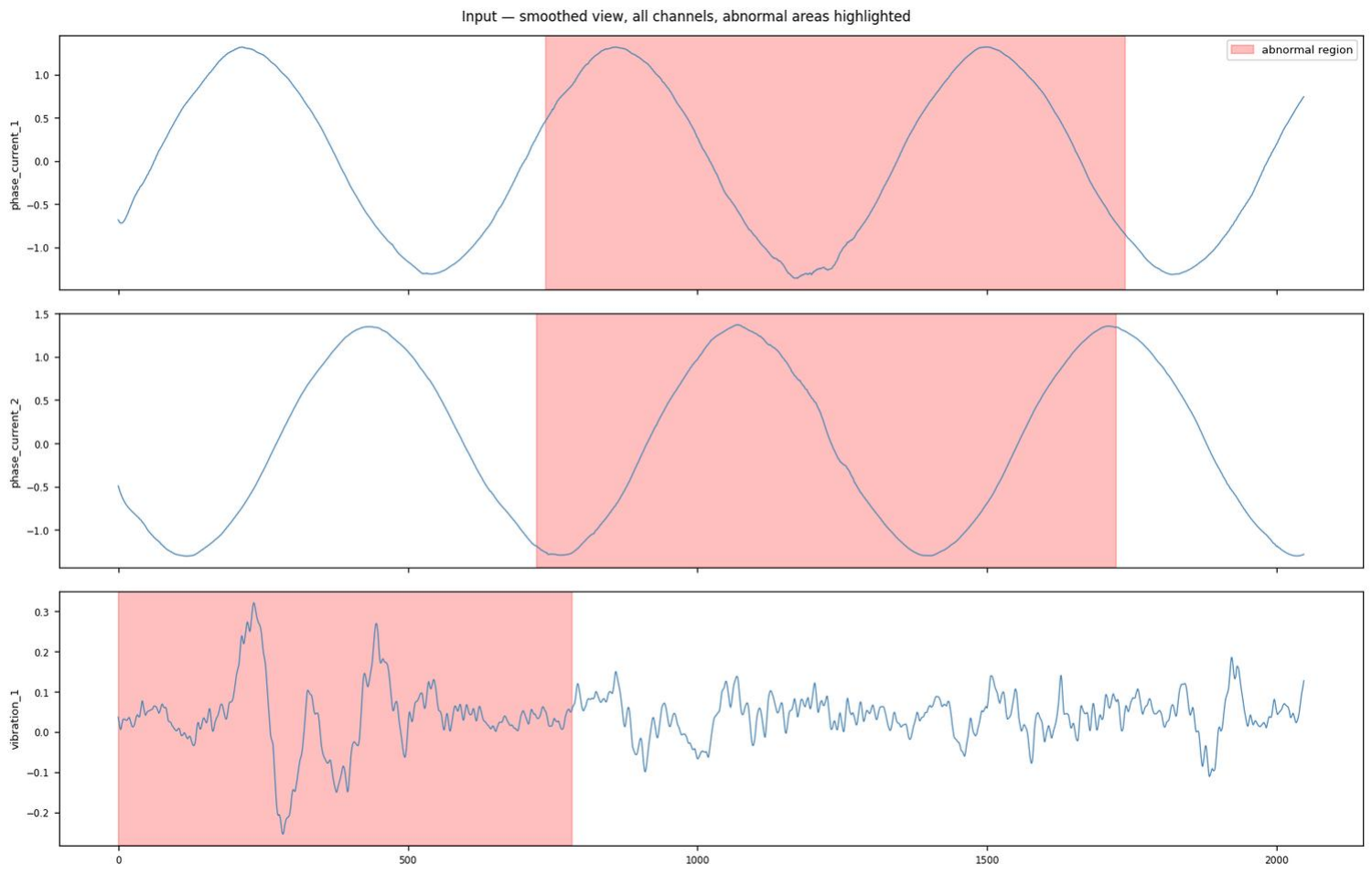}\end{center}

\vspace{2pt}
\note{What it found here.}\ Of 2048 \newred{observations}, 660 fall in stable regions and 1312 in deviant ones (\textnormal{\ttfamily normal\_ratio 0.32}, \textnormal{\ttfamily abnormal\_ratio 0.64}). The consolidated deviant span runs \textnormal{\ttfamily 76--1388} with a core at \textnormal{\ttfamily 222--1248}; on \textnormal{\ttfamily vibration\_1} the marked span is \textnormal{\ttfamily 0--784}, while both current channels are marked far later, around \textnormal{\ttfamily 723--1738}.\\[4pt]
\note{And the \host{} qualifies its own finding, in both directions.}\ The same record carries a separate recurrence assessment --- \textnormal{\ttfamily recurrence\_classification: recurring\_fault\_like}, \textnormal{\ttfamily recurrence\_strength: strong}, \textnormal{\ttfamily recurrence\_score: 0.4161} --- with guidance that cuts the other way from the usual caution: \emph{``recurring impulse clusters were detected; treat sparse abnormal coverage as diagnostically important, do not dismiss it as a healthy isolated transient merely because the marked abnormal region is small.''} The segmentor states how much weight its own marking deserves rather than leaving that to whoever reads it.

\hsep

\jturn{Step 2}{--- the agent ranks the channels, under \host{} guards}\\[4pt]
Only now is an agent asked which channels carry diagnostic weight. Here it keeps two and \emph{drops} one:\\[3pt]
\hspace*{1.2em}\begin{tabular}{@{}>{\ttfamily}l@{\quad}l@{\quad}p{0.58\linewidth}@{}}
vibration\_1 & \normalfont\textbf{high} & \normalfont\lab{``distinct transient events and large amplitude bursts specifically within the highlighted abnormal region''}\\
phase\_current\_1 & \normalfont medium & \normalfont\lab{kept as operating context}\\
phase\_current\_2 & \normalfont\emph{removed} & \normalfont\lab{listed under \textnormal{\ttfamily removed\_channels}}\\
\end{tabular}\\[4pt]
\note{The guard behind this.}\ Priorities are agent-assigned, but the \host{} scores each channel's diagnostic content from the segmentation mask and may \emph{restore} a channel the agent dropped or \emph{raise} one it under-weighted. That matters here: \textnormal{\ttfamily phase\_current\_2} carries the strongest deviant-region score of any channel (prominence $z=15.6$), and it is in use downstream at \textnormal{\ttfamily high} weight --- all three channels appear in the \newred{Analyst comparisons} and in every follow-up measurement. An agent may shape emphasis; it cannot quietly discard a channel that carries the signal.

\hsep

\note{What is frozen at the end of this stage.}\ The smoothed views, the segment masks, and the channel priorities. None of them may be revisited once comparison starts --- which is what makes the next stage's reports usable as evidence rather than as opinions formed alongside a conclusion.

\end{tcolorbox}

\vspace{4pt}
\begin{tcolorbox}[colback=wColor!5,colframe=wColor!45,boxrule=0.4pt,arc=2pt,
  left=7pt,right=7pt,top=6pt,bottom=6pt]
\textbf{What this stage establishes.} A view of the signal that neither party built alone: the agent names the transform family and the channel emphasis, the \host{} fixes every parameter, computes every region, and can override a channel judgement it can score. All of it is fixed before the first comparison.\\[5pt]
\textbf{And what it does not.} The paper reports this boundary as \emph{partial} rather than enforced, and the honest reason is visible above. The family choice is still the agent's, and a family can suppress a texture just as a parameter can. The fidelity floor bounds how much a transform may destroy; it does not prove the retained view is the diagnostic one. \ \lab{Continue to Appendix \tracesec{2}, where these frozen views are compared against the class references.}
\end{tcolorbox}

\FloatBarrier
\Needspace{6\baselineskip}
\FloatBarrier
\Needspace{18\baselineskip}
\refstepcounter{subsection}\label{app:workflow-comparative-reasoning}

{\large\bfseries \tracesec{2} --- Comparative Reasoning: two \newred{Analysts}, and the tie that starts everything}\\[4pt]
{\small Same \newred{sample} as Appendix~\tracesec{1} (trace steps \texttt{3--9}), figures included. The frozen views from Evidence Construction are now compared against the class references. Two \newred{Analysts} run in separate roles, each reports a band per candidate, and both reports are sealed before any model baseline, recurrence evidence or follow-up measurement is disclosed.}

\vspace{5pt}
\seriesstrip{\stepon\ $\rightarrow$ \ \here{\tracesec{2} Comparative Reasoning}\ $\rightarrow$ \ \stepthree\ $\rightarrow$ \ \stepfour\ $\rightarrow$ \ \stepfive}

\vspace{5pt}
\begin{tcolorbox}[colback=black!2,colframe=black!30,boxrule=0.4pt,arc=2pt,
  left=7pt,right=7pt,top=6pt,bottom=6pt,breakable]

\note{Two roles, deliberately separated.}\ The \emph{Normal Analyst} compares representative stable regions \newred{and gives every class a band; on its own it can rule out the healthy class but cannot settle a fault type}. The \emph{Abnormal Analyst} compares the focused deviant region against every candidate reference in turn. Neither sees the other's report, and neither sees the fitted-model \newred{predictions} or the recurrence evidence, which the \host{} discloses only afterwards.

\hsep

\jturn{The Normal Analyst}{--- is this \newred{sample} consistent with a healthy baseline?}\\[4pt]
\note{It does not choose where to look.}\ The stable \newred{segment} is selected deterministically by the \host{} --- \texttt{selection\_authority: deterministic}, \texttt{primary\_mode: best\_window} --- which picks \newred{observations} \texttt{502--1014} of the input, and a separate \newred{segment} in each reference: Class~0 \texttt{837--1349}, Class~1 \texttt{395--907}, Class~2 \texttt{1313--1825}. All three references have a qualifying quiet stretch here, so all three are compared.

\vspace{4pt}
\begin{center}\includegraphics[width=0.985\linewidth]{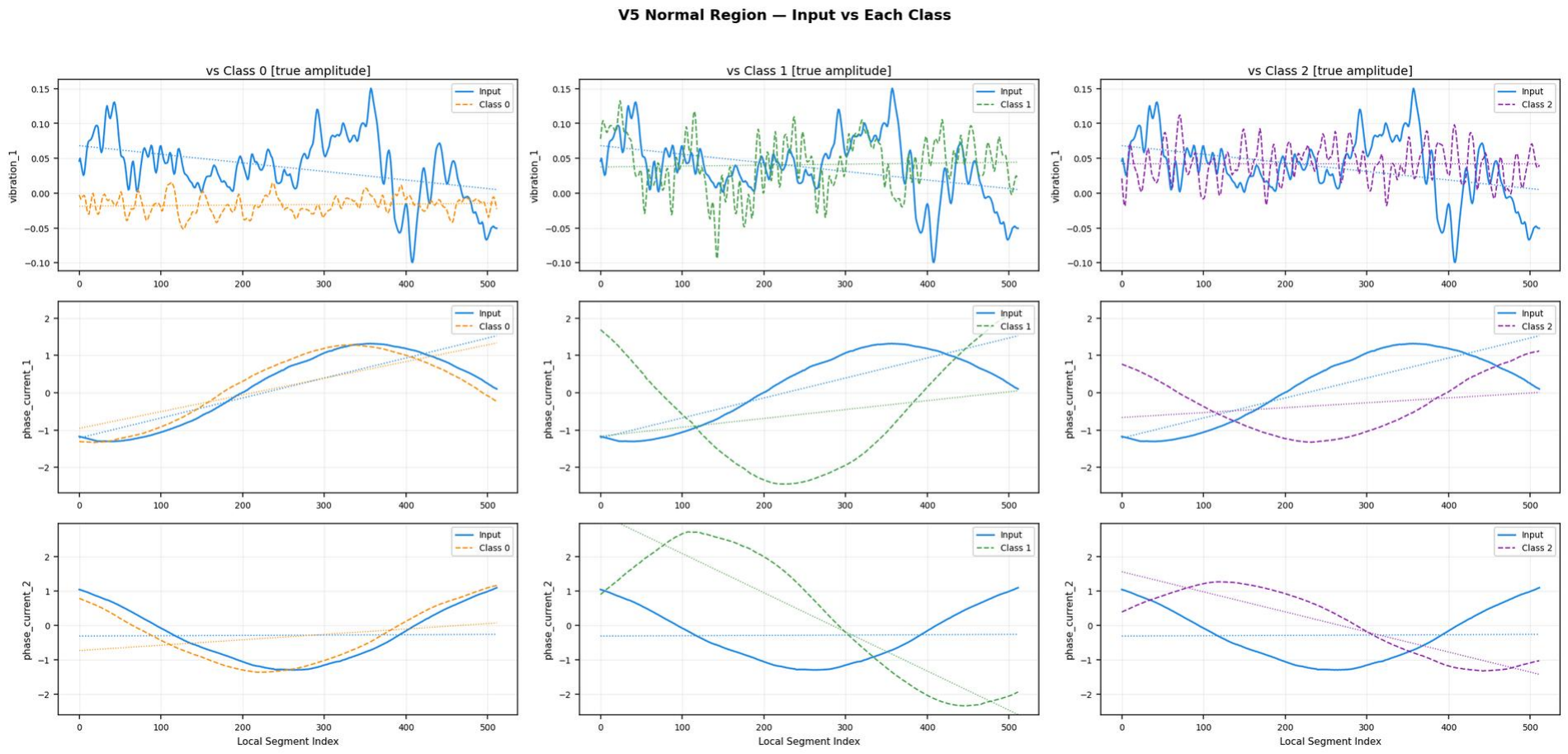}\end{center}
\vspace{2pt}
\note{What it finds.}\ Class~0 is banded \texttt{low} --- on \texttt{vibration\_1} the input sits well above the Class~0 reference, which stays flat near zero. Class~1 and Class~2 are both banded \texttt{high}, and they are high for the same reason: on the vibration channel all three traces share a comparable noisy envelope, and the current channels differ only in phase and offset.\\[4pt]
\note{Its verdict.}\ Equal highest bands stay tied, so the \newred{Analyst} reports the tie rather than picking between them:\\[3pt]
\hspace*{1.2em}\texttt{classification\_status: ambiguous}\quad
\texttt{predicted\_class: null}\quad \texttt{top\_tied\_classes: [1, 2]}\quad\texttt{confidence: medium}\\
\hspace*{1.2em}\lab{available\_class\_labels: [0, 1, 2] $\cdot$ evidence\_role: normal $\cdot$ source\_view: smoothed}\\[4pt]
\note{What it has settled, and what it has not.}\ Class~0 --- the healthy class --- is the one candidate it can rule out from the quiet region. It cannot separate the two fault classes, and says so. That single result is what makes the rest of the case a fault-\emph{type} question rather than a fault-\emph{existence} one.

\hsep

\jturn{The Abnormal Analyst}{--- and the region it is given to look at}\\[4pt]
\note{It does not choose its region either.}\ The deviant span comes from the same deterministic segmentor as the stable \newred{segment}, and it is scored rather than merely marked. On the consolidated all-channel view the input's deviant span runs \texttt{76--1388}, with a core at \texttt{222--1248} peaking at sample \texttt{1225}. Per channel the segmentor reports separately:\\[4pt]
{\footnotesize
\setlength{\tabcolsep}{5pt}\renewcommand{\arraystretch}{1.15}
\hspace*{1.2em}\begin{tabular}{@{}>{\ttfamily}lccccc@{}}
\toprule
\normalfont\textbf{Scope} & \textbf{\normalfont deviant span} & \textbf{\normalfont core} & \textbf{\normalfont peak} & \textbf{\normalfont prominence $z$} & \textbf{\normalfont diagnostic?}\\
\midrule
\normalfont all channels & \ \ 76--1388 & 222--1248 & 1225 & 10.21 & \normalfont yes\\
vibration\_1      & \ \ \ \ 0--784 & \ 220--303 & \ 284 & 10.94 & \normalfont yes\\
phase\_current\_1 & 738--1738 & 1147--1255 & 1238 & \ 7.46 & \normalfont yes\\
phase\_current\_2 & 723--1723 & 1180--1262 & 1223 & \boldnum{15.57} & \normalfont yes\\
\bottomrule
\end{tabular}}\\[4pt]
\note{Two things follow from that table.}\ First, the channels disagree sharply about \emph{when} the \newred{sample} is unusual: \texttt{vibration\_1} marks the opening 784 \newred{observations} while both current channels mark a span starting past 720, so the consolidated region spans nearly the whole \newred{sample} rather than being the intersection of confident channels. Second, the strongest deviant-region score in the whole record --- prominence $z=15.6$ --- belongs to \texttt{phase\_current\_2}, the channel the agent had proposed to remove in Appendix~\tracesec{1}. The segmentor scores channels independently of who wanted them kept.

\vspace{5pt}
\jturn{The Abnormal Analyst}{--- comparing the deviant region against all three references}\\[4pt]
\begin{center}\includegraphics[width=0.985\linewidth]{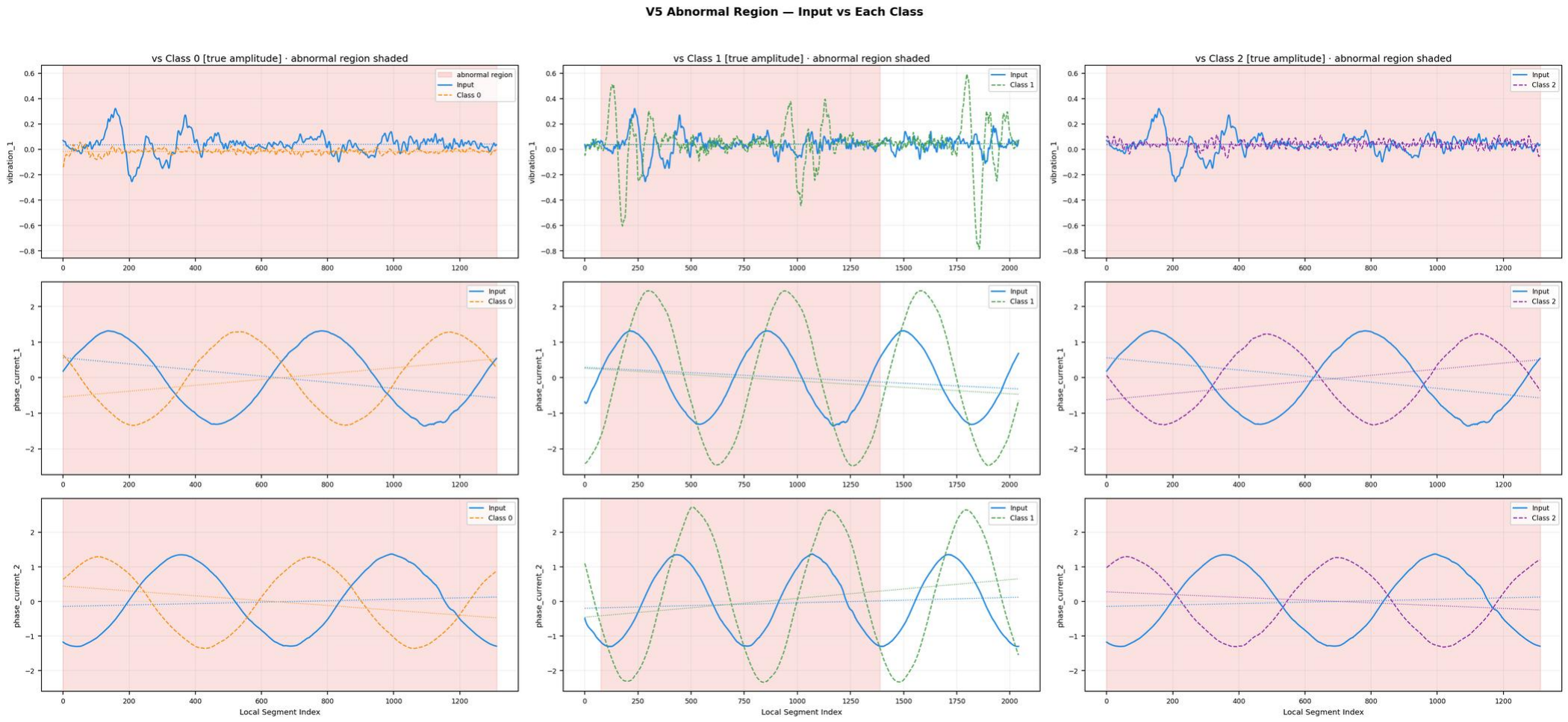}\end{center}
\vspace{2pt}
\note{Three candidates, three different bands.}\ Unlike the normal region, this comparison separates all three:\\[3pt]
\hspace*{1.2em}\begin{tabular}{@{}l@{\quad}l@{\quad}p{0.56\linewidth}@{}}
\textbf{Class 2} & \texttt{high} & \lab{vibration\_1 high $\cdot$ phase\_current\_1 medium $\cdot$ phase\_current\_2 high}\\
\textbf{Class 1} & \texttt{medium} & \lab{vibration\_1 low $\cdot$ phase\_current\_1 high $\cdot$ phase\_current\_2 high}\\
\textbf{Class 0} & \texttt{low} & \lab{vibration\_1 low $\cdot$ phase\_current\_1 medium $\cdot$ phase\_current\_2 medium}\\
\end{tabular}\\[4pt]
\note{And yet it still refuses to conclude.}\ Class~2 is the sole \texttt{high} band, with no tie to report (\texttt{top\_tied\_classes: []}). \newred{An Analyst} optimising for a confident answer would stop here. This one returns \texttt{classification\_status: ambiguous} with \texttt{predicted\_class: null} over \texttt{ambiguous\_classes: [2, 0, 1]}, and files the reason as a question it cannot answer itself: \emph{``does the downstream judge confirm that the \texttt{vibration\_1} magnitude match with Class~2 is diagnostic rather than coincidental?''} and \emph{``are there recurrence or envelope-spectrum features that further distinguish Class~2 from the other fault types?''}\\[4pt]
\note{Why that matters.}\ Leading on one band is not the same as being established. The \newred{Analyst} has a favourite and declines to promote it to a verdict, because the evidence separating Class~2 from Class~1 is a magnitude match on a single channel --- exactly the kind of thing a quantitative measurement can check and a picture cannot.

\hsep

\jturn{A closer look}{--- the focused view behind the leading band}\\[4pt]
For each candidate the analyst is also given a focused rendering of the deviant region at true amplitude. The Class~2 view is shown here --- the comparison that produced the only \texttt{high} band, and the one the analyst then declined to treat as decisive.

\vspace{3pt}
\begin{center}\includegraphics[width=0.86\linewidth]{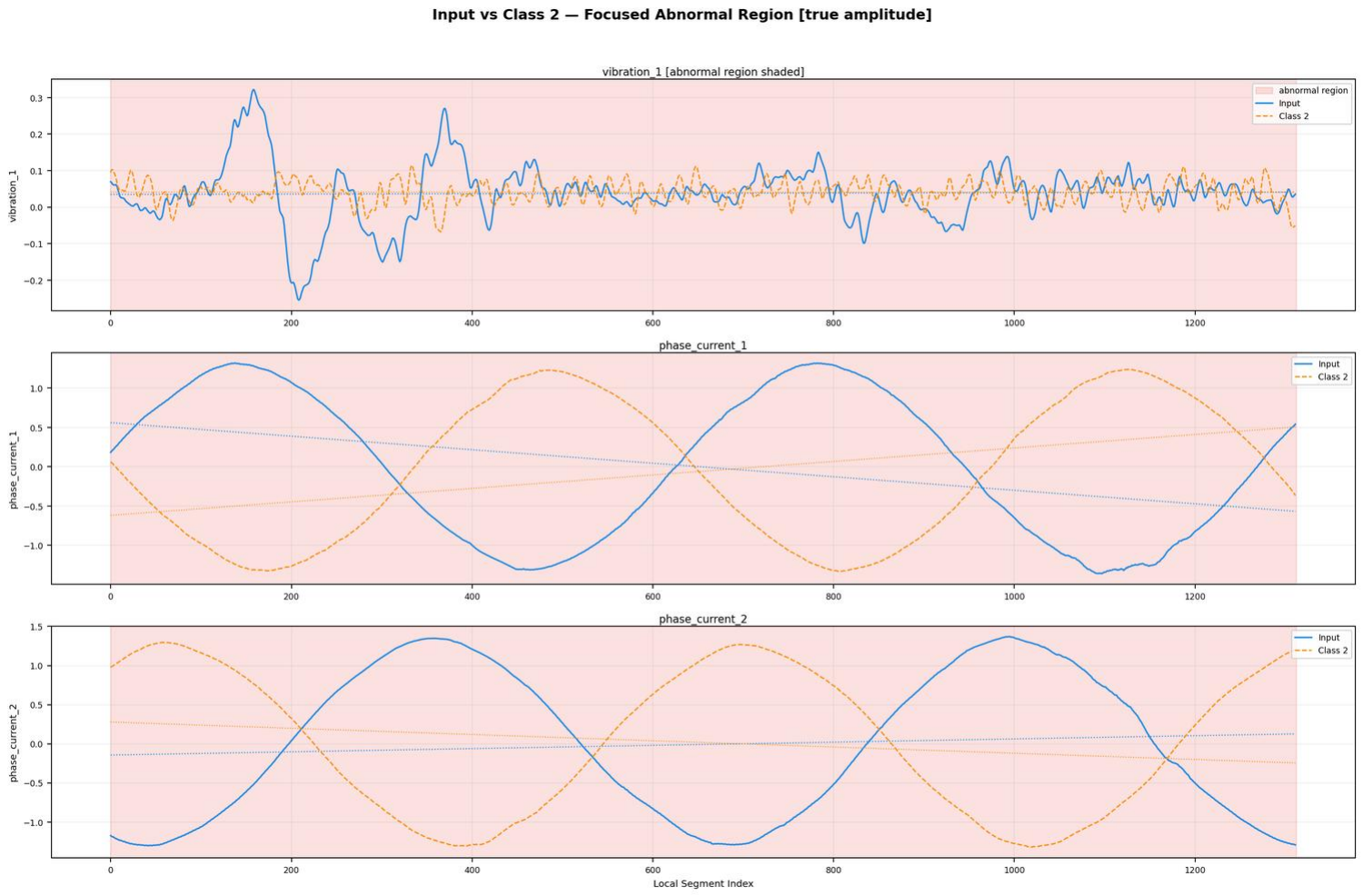}\end{center}
\vspace{1pt}
\lab{The match is an envelope and magnitude match on the vibration channel. That is real evidence, and it is also exactly what a load change could imitate --- which is what the analyst's own skeptical question asks about.}

\hsep

\hturn{The \host{}}{--- carries the uncertainty forward instead of resolving it}\\[4pt]
Two \newred{Analysts} have now run on the same \newred{sample}, over different regions, and neither has produced a class. The normal region ruled out Class~0 and tied Classes 1 and 2. The deviant region ranked Class~2 first but refused to certify that its lead is diagnostic. Neither result is discarded and neither is promoted: what survives is the intersection of what they could establish.\\[4pt]
\hspace*{1.2em}\texttt{normal:}\ \ \texttt{ambiguous}, \texttt{top\_tied\_classes: [1, 2]}, \texttt{confidence: medium}\\
\hspace*{1.2em}\texttt{abnormal:}\ \texttt{ambiguous}, \texttt{predicted\_class: null}, \texttt{confidence: low}\\[4pt]
\note{What that leaves.}\ Fault existence is settled --- both \newred{Analysts} put Class~0 at \texttt{low} --- so the remaining question is fault \emph{type}, between Class~1 and Class~2. That is the live candidate set $\mathcal H_t=\{1,2\}$ the Judge inherits, and it is why the request in Appendix~\tracesec{3} is scoped \texttt{type} rather than \texttt{both}.

\hsep

\note{What is handed forward.}\ Two frozen reports, a live candidate set $\mathcal H_t=\{1,2\}$, and a named uncertainty that both \newred{Analysts} independently pointed at the same way --- recurrence and envelope-spectrum structure, which neither of them is permitted to measure. This is the state the Judge inherits at the start of Appendix~\tracesec{3}, and the reason it opens a registered measurement round at all.

\end{tcolorbox}

\vspace{4pt}
\begin{tcolorbox}[colback=wColor!5,colframe=wColor!45,boxrule=0.4pt,arc=2pt,
  left=7pt,right=7pt,top=6pt,bottom=6pt]
\textbf{What this stage establishes.} A comparison that cannot be revised in the light of what comes later. Both reports are sealed before the fitted-model \newred{predictions}, the recurrence evidence and any follow-up measurement are disclosed, so no downstream result can reach back and rewrite an observation. The roles are split so that the question \emph{is this healthy?} is answered by \newred{an Analyst} that cannot \newred{settle} a fault type. And an unresolved comparison is allowed to stay unresolved: the tie is recorded and carried forward rather than broken by whichever candidate looked marginally better.\\[5pt]
\textbf{And what it does not.} A band is a judgement, elicited from a model looking at a picture. Freezing it makes it auditable, not correct --- and the same \newred{sample} can be read differently under a different frozen view, which is the limitation inherited from Appendix~\tracesec{1}. What the protocol adds here is only that the reading is fixed before the evidence that might have been used to justify it. \ \lab{Continue to Appendix \tracesec{3}, where the Judge must ask for a measurement to break this tie.}
\end{tcolorbox}

\FloatBarrier
\Needspace{6\baselineskip}
\FloatBarrier
\Needspace{18\baselineskip}
\refstepcounter{subsection}\label{app:workflow-request-gate}

{\large\bfseries \tracesec{3} --- The Request Gate: asking for a measurement without being able to shop for one}\\[4pt]
{\small This exchange reproduces trace steps \texttt{18--20} from one archived run: the identifiers, the tool descriptions and the sealed hashes are all copied from the log rather than written for the example\newred{; log fields that Section~\ref{sec:request-gate} does not use are omitted}. The exchange shows only how a measurement comes to be taken. What each measurement is then \emph{allowed to count for} is settled later, by the packet and influence gate, and we deliberately leave that out here so the acquisition step can be read on its own.}

\vspace{5pt}
\seriesstrip{\stepon\ $\rightarrow$ \ \steptwo\ $\rightarrow$ \ \here{\tracesec{3} Request Gate}\ $\rightarrow$ \ \stepfour\ $\rightarrow$ \ \stepfive}

\vspace{5pt}
\begin{tcolorbox}[colback=black!2,colframe=black!30,boxrule=0.4pt,arc=2pt,
  left=7pt,right=7pt,top=6pt,bottom=6pt,breakable]

\jturn{step 18}{--- naming what is unclear}\\[4pt]
\note{Where things stand.}\ Both \newred{Analysts} have finished and neither produced a class. Between them they ruled out Class~0 and left the two fault classes live, so $\mathcal H_t=\{1,2\}$ over three channels, \texttt{vibration\_\allowbreak 1}, \texttt{phase\_\allowbreak current\_\allowbreak 1} and \texttt{phase\_\allowbreak current\_\allowbreak 2}. Nothing has been measured yet, and no score of any kind exists anywhere in the system.\\[4pt]
\note{What the Judge asks for.}\ Rather than naming a tool, the Judge names the signal \newred{properties} it believes would settle the question, drawn from the registered vocabulary:\\[3pt]
\hspace*{1.2em}\texttt{measured\_\allowbreak signal\_\allowbreak properties: [extrema\_\allowbreak morphology, roughness\_\allowbreak across\_\allowbreak scales, recurrence\_\allowbreak match]}\\[4pt]
\note{Why the \host{} accepts it.}\ All three \newred{properties} are registered and none has been measured before in this case\newred{, so the declaration meets the no-repetition condition of Eq.~\ref{eq:request-property-coverage}}. What matters is the timing: the Judge has now committed to \emph{what} it wants measured while it still has no idea what is available to measure it.

\hsep

\hturn{step 19}{--- answering with an offer that carries no scores}\\[4pt]
The \host{} seals the request as \texttt{selection\_\allowbreak request\_\allowbreak id: qsel\_\allowbreak 5d8056a5830c1c16bd46} and returns it with \texttt{scores\_\allowbreak exposed: false} and \texttt{predicted\_\allowbreak class\_\allowbreak exposed: false}.

\vspace{5pt}
\note{What comes back.}\ \newred{Five ToolCards: the \host{} offers every default ToolCard that measures a requested property, and \texttt{roughness\_\allowbreak across\_\allowbreak scales} is measured by three of them, which is why three requested properties produce five cards.} Every card arrives with its complete description; here is the first one exactly as the \host{} returns it.

\vspace{5pt}
{\footnotesize\ttfamily
\begin{tabular}{@{\hspace*{1.2em}}l@{\ \ }p{0.76\linewidth}@{}}
title & Local-extrema profile similarity\\
purpose & Compare peak/trough density, prominence, and spacing over time.\\
answers & [extrema\_\allowbreak morphology, oscillation\_\allowbreak and\_\allowbreak transient\_\allowbreak profile]\\
strengths & [direct morphology summary, retains temporal profile]\\
blind\_\allowbreak spots & [noise creates extrema, window alignment matters]\\
avoid\_\allowbreak when & [noise scales are not comparable]\\
requires & \{full\_\allowbreak sample: true, minimum\_\allowbreak rows: 256, identical\_\allowbreak csv\_\allowbreak schema: true, \ldots\}\\
transform\_\allowbreak steps & [\{fixed\_\allowbreak params: \{distance: l1, window\_\allowbreak count: 32\}, support: complete\_\allowbreak sample\}]\\
scoring\_\allowbreak rule & higher similarity ranks first \quad direction: higher\_\allowbreak is\_\allowbreak better\\
group / cost & local\_\allowbreak extrema / medium \quad version: 1.0.0\\
impl\_\allowbreak hash & sha256:869a7cdc2327c81b\ldots\\
\end{tabular}}

\vspace{6pt}
\note{And the same for the other four.}\ Summarised by the \newred{property that brought them in}, what they do, and where the registry says they are weak.

\vspace{4pt}
{\footnotesize
\setlength{\tabcolsep}{4pt}\renewcommand{\arraystretch}{1.15}
\begin{tabular}{@{}p{0.20\linewidth}p{0.355\linewidth}p{0.40\linewidth}@{}}
\toprule
\textbf{Requested \newred{property}} & \textbf{ToolCard} & \textbf{What it does \textmd{/} where it is weak}\\
\midrule
\texttt{extrema\_\allowbreak morphology} & \texttt{quant\_\allowbreak local\_\allowbreak extrema\_\allowbreak profile\_\allowbreak windows\_\allowbreak 32\_\allowbreak l1}\newline \lab{local\_\allowbreak extrema $\cdot$ medium} & Compare peak/trough density, prominence, and spacing over time.\newline \lab{noise creates extrema; window alignment matters}\\
\texttt{roughness\_\allowbreak across\_\allowbreak scales} & \texttt{quant\_\allowbreak fractal\_\allowbreak scaling\_\allowbreak lags\_\allowbreak 128\_\allowbreak pearson}\newline \lab{multilag\_\allowbreak roughness $\cdot$ low} & Compare how signal roughness changes across lag scales.\newline \lab{trend sensitive; finite length and lag support affect results}\\
\texttt{recurrence\_\allowbreak match} & \texttt{quant\_\allowbreak autocorrelation\_\allowbreak lags\_\allowbreak 128\_\allowbreak pearson}\newline \lab{lag\_\allowbreak recurrence $\cdot$ low} & Compare periodic recurrence across lags without requiring cycle-phase alignment.\newline \lab{common operating cycles can match across classes; lag support limits interpretation}\\
\midrule
\multicolumn{3}{@{}p{0.97\linewidth}}{\lab{\newred{Also offered for the same requested property; the \agent{Judge} did not select them.}}}\\
\texttt{roughness\_\allowbreak across\_\allowbreak scales} & \texttt{quant\_\allowbreak gaussian\_\allowbreak scale\_\allowbreak space\_\allowbreak sigma\_\allowbreak 8\_\allowbreak pearson}\newline \lab{gaussian\_\allowbreak scale\_\allowbreak space $\cdot$ low} & Compare morphology after a declared smoothing scale.\newline \lab{smoothing can erase weak faults; point alignment remains relevant}\\
\texttt{roughness\_\allowbreak across\_\allowbreak scales} & \texttt{quant\_\allowbreak multiscale\_\allowbreak complexity\_\allowbreak scales\_\allowbreak 16\_\allowbreak l1}\newline \lab{multiscale\_\allowbreak complexity $\cdot$ medium} & Compare nonlinear complexity under progressive coarse graining.\newline \lab{finite-length sensitive; less directly interpretable}\\
\bottomrule
\end{tabular}}

\vspace{6pt}
\note{What the \host{} will not say.}\ \newred{Alongside the five cards,} the \host{} does not return any score, any ranking of the candidate classes, or any hint of a predicted class.\\[4pt]
\note{The line this draws.}\ Notice how much the card gives away. It describes exactly \emph{how} the measurement would be computed --- the transform and its fixed parameters, the similarity metric, the direction that counts as better, the row and schema requirements, even a hash of the precise implementation --- and it volunteers the tool's own weaknesses. What it never contains is any indication of \emph{what that tool would return on this particular \newred{sample}}. The Judge can therefore reason carefully about which instrument suits the question, and still cannot anticipate the answer.

\hsep

\jturn{step 19}{--- choosing, before any result exists}\\[4pt]
The Judge takes all three recommended cards and leaves the two alternates. It must give the \host{} a reason for each one, and a stated objective for the round as a whole:\\[4pt]
{\footnotesize
\hspace*{1.2em}\texttt{selection\_\allowbreak request\_\allowbreak id: qsel\_\allowbreak 5d8056a5830c1c16bd46}\quad
\texttt{round\_\allowbreak index: 1}\\[3pt]
\hspace*{1.2em}\parbox{0.93\linewidth}{\texttt{round\_\allowbreak objective:} \emph{``Resolve fault type ambiguity between Class 1 and Class 2 using morphology, roughness, and recurrence properties.''}}\\[4pt]
\hspace*{1.2em}\begin{tabular}{@{}>{\ttfamily}p{0.355\linewidth}p{0.56\linewidth}@{}}
\normalfont\textbf{Tool chosen} & \textbf{\normalfont The \texttt{selection\_\allowbreak reason} it had to give}\\[3pt]
quant\_\allowbreak local\_\allowbreak extrema\_\allowbreak profile\_\allowbreak windows\_\allowbreak 32\_\allowbreak l1 & \normalfont\emph{``Measures peak/trough density and spacing to distinguish burst morphology patterns between candidates.''}\\[2pt]
quant\_\allowbreak fractal\_\allowbreak scaling\_\allowbreak lags\_\allowbreak 128\_\allowbreak pearson & \normalfont\emph{``Assesses signal roughness across scales to differentiate structural complexity independent of absolute amplitude.''}\\[2pt]
quant\_\allowbreak autocorrelation\_\allowbreak lags\_\allowbreak 128\_\allowbreak pearson & \normalfont\emph{``Evaluates periodic recurrence and phase-shift tolerant repetition patterns to match fault dynamics.''}\\
\end{tabular}}\\[5pt]
All three are submitted with \texttt{channel\_\allowbreak policy: all\_\allowbreak channels\_\allowbreak independent}.\\[4pt]
\note{What the \host{} checks before it runs anything.}\ \newred{The three request conditions of Eq.~\ref{eq:request-property-coverage}: the chosen batch lies inside the offer, the three tools together cover all three requested properties, and none of these properties was measured in an earlier round.} The round is then recorded as \texttt{precommitted\_\allowbreak before\_\allowbreak any\_\allowbreak result: true} and \texttt{pre\_\allowbreak execution\_\allowbreak batch\_\allowbreak validation: passed}. Had any of these failed, the \host{} would have returned a \newred{rejection} and \emph{no tool would have run at all}.

\hsep

\hturn{step 20}{--- running the tools, the first moment anything is written}\\[4pt]
The three ToolCards execute and three \texttt{pending} records are appended to the evidence ledger:\\[4pt]
\hspace*{1.2em}\begin{tabular}{@{}l@{\ \ }l@{}}
\texttt{qm\_\allowbreak 1648602c58c5b3baad84} & sealed as \texttt{10139940e3f3ef8f\ldots}\\
\texttt{qm\_\allowbreak 90c0d7a919b3fd0ea53a} & sealed as \texttt{1cccfb014dda865d\ldots}\\
\texttt{qm\_\allowbreak 05fff8ed9ce86c3db2f2} & sealed as \texttt{f0ca18056c3f8ecc\ldots}\\
\end{tabular}\\[4pt]
This is the first point in the whole exchange at which a measurement value exists. Each hash seals the tool's identity together with its per-channel output and its stated limitations, so any later change to a stored result can be detected.\\[4pt]
\note{Where that leaves the round.}\ \texttt{extrema\_\allowbreak morphology}, \texttt{roughness\_\allowbreak across\_\allowbreak scales} and \texttt{recurrence\_\allowbreak match} are now marked as measured.

\hsep

\jturn{after the round}{--- what the ledger now forecloses}\\[4pt]
The three \newred{properties} are now marked as measured, and that is binding regardless of what they returned. Had the Judge wanted to continue, it could not have re-requested \texttt{extrema\_\allowbreak morphology}, \texttt{roughness\_\allowbreak across\_\allowbreak scales} or \texttt{recurrence\_\allowbreak match} under different settings; a later round has to name a \newred{property} the ledger does not already hold\newred{, and the round report must cover every packet (Section~\ref{par:evidence-admission})}. A result the Judge dislikes therefore cannot be quietly rerun, and an earlier one cannot be quietly dropped. In this \newred{sample} the Judge did not open a second round: it proceeded to account for these three packets, which is where the packet and influence gate takes over.

\end{tcolorbox}

\vspace{5pt}
\begin{tcolorbox}[colback=wColor!5,colframe=wColor!45,boxrule=0.4pt,arc=2pt,
  left=7pt,right=7pt,top=6pt,bottom=6pt]
\textbf{What this exchange does establish.} The four steps always happen in this order and the order cannot be reversed: the Judge says what is unclear, the \host{} answers with an offer that carries no scores, the Judge commits to a batch, and only then does anything run. Because no measurement value exists until that last step, no value can have influenced either the \newred{properties that were declared} or the tools that were chosen. Two further rules follow the same logic: a \newred{property} that has been measured cannot be measured again, so an unwelcome result cannot be retried under new settings; and \newred{the batch must cover every declared property, so the \agent{Judge} cannot declare several properties and then run only the tools for the ones it prefers}.\\[6pt]
\textbf{And what it does not.} Nothing here speaks to whether a measurement is any good. Cards are \newred{offered} on their declared \newred{properties} alone; the \host{} matches them to the request, it does not rank them, and it has no opinion about which will prove informative. Whether each of these packets ends up \texttt{excluded}, \texttt{advisory} or \texttt{corroborative} is decided afterwards by the packet and influence gate, \newred{on whether the tool ran, whether the top-tier channels agree on a winning class, and the reference preflight} --- none of which exist at any point in the exchange above.
\end{tcolorbox}

\FloatBarrier
\Needspace{6\baselineskip}
\FloatBarrier
\Needspace{18\baselineskip}
\refstepcounter{subsection}\label{app:workflow-packet-influence-gate}

{\large\bfseries \tracesec{4} --- The Packet and Influence Gate: deciding what a measurement is allowed to count for}\\[4pt]
{\small This continues the same case as the request-gate example, picking up at trace step \texttt{20}, the moment the three precommitted tools come back. The request gate has finished its work: the measurements exist. The question now is no longer whether a measurement \emph{may be run}, but what a measurement that already exists is \emph{allowed to count for}. The \host{} settles that in \newred{two steps, with the reference preflight between them}, and the Judge writes none of them.}

\vspace{5pt}
\seriesstrip{\stepon\ $\rightarrow$ \ \steptwo\ $\rightarrow$ \ \stepthree\ $\rightarrow$ \ \here{\tracesec{4} Packet \& Influence Gate}\ $\rightarrow$ \ \stepfive}

\vspace{5pt}
\begin{tcolorbox}[colback=black!2,colframe=black!30,boxrule=0.4pt,arc=2pt,
  left=7pt,right=7pt,top=6pt,bottom=6pt,breakable]

\stg{\newred{Step 1}}{--- did the registered operation really run on the declared inputs?}\\[4pt]
All three tools return, each with a complete ranking over the live candidates $\{1,2\}$ on all three channels, and none reports a failure. Before considering what any of them means, the \host{} establishes three facts per packet: that execution was valid and the ranking complete, that the comparison covered the full sample with equal row support, and that channel arbitration produced a single winner. The first two hold everywhere here. The third does not.

\vspace{5pt}
{\footnotesize
\setlength{\tabcolsep}{4pt}\renewcommand{\arraystretch}{1.2}
\begin{tabular}{@{}>{\ttfamily}p{0.295\linewidth}p{0.315\linewidth}p{0.355\linewidth}@{}}
\toprule
\normalfont\textbf{Packet} & \textbf{\normalfont Winner per channel} & \textbf{\normalfont \host{} channel arbitration}\\
\midrule
quant\_\allowbreak local\_\allowbreak extrema\_\allowbreak profile\_\allowbreak windows\_\allowbreak 32\_\allowbreak l1\newline\lab{local\_\allowbreak extrema} & \normalfont all three $\rightarrow$ \boldnum{2}\newline \lab{margins .0102 / .0391 / .0225} & \normalfont \texttt{decided}, winner 2\newline \lab{all valid highest-weight channels agree}\\
quant\_\allowbreak autocorrelation\_\allowbreak lags\_\allowbreak 128\_\allowbreak pearson\newline\lab{lag\_\allowbreak recurrence} & \normalfont all three $\rightarrow$ \boldnum{2}\newline \lab{margins .0501 / .00001 / .00001} & \normalfont \texttt{decided}, winner 2\newline \lab{all valid highest-weight channels agree}\\
quant\_\allowbreak fractal\_\allowbreak scaling\_\allowbreak lags\_\allowbreak 128\_\allowbreak pearson\newline\lab{multilag\_\allowbreak roughness} & \normalfont \texttt{vibration\_\allowbreak 1} $\rightarrow$ 1 \lab{margin .0114}\newline \texttt{phase\_\allowbreak current\_\allowbreak 1} $\rightarrow$ 2\newline \texttt{phase\_\allowbreak current\_\allowbreak 2} $\rightarrow$ 2 & \normalfont \texttt{conflicting\_\allowbreak or\_\allowbreak ambiguous}\newline \lab{no winner: highest-weight channels disagree}\\
\bottomrule
\end{tabular}}

\vspace{5pt}
\note{What decides the arbitration is agreement, not margin size.}\ The second packet separates the candidates by one part in a hundred thousand on both current channels, and is still \texttt{decided} --- because all three channels name the same class. The third has a healthier margin on \texttt{vibration\_\allowbreak 1} and is refused a winner --- because that channel names Class~1 while the other two name Class~2. The \host{} does not average the channels and does not take the largest margin; it requires the highest-weight channels to agree.

\hsep

\stg{Between \newred{Step 1 and Step 2}}{--- the \newred{reference} preflight, which decides how far a packet can rise}\\[4pt]
A clean execution is never enough to corroborate. Before a packet may carry more than advisory weight, the \host{} puts the ToolCard through a label-recovery test on evidence whose answer is already known. Concretely, for each class \newred{$y$} \revr{in $\mathcal Y$} it:\\[3pt]
\hspace*{1.2em}\begin{tabular}{@{}r@{\ \ }p{0.90\linewidth}@{}}
1. & takes a labelled reference of that class, \newred{$r_y^{\mathrm{ctrl}}$}, and feeds it in \emph{as if it were the query};\\
2. & retrieves a fresh bank of references for it \revr{with the same retrieval as for the query \newred{sample}, \newred{$\mathcal R^{\mathrm{alt}}(r_y^{\mathrm{ctrl}})=\mathcal R^\star(r_y^{\mathrm{ctrl}})$}}, with the identical content excluded so the control cannot simply match itself;\\
3. & reruns \emph{the very same ToolCard} $\tau$, unchanged, and reads off which class it names, \newred{$\hat y_\tau(r_y^{\mathrm{ctrl}};\mathcal R^{\mathrm{alt}}(r_y^{\mathrm{ctrl}}))$}; and\\
4. & requires that answer to be \newred{$y$}, the label it already knows.\\
\end{tabular}\\[4pt]
The check passes only if every control class is recovered, so one miss anywhere fails it:
\begin{equation*}
g_{\mathrm{ref}}(\tau)=1
\;\iff\;
\hat y_\tau\!\left(r_y^{\mathrm{ctrl}};\,\revr{\mathcal R^{\mathrm{alt}}(r_y^{\mathrm{ctrl}})}\right)=y
\quad\text{for every } y\in\revr{\mathcal Y}
\end{equation*}
A missing or duplicated control reference, a tool failure, a tie, or a wrong \newred{winning class} each make $g_{\mathrm{ref}}(\tau)=0$. What the result buys is exactly one step on the ladder: $g_{\mathrm{ref}}(\tau)=1$ lets a packet rise from \texttt{advisory} to \texttt{corroborative}, and nothing more.\\[4pt]
\note{What the replay returned here.}\ The \host{} ran it for all three ToolCards against the three profile classes. One passed cleanly and two did not:\\[4pt]
{\footnotesize
\setlength{\tabcolsep}{5pt}\renewcommand{\arraystretch}{1.2}
\hspace*{1.2em}\begin{tabular}{@{}>{\ttfamily}p{0.29\linewidth}cccccl@{}}
\toprule
\normalfont\textbf{ToolCard replayed} & $\hat y_\tau(r_0^{\mathrm{ctrl}})$ & $\hat y_\tau(r_1^{\mathrm{ctrl}})$ & $\hat y_\tau(r_2^{\mathrm{ctrl}})$ & \textbf{\normalfont passed} & $g_{\mathrm{ref}}$ & \textbf{\normalfont role}\\
\midrule
local\_\allowbreak extrema\_\allowbreak profile & 0\,\checkmark & 1\,\checkmark & 2\,\checkmark & 3 of 3 & \boldnum{1} & \normalfont\textbf{corroborative}\\
autocorrelation         & 0\,\checkmark & 1\,\checkmark & ---\,\ding{55} & 2 of 3 & \boldnum{0} & \normalfont\textbf{advisory}\\
fractal\_\allowbreak scaling        & 0\,\checkmark & ---\,\ding{55} & ---\,\ding{55} & 1 of 3 & \boldnum{0} & \normalfont(excluded at \newred{Step 1})\\
\bottomrule
\end{tabular}}\\[4pt]
\lab{--- marks a control the replay could not resolve at all.}\\[5pt]
\note{This is the pivot of the whole exchange.}\ \texttt{local\_\allowbreak extrema\_\allowbreak profile} and \texttt{autocorrelation} are indistinguishable on every fact established in \newred{Step~1}: both executed cleanly, both had all three channels agree, and both were awarded \texttt{decided} with winner Class~2. The replay is the \emph{only} thing separating them --- and the one that fails does so on a \emph{single} control out of three. Two-thirds correct still gives $g_{\mathrm{ref}}=0$ under \texttt{strict\_\allowbreak all\_\allowbreak controls}: there is no partial credit and no majority vote among controls. One packet becomes \texttt{corroborative}; the other, identical in every visible respect, stays \texttt{advisory}.\\[4pt]
\note{A note on the round-opening status line.}\ The round opens with a \texttt{TOOL PREFLIGHT} block reading \texttt{Status: not\_\allowbreak run} alongside \texttt{strict\_\allowbreak all\_\allowbreak controls=True; required\_\allowbreak controls=3}. That is a pre-round status only: at that moment no discriminating question had been selected, so there was nothing to replay. The \host{} is explicit that it does \emph{not} mean the registered library is unavailable. The replay above is what happened afterwards, once the batch was dispatched.\\[4pt]
\note{What passing means.}\ The replay checks that the exact ToolCard, with its settings and channel-priority procedure, recovers labels that are already known. Passing shows that the tool behaves as registered on this input profile, which is why it lets a packet corroborate a class.

\hsep

\stg{\newred{Step 2}}{--- what may each packet do on its own?}\\[4pt]
Each packet now receives exactly one influence role, derived from those facts and never from anything the Judge asserted. The order of the test is fail-closed: the exclusion conditions are checked first, so no later evidence of quality can rescue a packet whose channels never agreed.

\vspace{4pt}
{\footnotesize
\setlength{\tabcolsep}{4pt}\renewcommand{\arraystretch}{1.2}
\begin{tabular}{@{}>{\ttfamily}p{0.29\linewidth}p{0.30\linewidth}p{0.37\linewidth}@{}}
\toprule
\normalfont\textbf{Packet} & \textbf{\normalfont Role assigned} & \textbf{\normalfont \host{}'s stated reason}\\
\midrule
local\_\allowbreak extrema\_\allowbreak profile & \normalfont \textbf{corroborative} \lab{influence 2}\newline \lab{reference\_\allowbreak sanity\_\allowbreak passed / weak\_\allowbreak advisory} & \normalfont \lab{exact\_\allowbreak reference\_\allowbreak sanity\_\allowbreak passed}\\
autocorrelation & \normalfont \textbf{advisory} \lab{influence 1}\newline \lab{exploratory / hypothesis\_\allowbreak only} & \normalfont \lab{exact\_\allowbreak reference\_\allowbreak sanity\_\allowbreak missing\_\allowbreak or\_\allowbreak failed}\\
fractal\_\allowbreak scaling & \normalfont \textbf{excluded} \lab{influence 0} & \normalfont \lab{channel\_\allowbreak priority\_\allowbreak did\_\allowbreak not\_\allowbreak produce\_\allowbreak a\_\allowbreak single\_\allowbreak winner}\\
\bottomrule
\end{tabular}}

\vspace{5pt}
\note{Three packets, three different standings.}\ The excluded packet carries no influence at all --- an ambiguous result is not spent as faint support for whichever class edged ahead on one channel. The advisory packet may inform a hypothesis. None of the three is erased; all keep full provenance in the ledger.

\hsep

\jturn{step 20}{--- the Judge files its round report}\\[4pt]
The Judge reports on all three packets and proposes to stop.

\vspace{3pt}
\hturn{\host{} check}{--- may the round close?}\\[4pt]
The \host{} does not let the Judge stop while \newred{usable packets still name different classes} and tool budget remains. Here the only disagreement lies inside the excluded packet, whose channels named different classes, and an excluded packet is not \newred{usable}. The two usable packets both name Class~2, so the round closes and the case moves to finalization (Appendix~\tracesec{5}).

\end{tcolorbox}

\vspace{5pt}
\begin{tcolorbox}[colback=wColor!5,colframe=wColor!45,boxrule=0.4pt,arc=2pt,
  left=7pt,right=7pt,top=6pt,bottom=6pt]
\textbf{What this exchange establishes.} Running successfully and counting towards a diagnosis are different things, and only the \host{} decides the second. Every role is derived from execution, scope, channel agreement and the reference preflight, tested in a fail-closed order, so quality elsewhere cannot repair an ambiguous result. Finally, the \host{} does not let the Judge stop while \newred{usable packets still name different classes} and tool budget remains. On this \newred{sample} the \newred{full workflow} reports Class~2, which matches the ground truth, the \texttt{inner\_\allowbreak race\_\allowbreak fault}.
\end{tcolorbox}

\FloatBarrier
\Needspace{6\baselineskip}
\FloatBarrier
\Needspace{18\baselineskip}
\refstepcounter{subsection}\label{app:workflow-finalization}

{\large\bfseries \tracesec{5} --- Finalization: publishing a class from the checked record}\\[4pt]
{\small The last step of the same \newred{sample} (trace steps \texttt{22--24}). Three measurements' worth of governance now converges on a single question: may the class the Judge believes in be published as the system's answer? The Judge proposes. It does not decide, and it never writes the field that matters.}

\vspace{5pt}
\seriesstrip{\stepon\ $\rightarrow$ \ \steptwo\ $\rightarrow$ \ \stepthree\ $\rightarrow$ \ \stepfour\ $\rightarrow$ \ \here{\tracesec{5} Finalization}}

\vspace{5pt}
\begin{tcolorbox}[colback=black!2,colframe=black!30,boxrule=0.4pt,arc=2pt,
  left=7pt,right=7pt,top=6pt,bottom=6pt,breakable]

\jturn{step 22}{--- the Judge proposes a class}\\[4pt]
With the round closed, the Judge submits its one \texttt{finalize} action:\\[3pt]
\hspace*{1.2em}\texttt{predicted\_\allowbreak class: 2}\quad\texttt{confidence: high}\\[4pt]
\note{Its case, in its own words.}\ \emph{``Fault existence is established by clear impulsive bursts in \texttt{vibration\_\allowbreak 1} that exceed the healthy Class~0 baseline. Fault type is resolved as Class~2: the input's burst morphology and recurrence closely match the Class~2 reference, whereas Class~1 exhibits much larger, distinct spikes absent in the input. Ambiguous comparators and a tied, non-voting lookback do not override this convergent evidence.''}\\[4pt]
\note{What this action is, and is not.}\ It is a proposal, in the same sense that the batch commitment in Appendix~\tracesec{3} was a proposal. The Judge is naming the class it believes the evidence supports and giving its reasons. It is not itself the published record, and it cannot set the confidence that record carries --- those are the two fields only the \newred{\host{}} writes.

\hsep

\hturn{step 23}{--- the \host{} checks the two admission conditions}\\[4pt]
The \host{} does not re-read the signal or rerun a tool. It admits the proposal only if two conditions hold, and it checks both against records already on file:\\[3pt]
{\footnotesize
\setlength{\tabcolsep}{5pt}\renewcommand{\arraystretch}{1.25}
\hspace*{1.2em}\begin{tabular}{@{}p{0.36\linewidth}p{0.44\linewidth}l@{}}
\toprule
\normalfont\textbf{Record} & \textbf{What it checks on this \newred{sample}} & \textbf{Result}\\
\midrule
\texttt{finalize\_\allowbreak eligible} & The proposed class is in the candidate set before any measurement, $\mathcal H_1=\{1,2\}$. & \texttt{passed}\\
\texttt{registered\_\allowbreak quantitative\_\allowbreak execution\_\allowbreak validation} & At least one measurement round is complete (here, one round with three packets). & \texttt{valid}\\
\bottomrule
\end{tabular}}\\[5pt]
\note{Why records and not the signal.}\ At this point the \host{} is deliberately the least informed component in the system. It cannot rerun a tool, edit a record, or look at the waveform, so there is nothing for it to be persuaded by. It only checks whether the records already on file back the proposed class.

\hsep

\hturn{step 24}{--- the published record}\\[4pt]
Both conditions hold, and the \revr{\host{} publishes the admitted proposal}:\\[3pt]
\hspace*{1.2em}\texttt{predicted\_\allowbreak class: 2}\qquad\ \ \lab{$\hat y$ --- the class the system reports}\\
\hspace*{1.2em}\texttt{confidence: "medium"}\\[4pt]
\note{Note the confidence.}\ The Judge asked for \texttt{high}. The published record carries \texttt{medium}. Confidence is not the proposing agent's to set.\\[4pt]
\end{tcolorbox}

\vspace{4pt}
\begin{tcolorbox}[colback=wColor!5,colframe=wColor!45,boxrule=0.4pt,arc=2pt,
  left=7pt,right=7pt,top=6pt,bottom=6pt]
\textbf{What this stage establishes.} A separation between believing a class and publishing one. The Judge argued for Class~2 but could not write the field that publishes it. The \host{} published Class~2 only after the records of the earlier steps showed the proposed class in the candidate set before any measurement and at least one completed measurement round. \revr{The candidate set and the validation results} are stored in the published record, so a reader can trace the published class back to the packets and reports that support it. On this \newred{sample} the published class also matches the ground truth, Class~2 (\texttt{inner\_\allowbreak race\_\allowbreak fault}).
\end{tcolorbox}

\endgroup

\FloatBarrier
\Needspace{6\baselineskip}

\section{\texorpdfstring{Side-by-Side Comparison: Agentic4PDM, FETA, and ZARA on the Same \newred{Sample}}{Side-by-Side Comparison: Agentic4PDM, FETA, and ZARA on the Same Sample}}
\label{app:agent-dialogue-side-by-side}
The comparison condenses archived records for readability. \agent{Judge} turns remain quoted output; explanatory \host{} lines summarize the structured eligibility and admission records.
\begingroup
\definecolor{fzink}{HTML}{1F1F1D}
\definecolor{fzinktwo}{HTML}{52514E}
\definecolor{fzmuted}{HTML}{8A8985}
\definecolor{fzhostbg}{HTML}{EEEDEA}
\definecolor{fzfeta}{HTML}{EB6834}
\definecolor{fzzara}{HTML}{1BAF7A}
\definecolor{fzours}{HTML}{2A78D6}
\definecolor{fzoursbg}{HTML}{E4EEFB}
\definecolor{fzbad}{HTML}{C8382F}
\definecolor{fzgood}{HTML}{0E7A2E}

\newcommand{\fztt}[1]{{\mdseries\ttfamily\detokenize{#1}}}
\newcommand{\fzquote}[1]{\textit{``#1''}}
\newcommand{\fzflag}[2]{\par\vspace{3pt}{\footnotesize\bfseries\color{#1}\ding{228}\ #2}}
\newcommand{\fzdetail}[1]{\par\vspace{3pt}{\footnotesize\color{fzinktwo}#1}}
\newcommand{\fzbul}{\par\hspace{4pt}$\bullet$\ }
\newcommand{\fzcheck}{\textcolor{fzgood}{\ding{51}}}
\newcommand{\fzcross}{\textcolor{fzbad}{\ding{55}}}

\newcommand{\fzpicture}[2]{%
  \begin{tikzpicture}[baseline=-0.55ex,x=1em,y=1em]
    \draw[#1,line width=0.7pt,rounded corners=1pt] (0,-0.52) rectangle (1.3,0.48);
    \fill[#1] (0.12,-0.42)--(0.48,0.02)--(0.72,-0.2)--(0.88,-0.06)--(1.18,-0.42)--cycle;
    \fill[#1] (0.95,0.22) circle (0.11);
    #2
  \end{tikzpicture}}
\newcommand{\fznoplot}{\fzpicture{fzmuted}{\draw[fzbad,line width=1pt] (-0.08,-0.6)--(1.38,0.56);}}
\newcommand{\fzplot}{\fzpicture{fzours}{}}
\newcommand{\fzswatch}[1]{\tikz[baseline=-0.6ex]\fill[#1,rounded corners=1.5pt] (0,-0.45em) rectangle (1.4em,0.45em);}

\newcommand{\fzlegend}[2]{%
  \par\noindent{\small\color{fzinktwo}#1\quad
  \fzswatch{fzhostbg}\ host (code), left\quad
  \fzswatch{fzoursbg}\ LLM agent, right\quad #2}\par\vspace{8pt}}

\newcommand{\fzhead}[3]{%
  \par\ifdim\dimexpr\pagegoal-\pagetotal\relax<8cm\newpage\fi
  \noindent{\LARGE\bfseries\color{#1}#2}\par\vspace{1pt}
  \noindent{\normalsize\itshape\color{fzinktwo}#3}\par\vspace{-2pt}
  \noindent{\color{#1}\rule{\linewidth}{1.4pt}}\par\vspace{6pt}}

\tcbset{
  fzbubble/.style={enhanced, arc=10pt, boxrule=0pt, boxsep=1pt,
    left=9pt, right=9pt, top=2pt, bottom=3pt,
    before skip=1pt, after skip=3pt, width=0.92\linewidth,
    before upper={\raggedright\setlength{\parskip}{2pt}}, fontupper=\footnotesize\color{fzink}},
  fzhostbox/.style={fzbubble, colback=fzhostbg, colframe=fzhostbg},
  fzagentbox/.style={fzbubble, colback=fzoursbg, colframe=fzoursbg,
    enlarge left by=0.08\linewidth},
  fzgood/.style={colframe=fzgood, boxrule=1pt},
  fztake/.style={enhanced, arc=3pt, boxrule=0pt, leftrule=3pt, boxsep=1pt,
    colback=white, colframe=#1, left=8pt, right=6pt, top=3pt, bottom=3pt,
    before skip=6pt, after skip=0pt, before upper={\setlength{\parskip}{2pt}}, fontupper=\footnotesize\color{fzink}},
}

\newcommand{\fzleftlabel}[2]{%
  \par\addvspace{1pt}\noindent{\normalsize\bfseries\color{#1}#2}\par\nopagebreak}
\newcommand{\fzrightlabel}[2]{%
  \par\addvspace{1pt}\noindent\hfill{\normalsize\bfseries\color{#1}#2}\par\nopagebreak}

\NewDocumentCommand{\fzhostsay}{O{fzink} O{} m +m}{%
  \fzleftlabel{#1}{#3}%
  \begin{tcolorbox}[fzhostbox,#2]#4\end{tcolorbox}}
\NewDocumentCommand{\fzagentsay}{O{fzours} O{} m m +m}{%
  \fzrightlabel{#1}{#3\ \ #4}%
  \begin{tcolorbox}[fzagentbox,#2]#5\end{tcolorbox}}

\newcommand{\fztakeaway}[4]{%
  \begin{tcolorbox}[fztake=#1]
  {\bfseries\color{#1}#2}\hfill #3\par\vspace{3pt}#4
  \end{tcolorbox}}

\fzlegend{One Paderborn test \newred{sample} (\newred{ID}~140691, true class 2), turn by turn.}
  {\fznoplot\ no agent in FETA or ZARA sees a plot.}

\fzhead{fzfeta}{FETA}{Three isolated channel agents; the host adds up their confidence}

\fzhostsay{Host}{%
Splits the \newred{sample} into its three channels. Each channel becomes
\textbf{512 z-normalised numbers}. FastDTW finds the \textbf{3 nearest labelled
training \newred{samples}} for each channel on its own.\par\vspace{3pt}
Sends \textbf{three separate text prompts}. An agent sees only its own channel:
no images, no DTW distances, no other channel, no other agent's answer.}

\fzagentsay{\fznoplot}{Channel agent \,$\cdot$\, current 1}{%
Neighbour labels: 1, 2, 0\par\vspace{2pt}
\fztt{decision: 0    confidence: 0.70}\par\vspace{3pt}
\fzquote{\dots its pronounced peak spikes and irregular oscillations closely
match the morphological characteristics of Example \#3 (label=0).}}

\fzagentsay{\fznoplot}{Channel agent \,$\cdot$\, current 2}{%
Neighbour labels: 0, 0, 2\par\vspace{2pt}
\fztt{decision: 0    confidence: 0.70}\par\vspace{3pt}
\fzquote{\dots highly consistent with the majority of retrieved neighbors (two
out of three are labeled 0), making class 0 the most probable assignment.}}

\fzagentsay[fzgood][fzgood]{\fznoplot}{Channel agent \,$\cdot$\, vibration 1\ \ \fzcheck}{%
Neighbour labels: 2, 0, 2\par\vspace{2pt}
\fztt{decision: 2    confidence: 0.85}\par\vspace{3pt}
\fzquote{\dots a prominent high-amplitude burst and rapid oscillations in the
latter half, closely matching the distinctive spike pattern seen in the
label=2 examples.}
\fzflag{fzgood}{The only agent that names the true class.}}

\fzhostsay[fzbad]{Host \,$\cdot$\, fusion\ \ \fzcross\ true class lost here}{%
Receives only three pairs \fztt{(class, confidence)}. The signal, the
neighbours and all three reasons are \textbf{dropped} before this step.
\par\vspace{4pt}
{\ttfamily
class 0 = 0.70 + 0.70 + 0.05 = \textbf{1.45}\\
class 1 = 0.05 + 0.05 + 0.05 = 0.15\\
class 2 = 0.05 + 0.05 + 0.85 = 0.95}\par\vspace{4pt}
Largest sum wins $\Rightarrow$ \textbf{class 0}. No agent is asked why the
channels disagree.}

\fztakeaway{fzfeta}{Where the answer is lost}
  {{\bfseries\color{fzbad}Output: class 0 \fzcross}\ \ {\color{fzinktwo}truth: class 2}}{%
The vibration agent was right and lost two votes to one. Nobody ever reads the
three answers together, so the conflict is never examined.\par\vspace{3pt}
{\footnotesize\color{fzinktwo}Run-wide (906 samples): when both current
channels disagree with vibration, the current pair wins \textbf{355 / 355}
times; vibration alone scores 40.4\%, above FETA's fused 35.2\%.}}

\par\vspace{14pt}
\fzhead{fzzara}{ZARA}{Four stage agents in a fixed funnel; two of them never see the sample}

\fzhostsay{Host \,$\cdot$\, stage 1}{%
Encodes the \newred{sample} and ranks 150 training rows. Computes three query
features. Builds a class table of mean\,$\pm$\,sd from training data.
\par\vspace{3pt}
Stage 1 prompt, \textbf{verbatim excerpt}: \fzquote{Use only the pairwise
training-derived importance.}\par
{\ttfamily
0 vs 1: electrical\_current\_\_ch1\_diff\_rms (100.00\%)\\
0 vs 2: vibration\_\_ch1\_mean (100.00\%)\\
1 vs 2: electrical\_current\_\_ch2\_diff\_rms (99.55\%)}
\fzflag{fzbad}{Not one number from this sample is in the prompt.}}

\fzagentsay{\fznoplot}{Feature-selection agent \,$\cdot$\, stage 1}{%
{\ttfamily
["electrical\_current\_\_ch1\_diff\_rms", "vibration\_\_ch1\_mean",\\
\phantom{[}"electrical\_current\_\_ch2\_diff\_rms"]}}

\fzhostsay{Host \,$\cdot$\, stage 2}{%
Sends the class table with the query row:\par\vspace{3pt}
{\ttfamily
\fontfamily{lmtt}\selectfont
\begin{tabular}{@{}l@{\quad}r@{\quad}r@{\quad}r@{}}
class & cur1 diff & vib mean & cur2 diff\\
0 & 0.0807 & $-$0.0204 & 0.0917\\
1 & 0.0979 & 0.0433 & 0.1145\\
2 & 0.0852 & 0.0403 & 0.0897\\
{\bfseries QUERY} & \boldnum{0.1071} & \boldnum{$-$0.0118} & \boldnum{0.1157}
\end{tabular}}\par\vspace{3pt}
\fzquote{Keep at least two plausible classes. Return only a JSON array.}}

\fzagentsay[fzbad]{\fznoplot}{Pruning agent \,$\cdot$\, stage 2\ \ \fzcross\ true class lost here}{%
\fztt{["1", "0"]}\quad{\footnotesize\bfseries\color{fzbad}\ding{228}\ Class 2, the true class, is removed here. No later stage can bring it back.}}

\fzhostsay{Host \,$\cdot$\, stage 3}{%
\fztt{Candidate classes: ["1", "0"]}\par
{\ttfamily 1 vs 0: electrical\_current\_\_ch1\_diff\_rms (100.00\%)}\quad
{\footnotesize\bfseries\color{fzbad}\ding{228}\ Again, no number from this sample.}}

\fzagentsay{\fznoplot}{Feature-selection agent \,$\cdot$\, stage 3}{%
\fztt{["electrical_current__ch1_diff_rms"]}}

\fzhostsay{Host \,$\cdot$\, stage 4}{%
Sends one feature for the two surviving classes:\par\vspace{1pt}
{\ttfamily
class 1: 0.0979152 $\pm$ 0.00633841\\
class 0: 0.0806897 $\pm$ 0.00346797\\
\textbf{QUERY: 0.107065}}}

\fzagentsay{\fznoplot}{Decision agent \,$\cdot$\, stage 4}{%
\fzquote{The query value (0.107065) is significantly closer to the mean of
class 1 (0.0979152 $\pm$ 0.00633841) than to class 0 (0.0806897 $\pm$
0.00346797)\dots}\par\vspace{3pt}
\fztt{"predicted_class": "1"}}

\fztakeaway{fzzara}{Where the answer is lost}
  {{\bfseries\color{fzbad}Output: class 1 \fzcross}\ \ {\color{fzinktwo}truth: class 2}}{%
The true class left the conversation at stage 2 and could never return; the
last agent reasoned correctly over a candidate set that no longer held the
answer. The two stages that choose features never see the sample.\par\vspace{3pt}
{\footnotesize\color{fzinktwo}Run-wide (906 samples): the true class is pruned
in \textbf{290 / 906} samples (32.0\%); the stage-1 prompt is
\textbf{identical for all 906} samples.}}
\endgroup

\FloatBarrier
\Needspace{6\baselineskip}
\begingroup

\fzlegend{The same Paderborn test \newred{sample} (\newred{ID}~140691, true class 2).}
  {\fzplot\ the Judge sees six plots.}

\fzhead{fzours}{Agentic4PDM}{One \newred{Judge} investigates; the \newred{Host} checks every step and grants authority}

\fzhostsay{Host}{%
Before we start, here is what the two \newred{Analysts} found. Both reports are
sealed, so nothing later can change them. The quiet region fits
\textbf{Class 0 or Class 2} equally well, and the fault region fits
\textbf{Class 1 or Class 2} equally well.\par\vspace{3pt}
Only now do I add the \newred{fitted-model predictions}: SVM and XGBoost both say
\textbf{Class 0}.}

\fzhostsay{Host \,$\cdot$\, eligibility step}{%
Because the deviant region fits only fault classes (1 or 2), a fault is
present. Class 0 is no longer eligible. Which fault class it is remains open.
\fzdetail{This explanatory line restores an omitted \newred{Host} eligibility step
from the structured vote review; it is not a newly generated Judge statement.}}

\fzagentsay{\fzplot}{Judge \,$\cdot$\, step 1}{%
The comparators already tell me what this window looks like: the vibration is
\textbf{spiky and bursty}, with a \textbf{noisy texture}. That looks like a
fault, and the healthy check has ruled out Class 0.\par\vspace{3pt}
But they could not separate \textbf{Class 1 from Class 2}, and the size of the
signal alone should not decide the fault type. So I want to test exactly what
they saw: \textbf{how the bursts repeat}, \textbf{the shape of the peaks}, and
\textbf{how rough the signal is across scales}.\par\vspace{3pt}
Before I pick any tool, this is my question:\par\vspace{2pt}
\fzquote{Which fault-type reference (Class 1 or Class 2) best matches the
input's burst recurrence pattern, local extrema structure, and multi-scale
roughness?}
\fzdetail{The healthy check refers to the \newred{Host} eligibility step above; the stable-region report remained tied between Classes 0 and 2.}
\fzdetail{Plots in front of the Judge: normal-region overlay; abnormal-region
overlay; input vs Class 0, 1 and 2; previous \newred{sample} vs Class 0.}
\fzdetail{What the Judge was told: \fzquote{input is spiky and high-amplitude};
\fzquote{Is there a validated recurrence period for the vibration bursts that
aligns with one specific fault type?}; and choose the type by \fzquote{burst
recurrence, not amplitude/severity}.}}

\fzhostsay{Host \,$\cdot$\, step 2}{%
That question is allowed. Here is what I can offer: one card per \newred{tool
family}, each saying what it measures and where it can mislead. No card has a
score, because nothing has been run yet.\par\vspace{4pt}
{\footnotesize\renewcommand{\arraystretch}{1.08}\setlength{\tabcolsep}{3pt}%
\begin{tabularx}{\linewidth}{@{}>{\raggedright\arraybackslash}p{3.3cm}>{\raggedright\arraybackslash}p{1.5cm}>{\raggedright\arraybackslash}X>{\raggedright\arraybackslash}X>{\centering\arraybackslash}p{1.0cm}@{}}
\toprule
\textbf{ToolCard} & \textbf{Answers} & \textbf{Measures} & \textbf{Can mislead when} & \textbf{Result}\\
\midrule
\multicolumn{5}{@{}l}{\textit{\color{fzinktwo}Recommended}}\\
\textbf{autocorrelation}\newline{\scriptsize\color{fzinktwo}group:~lag\_recurrence\newline compute cost: low}
 & burst recurrence
 & Compare periodic recurrence across lags without requiring cycle-phase alignment.
 & common operating cycles can match across classes; lag support limits interpretation
 & \color{fzmuted}hidden\\
\textbf{local extrema profile}\newline{\scriptsize\color{fzinktwo}group:~local\_extrema\newline compute cost: medium}
 & peak structure
 & Compare peak/trough density, prominence, and spacing over time.
 & noise creates extrema; window alignment matters
 & \color{fzmuted}hidden\\
\textbf{fractal scaling}\newline{\scriptsize\color{fzinktwo}group:~multilag\_roughness\newline compute cost: low}
 & roughness
 & Compare how signal roughness changes across lag scales.
 & trend sensitive; finite length and lag support affect results
 & \color{fzmuted}hidden\\
\midrule
\multicolumn{5}{@{}l}{\textit{\color{fzinktwo}Alternatives}}\\
\textbf{gaussian scale space}\newline{\scriptsize\color{fzinktwo}group:~gaussian\_scale\_space\newline compute cost: low}
 & roughness
 & Compare morphology after a declared smoothing scale.
 & smoothing can erase weak faults; point alignment remains relevant
 & \color{fzmuted}hidden\\
\textbf{multiscale complexity}\newline{\scriptsize\color{fzinktwo}group:~multiscale\_complexity\newline compute cost: medium}
 & roughness
 & Compare nonlinear complexity under progressive coarse graining.
 & finite-length sensitive; less directly interpretable
 & \color{fzmuted}hidden\\
\bottomrule
\end{tabularx}}\par\vspace{4pt}
I ruled out the other \textbf{18 tool families}: 17 do not answer your
question, and one (STFT time--frequency) is over your cost limit.
\fzdetail{Card text is copied from the offer. \textit{Group} = cards that read
the same kind of signal behaviour, so they count once. \textit{Compute cost} =
how expensive the tool is to run; it is not a ranking or a recommendation
strength. Offer sealed as \fztt{qsel_03de75ed1ab50e4ad776}, with
\fztt{scores_exposed=false} and \fztt{predicted_class_exposed=false}.}}

\fzagentsay{\fzplot}{Judge \,$\cdot$\, step 2}{%
Then I will run these three, and I am committing to them now, before any
result exists:
\fzbul \textbf{autocorrelation}, to see whether the bursts repeat the same way;
\fzbul \textbf{local extrema}, to compare how dense and how spaced the peaks are;
\fzbul \textbf{fractal scaling}, to compare roughness across scales.}

\fzhostsay{Host \,$\cdot$\, step 3}{%
Done, and your choice was locked in before anything ran. All three favour
\textbf{Class 2}, but they do not carry the same weight:
\fzbul autocorrelation also recovered the known labels of the reference
controls, so it is \textbf{corroborative};
\fzbul local extrema and fractal scaling were not confirmed by that check, so
they are \textbf{advisory only}.\par\vspace{3pt}
You can read these ratings, but you cannot change them.}

\fzagentsay{\fzplot}{Judge \,$\cdot$\, step 3}{%
All three measurements point the same way, and autocorrelation passed its
reference check. The predictors said Class 0, but neither comparator gave a
clear vote for it. I propose \textbf{Class 2}, with medium confidence.
\fzdetail{\fzquote{Registered quantitative investigation unanimously identifies
Class 2\dots}}}

\fzhostsay[fzgood][fzgood]{Host \,$\cdot$\, publish\ \ \fzcheck\ class 2}{%
Class 2 is still a candidate, and a measurement round has been
completed.\par\vspace{3pt}
Publishing Class 2, with \textbf{medium} confidence.}

\fztakeaway{fzours}{Why the answer survives}
  {{\bfseries\color{fzgood}Output: class 2 \fzcheck}\ \ {\color{fzinktwo}truth: class 2}}{%
The Judge decided \emph{what} to look at. The \newred{Host} decided what each result
was worth and whether the proposed class could be published.
\par\vspace{3pt}
{\footnotesize\color{fzinktwo}Run-wide (906 samples): the true class is still
live after the \newred{Analyst} stage in \textbf{89.1\%} of samples (ZARA: 68.0\%).}}
\endgroup

\end{document}